\documentclass[10pt,twocolumn,letterpaper]{article}
\newcommand{\M}{TINC}
\usepackage[title]{appendix}
\usepackage{cvpr}      % To produce the REVIEW version
\usepackage{titling}

\date{}
\usepackage{graphicx}
\usepackage{amsmath}
\usepackage{amssymb}
\usepackage{booktabs}
\usepackage{multirow}
\usepackage{float}
\usepackage[pagebackref,breaklinks,colorlinks]{hyperref}

\usepackage[capitalize]{cleveref}
\crefname{section}{Sec.}{Secs.}
\Crefname{section}{Section}{Sections}
\Crefname{table}{Table}{Tables}
\crefname{table}{Tab.}{Tabs.}

\def\confName{CVPR}
\def\confYear{2023}

\begin{document}
\title{TINC: Tree-structured Implicit Neural Compression}

\author{Runzhao Yang, Tingxiong Xiao, Yuxiao Cheng, Jinli Suo, Qionghai Dai\\
Department of Automation, Tsinghua University\\
Beijing 100084, China\\
{\tt\small jlsuo@tsinghua.edu.cn}
% \and
}
\maketitle
%%%%%%%%% ABSTRACT
\begin{abstract}
Implicit neural representation (INR) can describe the target scenes with high fidelity using  a small number of parameters, and is emerging as a promising data compression technique. However, limited spectrum coverage is intrinsic to INR, and it is non-trivial to remove redundancy in diverse complex data effectively. Preliminary studies can only exploit either global or local correlation in the target data and thus of limited performance. 
In this paper, we propose a Tree-structured Implicit Neural Compression (\M) to conduct compact representation for local regions and extract the shared features of these local representations in a hierarchical manner. Specifically, we use Multi-Layer Perceptrons (MLPs) to fit the partitioned local regions, and these MLPs are organized in tree structure to share parameters according to the spatial distance. The parameter sharing scheme not only ensures the continuity between adjacent regions, but also jointly removes the local and non-local redundancy. 
Extensive experiments show that \M~improves the compression fidelity of INR, and has shown impressive compression capabilities over commercial tools and other deep learning based methods. Besides, the approach is of high flexibility and can be tailored for different data and parameter settings.
The source code can be found at \href{https://github.com/RichealYoung/TINC}{https://github.com/RichealYoung/TINC}.
\end{abstract}

%%%%%%%%% BODY TEXT
\section{Introduction}
\label{sec:intro}
In the big data era, there exist massive and continuously growing amount of visual data from different fields, from surveillance, entertainment, to biology and medical diagnosis. The urgent need for efficient data storage, sharing and transmission all requires effective compression techniques. Although image and video compression have been studied for decades and there are a number of widely used commercial tools, the compression of medical and biological data are not readily available.

Implicit neural representation (INR) is becoming widely-used for scene rendering\cite{sitzmann2019scene,nerf,martin2021nerf,yu2021pixelnerf,ingp}, shape estimation\cite{park2019deepsdf,chen2019learning,saito2019pifu,genova2019learning}, and dynamics modeling\cite{li2021neural,pumarola2021d,park2021nerfies}. Due to its powerful representation capability, INR can describe nature scenes at high fidelity with a much smaller number of parameters than raw discrete grid representation, and thus serves as a promising data compression technique. 

%\textcolor{red}{nerv for video. but if extend its convolution into 3d conv for the 3d data }
In spite of the big potential, INR based compression is quite limited confronted with large sized data \cite{mehta2021modulated}, since INR is intrinsically of limited spectrum coverage and cannot envelop the spectrum of the target data, as analyzed in \cite{yang2022sci}. 
Two pioneering works using INR for data compression, including NeRV\cite{chen2021nerv} and SCI\cite{yang2022sci}, have attempted to handle this issue in their respective ways.
Specifically, NeRV introduces the convolution operation into INR, which can reduce the required number of parameters using the weight sharing mechanism. However,  convolution is spatially invariant and thus limits NeRV's representation  accuracy on complex data with spatial varying feature distribution. 
Differently, to obtain high fidelity on complex data, SCI adopts divide-and-conquer strategy and partitions the data into blocks within INR's concentrated spectrum envelop. This improves the local fidelity, but cannot remove non-local redundancies for higher compression ratio and tend to cause blocking artifacts. 

%demonstrated INR's superior compression capability to commercial tools and other deep learning based methods, on natural scenarios and biomedical data respectively.

\begin{figure*}[t]
  \centering
   \includegraphics[width=\linewidth]{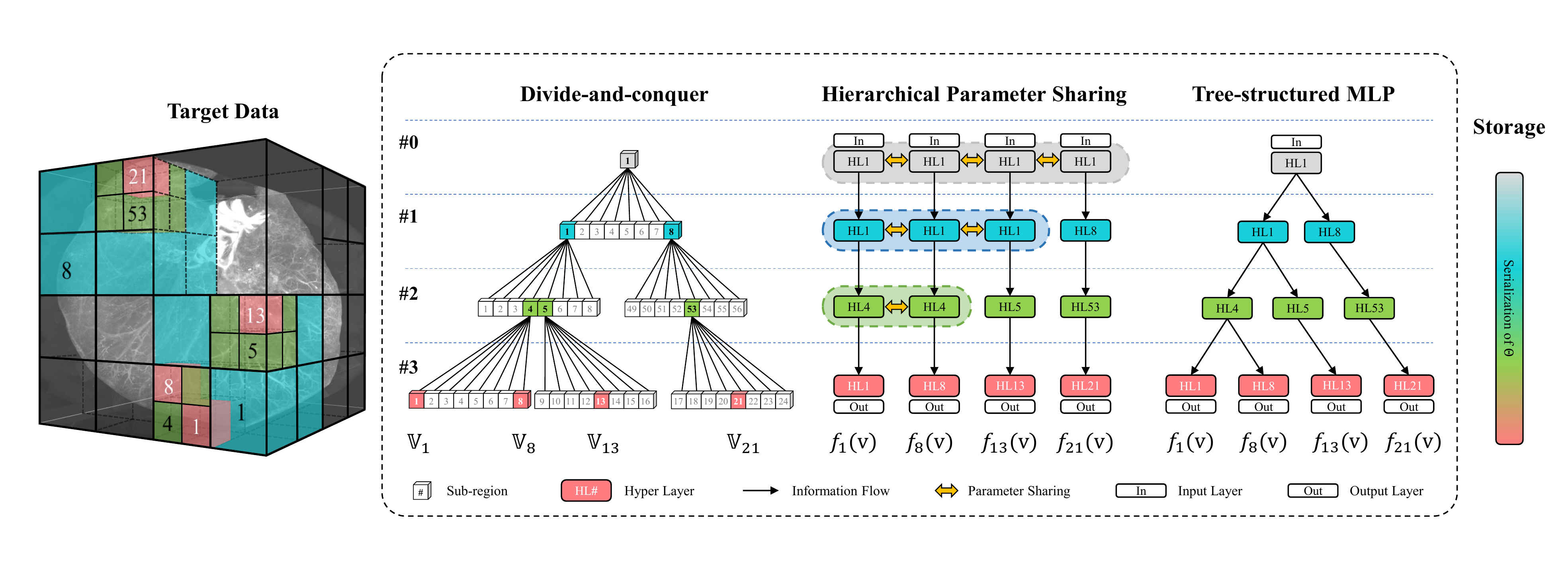}
   \caption{The scheme of the proposed approach \M. For a target data, we divide the volume into equal-size blocks via octree partitioning, with some neighboring and far-apart blocks of similar appearances. Then  each block can be representation with an implicit neural function, implemented as an MLP. After sharing parameters among similar blocks, we can achieve a more compact  neural network with a tree shaped structure. Here we highlight the similar blocks sharing network parameters with the same color.}
   \label{fig: method}
\end{figure*}

To compress large and complex data with INR, in this paper we propose to build a tree-structured Multi-Layer Perceptrons (MLPs), which consists of a set of INRs to represent local regions in a compact manner and organizes them under a hierarchical architecture for parameter sharing and higher compression ratio. 
Specifically, we first draw on the idea of ensemble learning\cite{kadarvish2021ensemble} and use a divide-and-conquer strategy\cite{mehta2021modulated,yang2022sci} to compress different regions with multiple MLPs separately. 
Then we incorporate these MLPs with a tree structure to extract their shared parameters hierarchically, following the rule that spatially closer regions are of higher similarity and share more parameters.
Such a joint representation strategy can remove the redundancy both within each local region and among non-local regions, and also ensures the continuity between adjacent regions. 
The scheme of the proposed network is illustrated in Fig.~\ref{fig: method}, and we name our approach  \M~(Tree-structured Implicit Neural Compression).

Using the massive and diverse biomedical data, we conduct extensive experiments to validate that \M~greatly improves the capability of INR and even outperforms the commercial compression tools (H.264 and HEVC) under high compression ratios. 
%We observe that \textcolor{red}{xxx}. Experiments show that our method has better results on \textcolor{red}{xxx} datasets compared to \textcolor{red}{xxx} algorithm, and even outperforms \textcolor{red}{xxx} algorithm on \textcolor{red}{xxx} data. 
The proposed \M~is also a general framework, which can be flexibly adapted to diverse data and varying settings. 
%We have also conducted an exhaustive study of the introduced parameter sharing mechanism and tree-structured MLP. We have studied the effect of number of tree levels, intra-level and inter-level parameter allocation methods on \M's compression fidelity on biomedical data with different characteristics under different compression ratios. And accordingly, the usage strategies of these features and their mechanisms are given for different experimental scenarios.

\section{Related work}
\noindent\textbf{Data compression.}
Data compression is a key technique for  efficient data storage, transmission and sharing, especially in the big data era. In recent decades, many commercial compression tools for massive visual data have become increasingly mature, like JPEG\cite{wallace1992jpeg}, JPEG2000\cite{skodras2001jpeg}, H.264\cite{wiegand2003overview}, and HEVC\cite{sullivan2012overview}, while emerging deep learning-based approaches are also gaining momentum and demonstrating continuously improving compression  capabilities\cite{lu2019dvc,minnen18joint,johannes18variational,yang20improving,agustsson2020scale,yang2020learning,yang2021hierarchical}. 
Compression for nature images and videos have been widely and successfully used, however, there still lack effective compression techniques for the huge amount of three or higher dimensional data such as biomedical volume. Image/Video compressors cannot be applied directly, because these high dimensional data is a discrete sampling of 3D organism, instead of a video describing temporal evolution of 2D slices. Hence, the optical flow algorithms that are widely applicable in video compression do not work well on such inherently spatially structured data\cite{yang2022sci}, and tend to cause discontinuity and degenerate the downstream tasks. 
Our approach differs completely from existing image/video compression methodologies in that we encode the intrinsic data structure with INR. Besides, our approach searches for the best network structure and parameters representing the target data and is free of generalization issues. 
%which is orthogonal to the parameter sharing mechanism proposed in this paper.

\vspace{2mm}
\noindent\textbf{Implicit neural compression.}
Recent works\cite{chen2021nerv,lu2021compressive,yang2022sci} have introduced INR to the field of data compression with exciting results. They fit the original discrete grid-based data with continuous implicit functions in a transform-coding-like manner, which takes advantage of the powerful representation capabilities of neural networks while avoids the generalization problem.
Obviously, the compression capability of INR is closely related to its representation capability, which can be improved from proper design of network structure \cite{chen2021nerv,mfn,lu2021compressive}, selection of activation function \cite{klocek2019hypernetwork,mehta2021modulated,siren}, embedding \cite{benbarka2021seeing,nerf,ffn,zhong2019reconstructing,ingp}, and learning strategies such as manifold learning \cite{du2021learning,mehta2021modulated,siren,sitzmann2019scene}, meta learning \cite{bergman2021fast,du2021learning,tancik2021learned} and ensemble learning \cite{aftab2021multi,kadarvish2021ensemble,niemeyer2021giraffe}.

However, improving these capability does not necessarily head for a ready compressor for general scenes or objects, since there exist large diversity in the target data and pursuing a general network with high fidelity for all the data under given parameter budget. In order to pursue a more compact and efficient representation, we draw on the idea of ensemble learning\cite{kadarvish2021ensemble} and use a divide-and-conquer strategy\cite{mehta2021modulated,yang2022sci} to compress the data in each region separately.
% We also take reference from  the activation function selection strategy\cite{mehta2021modulated}, using ReLU, which has more generalization performance, as the activation function of the shared layer and sine, which has more representation capability, as the activation function of the private layers.
On top of this, we improve INR by introducing a hierarchical parameter sharing mechanism in terms of data compression.

\vspace{2mm}
\noindent\textbf{Parameter sharing in INR.}
To improve the representation accuracy of INR for large and complex scenes, many algorithms use the parameter sharing strategy. One implementation is introducing convolution operations\cite{chen2021nerv,peng2020convolutional,boulch2022poco}, which allows hierarchical extraction of local features, reduces the required parameters, and enables describing high resolution data. However, for those complex data with uneven distribution of features, the spatial invariance of convolution would lead to lower local representation accuracy. 
Another parameter sharing mechanism is positional encoding\cite{nerf,ffn}, in which nearby spatial positions share partial mapping features hierarchically. However, this can only achieve higher representation accuracy when equipped with random or learnable mappings\cite{ffn,siren,benbarka2021seeing}, which can be considered as an embedding that encodes only spatial distance.  In addition, using embedding\cite{acorn,takikawa2021neural,ingp} intrinsically reduces the computational complexity at expenses of increased parameters, and thus not optimal for compression.

To make use of advantageous and bypass the above issues, we borrow the idea of hierarchical parameter sharing into local INRs, and build our compressor on a widely used tree structured data partitioning\cite{yang2022sci,acorn,takikawa2021neural}. 
One the one hand, the parameter sharing is applied among block-specific INRs to cope with the large data with high local fidelity. On the other hand, we share parameters on INR rather than the data itself or its embedding, to encode richer information other than spatial distance. 

\section{Tree-structured implicit neural compression}
%We now present our proposed hierarchical sharing mechanism built on ensembled INRs, and tree structured network design. 
%. We first revisit implicit neural compression and then analyze its usage under ensemble learning. We will derive the hierarchical parameter sharing mechanisms in turn and finally conclude our tree-structured implicit neural compression. 

\subsection{Ensemble of implicit neural compressors}
Implicit neural compression uses a compact neural network to parameterize a function describing the target data defined over a spatial coordinates system. More specifically, for an $N$ dimensional data, let $\mathbb{V}:=[-1,1]^N$ denote its grid coordinates and $\mathbf{d}(\mathbf{v})$ with $\mathbf{v}\in\mathbb{V}$ denote the intensity value at coordinate $\mathbf{v}$, one can learn a compact implicit function $f$ with parameter $\mathbf{\Theta}$ to estimate $\mathbf{d}(\mathbf{v})$ by $f(\mathbf{v};\mathbf{\Theta})$ and finally serialize them to achieve a compressed representation. Mathematically, we can formulate the compression task as an optimization problem
\begin{align}
    \min\limits_{\mathbf{\Theta}}\quad\int\limits_{\mathbf{v}\in \mathbb{V}}\mathcal{L}(f(\mathbf{v};\mathbf{\Theta}),\mathbf{d}(\mathbf{v})),
    \label{eq:network optimization objective}
\end{align}
where $\mathcal{L}$ measures the difference between the decompressed and target data.
For decompression, one can retrieve $f$ and $\mathbf{\Theta}$ from the compressed file and then calculat function values $f(\mathbf{v};\mathbf{\Theta})$ over grid $\mathbb{V}$ via simple forward propagation.

Due to limited spectrum coverage, INR is incapable of compactly describe large and complex data \cite{mehta2021modulated,yang2022sci}, we borrow the idea of ensemble learning to partition the target volume into blocks and use multiple less expressive $f_k(\cdot,\mathbf{\Theta}_k)$ to achieve a powerful representation. 
Specifically, we adopt the divide-and-conquer strategy as in \cite{mehta2021modulated,kadarvish2021ensemble,yang2022sci} to ensemble all implicit functions $\{f_k, k = 1,\cdots,K\}$ that represents data at its corresponding coordinate regions $\{\mathbb{V}_k, k = 1,\cdots,K\}$ and compose an $f$, which is equivalent to approximating a complex function by a piece-wise function composed of $K$ simple functions
\begin{align}
    f(\mathbf{v};\mathbf{\Theta}):=&\sum\limits_{k=1}^{K}\mathbf{1}_{\mathbb{V}_k}(\mathbf{v},f_k(\mathbf{v},\mathbf{\Theta}_k)),\\
    \mathbf{1}_{\mathbb{V}_k}(\mathbf{v},x)=&\left\{
    \begin{aligned}
    &x,\mathbf{v}\in\mathbb{V}_k\\
    &0,else
    \end{aligned}
    \right..
    \label{eq:ensemble learning}
\end{align}
In addition, we must constrain the total number of parameters to be consistent with the parameter budget, i.e., $|\mathbf{\Theta}|=\sum\limits_{k=1}^{K}|\mathbf{\Theta}_k|$, where $|\cdot|$ is the support operation. Here we omit the parameters for serializing $f_k$ which consists of only a few structural hyperparameters of neural network.

For a target data volume shown in the first column of Fig.~\ref{fig: method}, we use an $L$ level tree structure containing $K$ leaf nodes to organize $f_k$ and $\mathbb{V}_k$, as illustrated in the 2nd column of Fig.~\ref{fig: method}. The root node represents the whole coordinate region $\mathbb{V}$ and each child node represents $\frac{1}{2^N}$ region of its parent node. We use $f_k$ to compress the data in coordinate region $\mathbb{V}_k$ represented by the $k$th leaf node. This structure allows us to organize ensemble learning in high degree of flexibility. % and fits well with the parameter sharing mechanism proposed below.

\subsection{Hierarchical parameter sharing mechanism}
\label{sec:hierarchical parameter sharing mechanism}
We let these $\{f_k\}$ share their neural network parameters hierarchically with each other according to the spatial distance between corresponding regions $\mathbb{V}_k$. The sharing mechanism is defined on the octree structure. Specifically, for a leaf node at level $l$, its corresponding MLP-implemented $f_k$'s hidden layers can be divided into $l$ segments, i.e. $f_k=f_k^{out}\circ f_k^l\circ f_k^{l-1}\circ\cdots\circ f_k^{1}\circ f_k^{in}$. For a set of leaf nodes, we determine the number of shared segments based on the number of their common ancestor nodes. For example, if $f_i$ and $f_j$ share the same ancestor nodes at $1\sim3$ levels, three pairs of hidden layer segments $(f_i^1,f_j^1),(f_i^2,f_j^2),(f_i^3,f_j^3)$ will share the same parameters. 

The sharing mechanism is illustrated in the 3rd column in Fig.~\ref{fig: method}. Since the distance between two leaf nodes can also be inferred from the number of their shared ancestor nodes, above % distributing the shared parameters across the tree nodes is equivalent to a
hierarchical parameter sharing mechanism implies that closer blocks share more parameters and are of higher similarity in the implicit representation space.

%The detailed implementation of hierarchical parameter sharing mechanism in our approach \M~will be described in \cref{sec:TINC}

\subsection{Tree-structured network architecture}
\label{sec:TINC}
We propose a tree-structured MLP based on the $L$ level octree partitioning, as illustrated in the 4th column in Fig.~\ref{fig: method}. Each node contains a hyper layer consisting of some fully connected layers and takes the output of its parent node's hyper layer as input. Root node and leaf nodes additionally contain the input and output layers respectively.

Following the depth direction, from leaf nodes upward to the root of the tree, 
%We can first analyze this tree-structured MLP in terms of the information flow in the depth direction. 
the output information of the leaf node is processed by the hyper layers in its ancestor nodes. If we consider the hyper layer as a part of hidden layers, then any leaf node can be considered as an MLP, whose hidden layers are explicitly divided into multiple segments contained in its ancestor nodes across different levels. In this way we can ensemble all $\{f_k\}$ into a tree-structured MLP.

%Then we can analyze from the width direction.
At the same level, all sibling nodes share the same parent node and thus take the same information as input. Therefore, for any two leaf nodes sharing the same ancestor nodes, their output information have flown through the same hyper layers contained in the ancestor nodes. In other words, %If we adopt the context in the previous paragraph, 
the MLPs of these two leaf nodes will share a part of hidden layers, which implements the hierarchical parameter sharing mechanism described in \cref{sec:hierarchical parameter sharing mechanism}.

\section{Cross level parameter allocation}
\label{sec:cross level param alloc}
After the network design, we are left with some key implementation issues: (i) how to set a proper number of levels to finalize the tree structured network, (ii) how to allocate available parameters across MLPs at the tree nodes. 

\vspace{2mm}
\noindent{\bf The number of tree levels $L$.\quad}
The number of tree levels $L$ determines the number of divided segments of $f_k$'s hidden layers, i.e., a larger $L$ implies the parameter sharing mechanism proceeds to higher spatial resolution. %Intuitively, we should use a larger $L$ for the data containing more more high-frequency information.
Setting a large $L$ has both advantageous and disadvantages. On the one hand, a larger $L$ means finer partitioning with a larger number of blocks sharing parameters (with the same hyper layers as in the ancestor nodes), which is beneficial for describing the regions with rich details and thus improve the compression fidelity. On the other hand, with the same total number of parameters, a larger $L$ will lead to more leaf nodes $K$ and less parameters $|\mathbf{\Theta}_k|$ each $f_k$, which might reduce $f_k$'s representation capability and might harm the compression fidelity in turn.

Therefore, we should set the tree level $L$ by trading off between increased fineness in partitioning and parameter allocation, and $f_k$ decreased representation capability.  In \cref{sec: The number of tree levels exp}, we experimentally prove that it is favourable using large tree levels for the data with rich details or under low compression ratios.

\vspace{2mm}
\noindent{\bf Inter-level parameter allocation.\quad}
%Different inter-level parameter allocation strategies will also have an effect on \M. Here we only do a preliminary exploration of this feature and and only explore when to allocate more parameters to the shallow or deep level. 
The parameters at the shallow level describe the shared features at large scale,  i.e., the non-local redundancy distributed over distant regions.  
%the hierarchical parameter sharing mechanism will be carried out on a larger spatial scale, since the same ancestor nodes between nodes corresponding to spatially distant regions are located in the shallow level. 
Similarly, from the perspective of data features, if some far apart sub-regions at a certain level are highly similar, one can allocate more parameters to their common ancestor nodes, and vice versa. % then by moving a small number of parameters from deeper level to this corresponding level, these very similar regions will likely gain better compression fidelity because similar features are better modeled. 

Therefore, for data with high global redundancy or repetitiveness, i.e., high non-local similarity among sub-regions at large scales, allocating more parameters to the shallow level will be more beneficial to improve its compression fidelity. 
Conversely, for data with weak global similarity, it is more efficient to allocate more parameters to the nodes at deep levels for better representation of the unique local features.
%If there are many representations that can be shared among those $f_k$ that are spatially distant, then more parameter allocated on the shallow level will help improve the cost effectiveness of parameter usage, thus improving the overall compression fidelity.

%最后，至于如何根据我们算法的这一特性，去设计一套优美的参数分配方法，甚至是自适应的方法，我们将在未来工作中进一步探究。我们在后续试验中统一采用1。

\vspace{2mm}
\noindent{\bf Intra-level parameter allocation.}
In addition to allocating parameters across levels, we can also allocate more parameters within a level to some nodes according to certain criteria instead of even allocation. For example, the most valuable information in neuronal data is often distributed in sparser regions. If we allocate more parameters to the more important regions, then we can sacrifice the unimportant regions in exchange for the improvement of compression fidelity of the important ones. %Since \M~uses a tree-structured MLP, we can allocate the number of parameters within multiple levels, thus achieving a hierarchical distribution of compression fidelity across multiple spatial resolutions and ultimately maximizing the use of a limited number of parameters.
\begin{table*}[t]
\centering
\small
%\vspace{-2mm}
    \caption{The scores on each dataset as compression ratio ranges from 64 to 1024, with the \textcolor{red}{best} and the \textcolor{blue}{second best} marked by color. The metrics include PSNR(dB), SSIM, and Acc.$\tau$ with $\tau$ acts as a threshold for binarization. The ``All'' or ``High'' suffixes after the metrics name indicate that it is averaged for all compression rations or only high compression ratios (around 1024$\times$).
    }
    \setlength{\tabcolsep}{0.8mm}{
    {
    \begin{tabular}{c||cccc||cccc} 
    \hline
        &
        \multicolumn{4}{c||}{Medical data} & 
        \multicolumn{4}{c}{Biological data} \\
        \cline{2-9}
        Method 
        & {\scriptsize ~~~PSNR All. (dB)~~~} & {\scriptsize ~~~SSIM All.~~~}
        & {\scriptsize ~~~PSNR High. (dB)~~~} & {\scriptsize ~~~SSIM High.~~~}
        & {\scriptsize ~~~Acc.200 All.~~~} & {\scriptsize ~~~Acc.500 All.~~~} 
        & {\scriptsize ~~~Acc.200 High.~~~} & {\scriptsize ~~~Acc.500 High.~~~} 
        \\ 
    \hline
        \M~(ours) & \textcolor{blue}{52.02} & \textcolor{blue}{0.9897} & \textcolor{red}{50.59} & \textcolor{red}{0.9878} & \textcolor{blue}{0.9945} & \textcolor{blue}{0.9970} & \textcolor{red}{0.9934} & \textcolor{red}{0.9958} \\ \hline
        JPEG & 41.41 & 0.9722 & 30.49 & 0.9374 & 0.6612 & 0.9834 & 0.0197 & 0.9882 \\ \hline
        H.264 & 51.18 & 0.9896 & 47.28 & 0.9860 & 0.9919 & 0.9959 & 0.9860 & 0.9926 \\ \hline
        HEVC & \textcolor{red}{52.31} & \textcolor{red}{0.9903} & \textcolor{blue}{50.51} & \textcolor{blue}{0.9877} & \textcolor{red}{0.9955} & \textcolor{red}{0.9975} & 0.9917 & 0.9930 \\ \hline
        SCI & 51.90 & 0.9894 & 50.39 & 0.9876 & 0.9943 & 0.9965 & \textcolor{blue}{0.9921} & \textcolor{blue}{0.9951} \\ \hline
        NeRF & 50.93 & 0.9875 & 49.66 & 0.9863 & 0.9935 & 0.9962 & 0.9903 & 0.9940 \\ \hline
        NeRV & 47.11 & 0.9859 & 40.11 & 0.9800 & 0.9815 & 0.9901 & 0.9732 & 0.9867 \\ \hline
        DVC & 47.39 & 0.9865 & 45.74 & 0.9840 & 0.9827 & 0.9900 & 0.9692 & 0.9789 \\ \hline
        SGA+BB & 46.56 & 0.9836 & 43.02 & 0.9808 & 0.8038 & 0.9883 & 0.4817 & 0.9798 \\ \hline
        SSF & 46.25 & 0.9807 & 43.70 & 0.9773 & 0.7221 & 0.9603 & 0.7790 & 0.9542 \\
    \hline
    \end{tabular}}
    }
    \label{Table:1}
\end{table*}

\begin{figure*}[t]
	\centering
 \begin{subfigure}[b]{0.24\textwidth}{
 \centering
\includegraphics[width=\textwidth]{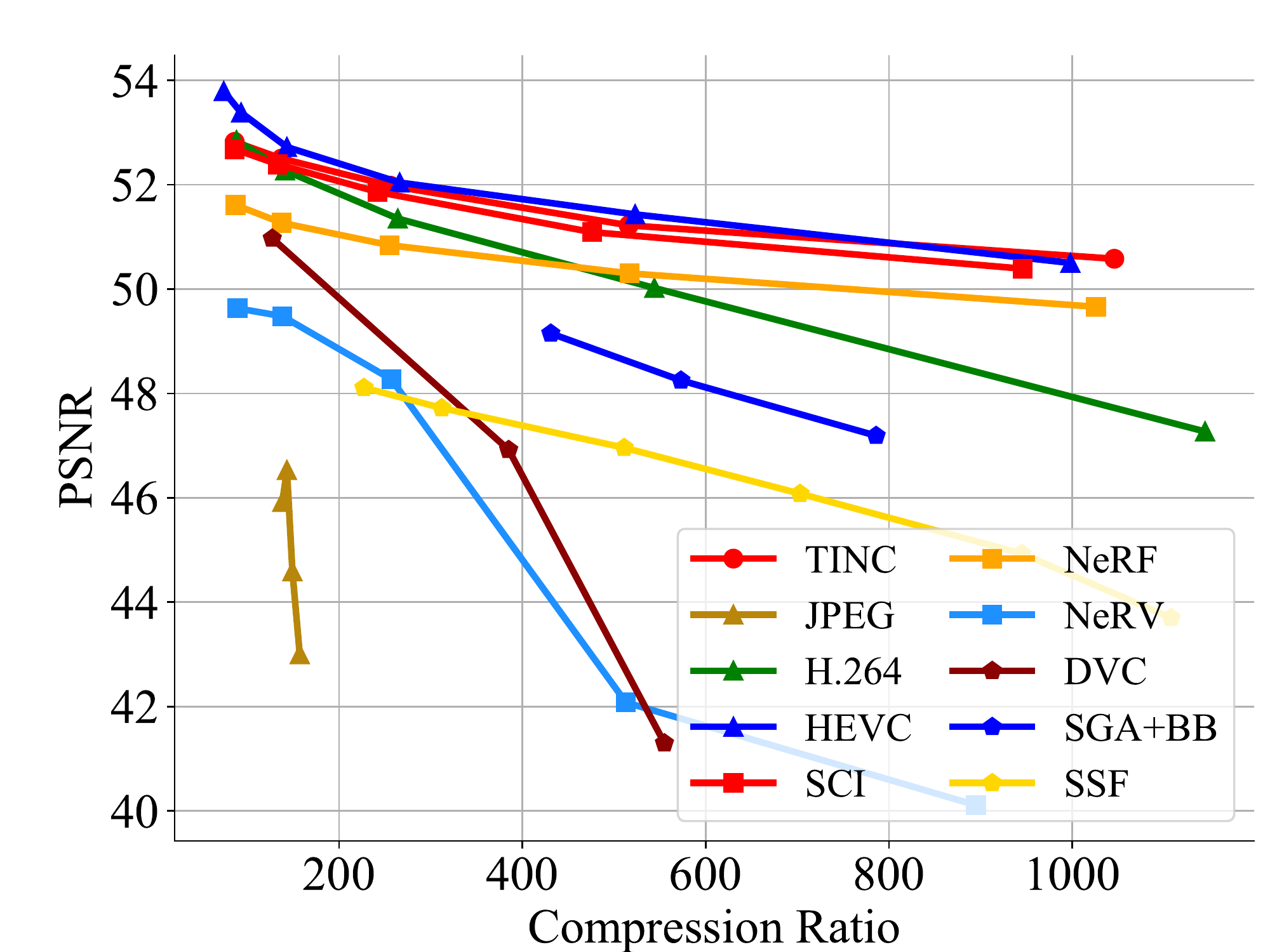}
\caption{}
}
\end{subfigure}
\begin{subfigure}[b]{0.24\textwidth}{
 \centering
\includegraphics[width=\textwidth]{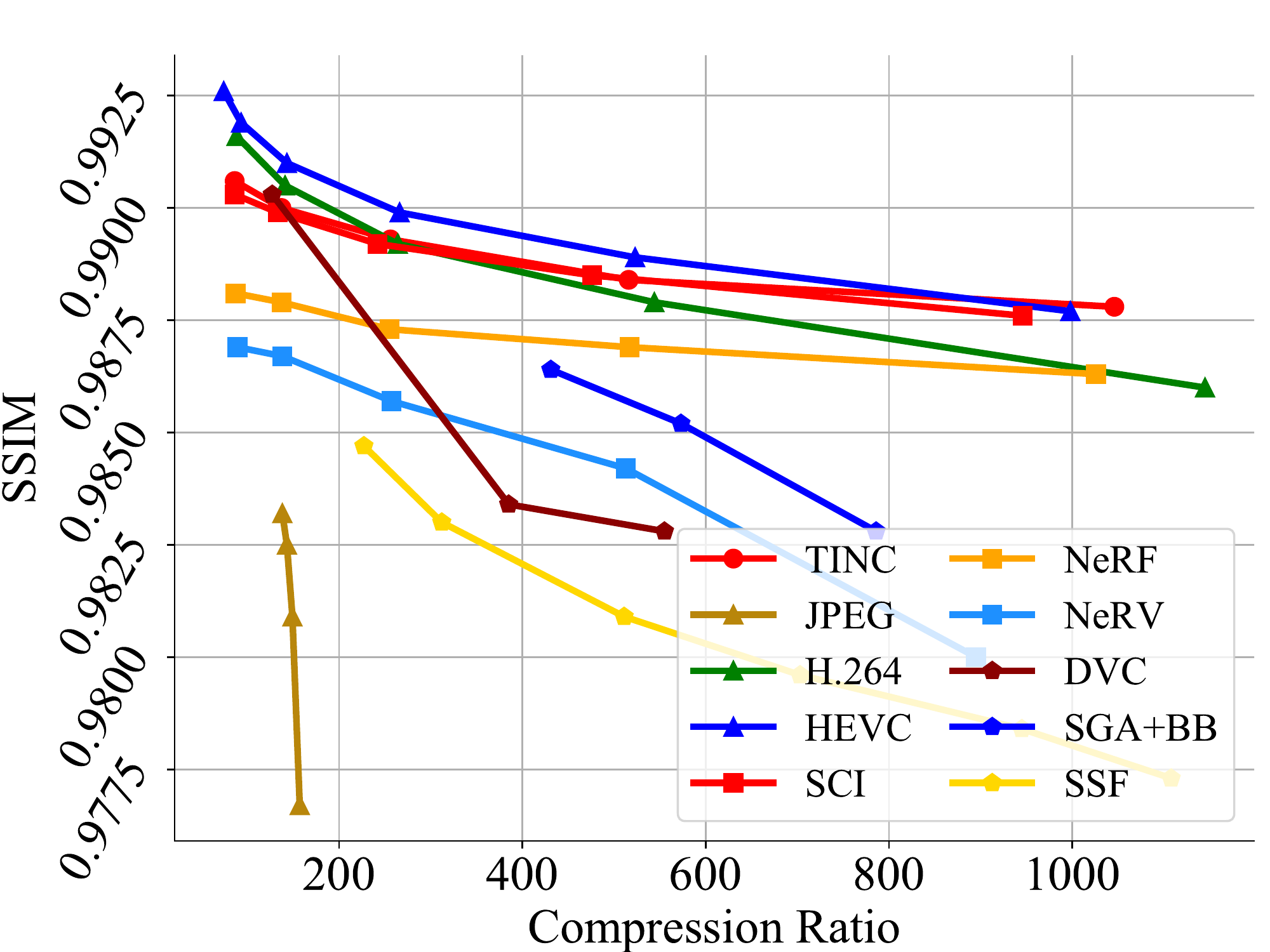}
\caption{}
}
\end{subfigure}
\begin{subfigure}[b]{0.24\textwidth}{
 \centering
\includegraphics[width=\textwidth]{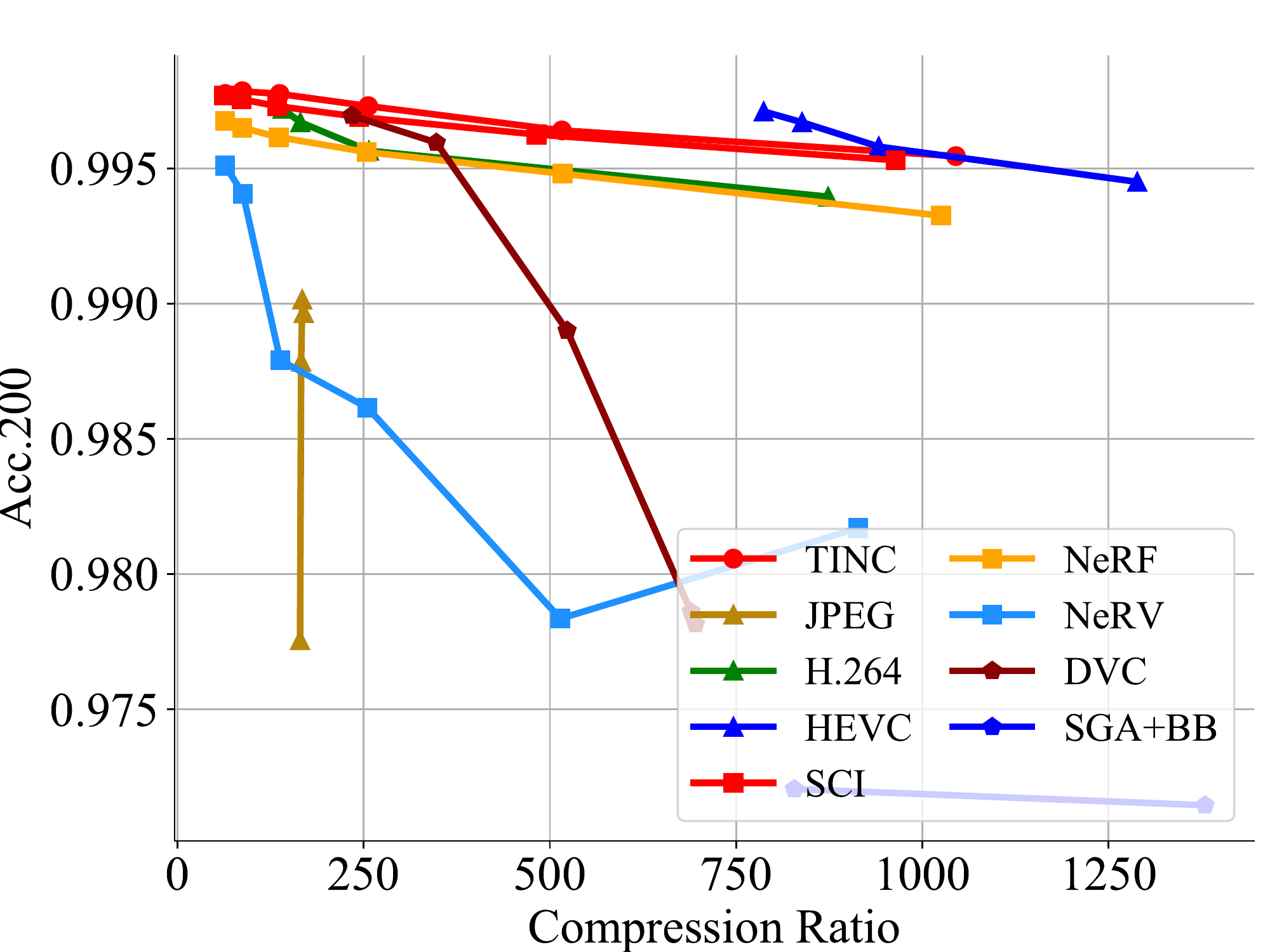}
\caption{}
}
\end{subfigure}
\begin{subfigure}[b]{0.24\textwidth}{
 \centering
\includegraphics[width=\textwidth]{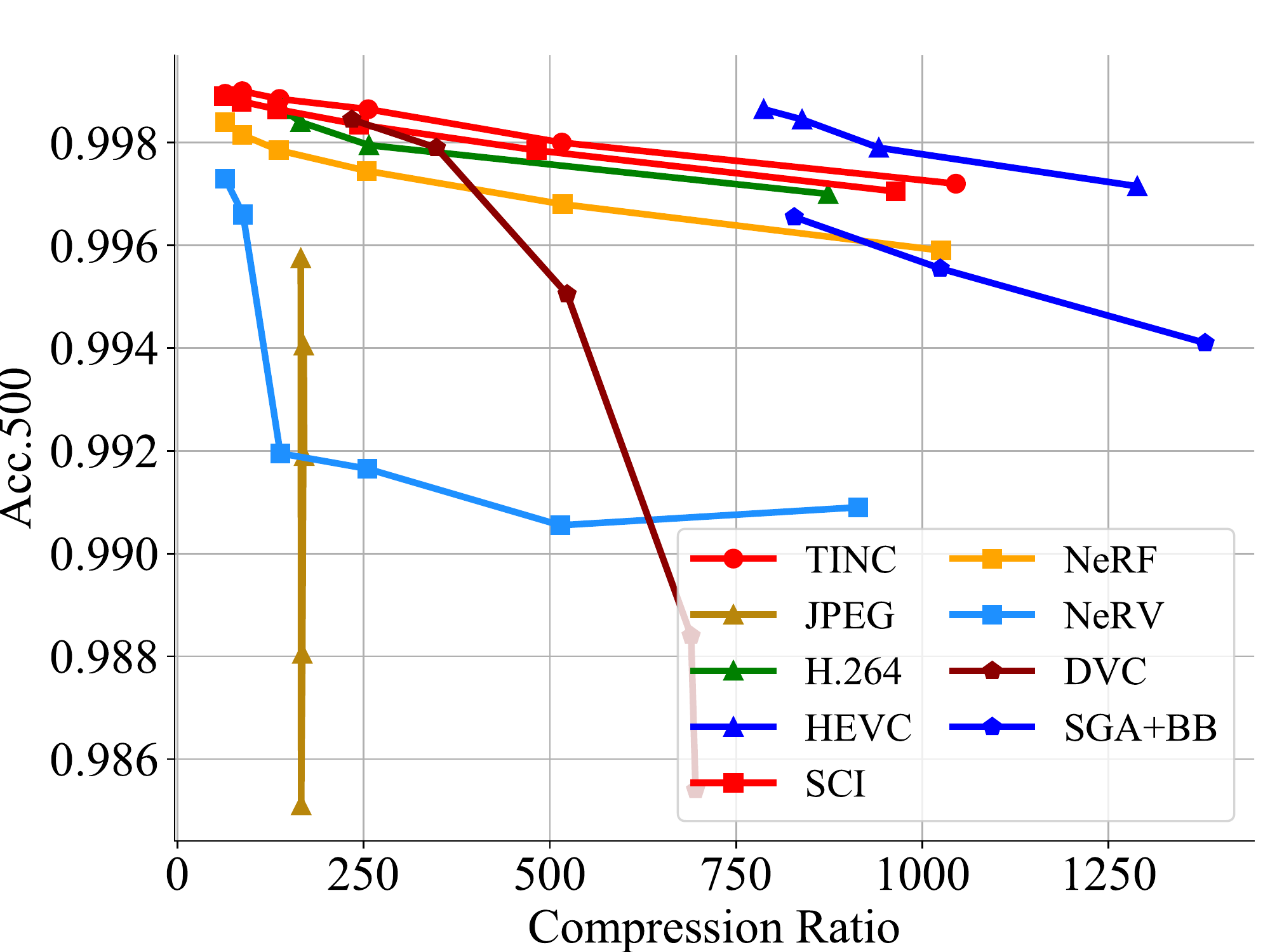}
\caption{}
}
\end{subfigure}
	\caption{Comparison of different compression methods on medical (a,b) and biological data (c,d) at different compression ratios.}
	\vspace{-2mm}
	\label{fig:Comparison}
\end{figure*}

A simple way to take advantage of the \M's hierarchy and flexibility is to allocate the number of parameters per node at each level according to the importance of the corresponding regions. Following this allocation, at each spatial resolution, the nodes corresponding to more important regions will have more parameters compared to other nodes in the same level. We demonstrate in \cref{sec: Intra-level parameter allocation exp} that this simple approach can significantly improve the compression fidelity of biological data, especially in regions with very low compression fidelity in the even allocation case, if an appropriate measure of the importance of the data region is adopted. Obviously there are more complex and efficient allocation methods, but this preliminary exploration is sufficient to demonstrate the advantages of \M.
%最后，至于如何根据我们算法的这一特性，去设计一套优美的参数分配方法，甚至是自适应的方法，我们将在未来工作中进一步探究。

\section{Experiments}
\vspace{2mm}
\subsection{Implementation details}
All the hyper layers in each node are implemented using MLP, where the activation functions are all sine function and the parameters are initialized following the scheme proposed in \cite{siren}. All MLPs are optimized by Adamax\cite{kingma2014adam} with initiate learning rate of 0.001 and decreased by a factor of 0.2 at the 2000th and 5000th iterations, respectively. We adopted a complete octtree, which means $K=8^{(L-1)}$. Before optimization, all the coordinates and the intensities are normalized to $[-1,1]$ and $[0,100]$ respectively. After optimization, all the parameters of each hyper layers are serialized and stored in binary. There are also some information that need to be saved for data decompression, such as MLP structure, original data bit depth, image size, inverse normalization basic information. Because all of the items are discrete, countable, and have a limited set of possible values, we can easily create a table listing all potential values and store their indices, to reduce the file size.

We evaluated our approach on both massive medical and biological data. For the former we used HiP-CT dataset\cite{walsh2021imaging},  which provides cellular level imaging of several organisms at multiple anatomical levels. We used four human organs (\textit{Lung}, \textit{Heart}, \textit{Kidney} and \textit{Brain}) from this dataset. For the latter we used the sub-micrometer resolution mouse whole-brain microscopic data (\textit{Neurons} and \textit{Vessels}) captured by our self-developed imaging system, which has varying brightness and structures across different regions.

For convenient and fair comparison, we cropped the data into the same size ($256\times256\times256$ voxels for medical data and $64\times512\times512$ voxels for biological data,).

\subsection{Performance comparison with state-of-the-arts}

\noindent{\bf Benchmark methods and evaluation metrics.\quad}
The proposed method is comprehensively compared with state-of-the-art methods, which can be classified into three groups: (i) commercial compression tools, including JPEG \cite{wallace1992jpeg}, H.264 \cite{wiegand2003overview}, and HEVC \cite{sullivan2012overview}; (ii) recently proposed data-driven deep learning based ones, including DVC \cite{lu2019dvc}, SGA+BB \cite{yang20improving}, and SSF\cite{agustsson2020scale}; (iii) rcently proposed INR based ones, including SCI \cite{yang2022sci}, NeRV \cite{chen2021nerv}, and NeRF \cite{nerf}. We evaluated all the methods at 6 different compression ratios ranging from 64$\times$ to 1024$\times$. It is worth noting that only INR based methods can precisely regulate the compression ratio. We adopted PSNR and SSIM for medical data including rich details, while Accuracy with threshold for binarization as evaluation metrics for biological data with sparse structures. For convenience we abbreviate Accuracy with threshold as ``Acc.threshold'' in the following descriptions.
\vspace{2mm}

% How do we use each method?
JPEG was tested on OpenCV. H.264 and HEVC were tested with their FFmpeg implementations.
For SGA+BB, we adopted the settings with the best performance mentioned in the original paper.
For DVC, we used its Pytorch implementation, and compressed the ``I Frames" using JPEG method with quality parameter set to 50.
For SSF, we used its CompressAI's implementation.
In addition, for the data-driven methods, i.e. DVC, SGA+BB, and SSF, we fine-tuned the provided pre-trained models on 300 volumes randomly selected from the dataset.
For SCI, we set the hyper parameter $a_{max}=32$.
% For SIREN, we set the hyper parameter $w_0=10$. 
For NeRF, the hyper parameter ``frequency" was set to 10.
For NeRV, we control the compression ratio by adjusting the number and width of the fully connected layers.
For 2D images compression methods, i.e. JPEG and SGA+BB, we firstly segmented the 3D data int 2D images before compression. 

\vspace{2mm}
\noindent{\bf Quantitative performance evaluation and analysis.\quad}
The average performance of each method under different compression ratios is plotted as a rate-distortion curve in Fig.~\ref{fig:Comparison}. We can see that our algorithm outperforms almost all methods at high compression ratios and is comparable to HEVC at other compression ratios.
For more detailed analysis, we summarize the scores on each dataset in Table~\ref{Table:1}. 
%Since only the INR based method can control the BPV precisely, the others fluctuate around the specified BPV, we calculate the average performance of all the algorithms within the intersection of their BPVs range on each dataset.
% 效果描述，待定
% Generally, our method and HEVC rank top two on all organs from CT. Since our method is a lossy compression and tends to suppress noise, on noisier data such as \textit{Lung} and \textit{Kidney}, our method is slightly inferior to HEVC on SSIM, while on cleaner data such as \textit{Heart} and \textit{Brain}, the PSNR of our method surpasses HEVC by 1.69dB and 0.51dB respectively. For \textit{Neurons} and \textit{Vessels}, the INR based methods have significant advantages and ours works best.
It should be noted that even if the PSNR of the medical data are higher than 40dB, it does not mean that there are no distortion in the decompressed data, which is different from the natural scene.

% \begin{figure*}[t]
%   \centering
%   \fbox{\rule{0pt}{2in} \rule{0.9\linewidth}{0pt}}
%   %\includegraphics[width=0.8\linewidth]{egfigure.eps}
%   \caption{}
%   \label{fig: benchmark}
% \end{figure*}

% From visualization comparisons what can we see?
Some visual comparisons of decompressed organ slices are presented in Fig.~\ref{fig: compare_details} (see Supplementary Figure.~\ref{figsup: compare_details_kidney} for more results). For better visual comparison, we applied the same post-processing to the decompressed data.
%For better visual comparison, we did the same processing on the decompressed data to improve contrast and clarity. 
Under compression ratio around $87\times$(since HEVC and SGA+BB cannot precisely regulate compression ratio), we consistently outperform existing INR based (SCI and NeRV), data-driven (SGA+BB) and commercial (HEVC) compression methods.
Although SCI achieved comparable results in the regions away from boundary, as shown in Fig.~\ref{fig: compare_details}(a), it produced severe blocking artifacts at the boundary, as shown by the arrows in (b), which will affect the medical research as well as other downstream tasks.
NeRV retained the general structures but suffered from the checkerboard artifacts caused by the aliasing in convolution. 
HEVC captured high fidelity details in ``I Frames" but suffered from the artifacts in ``P Frames", which distorted the edges in the volumes, such as the white matters in \textit{Brain} data and muscles in \textit{Heart} data, % and even change the original object,
which would be unusable for medical research and other downstream tasks.
SGA+BB struggled with retaining the rich details, since it cannot represent the similarity within the 3D volume as a 2D image compression algorithm.
In addition, HEVC produced ghosting artifacts and position offset in neighbor slices, since the optical flow algorithms that are widely applicable in video compression do not work well on such inherently spatially structured biomedical data. In contrast, our approach use INR to encode the intrinsic data structure.

\begin{figure}[t]
  \centering
   \includegraphics[width=\linewidth]{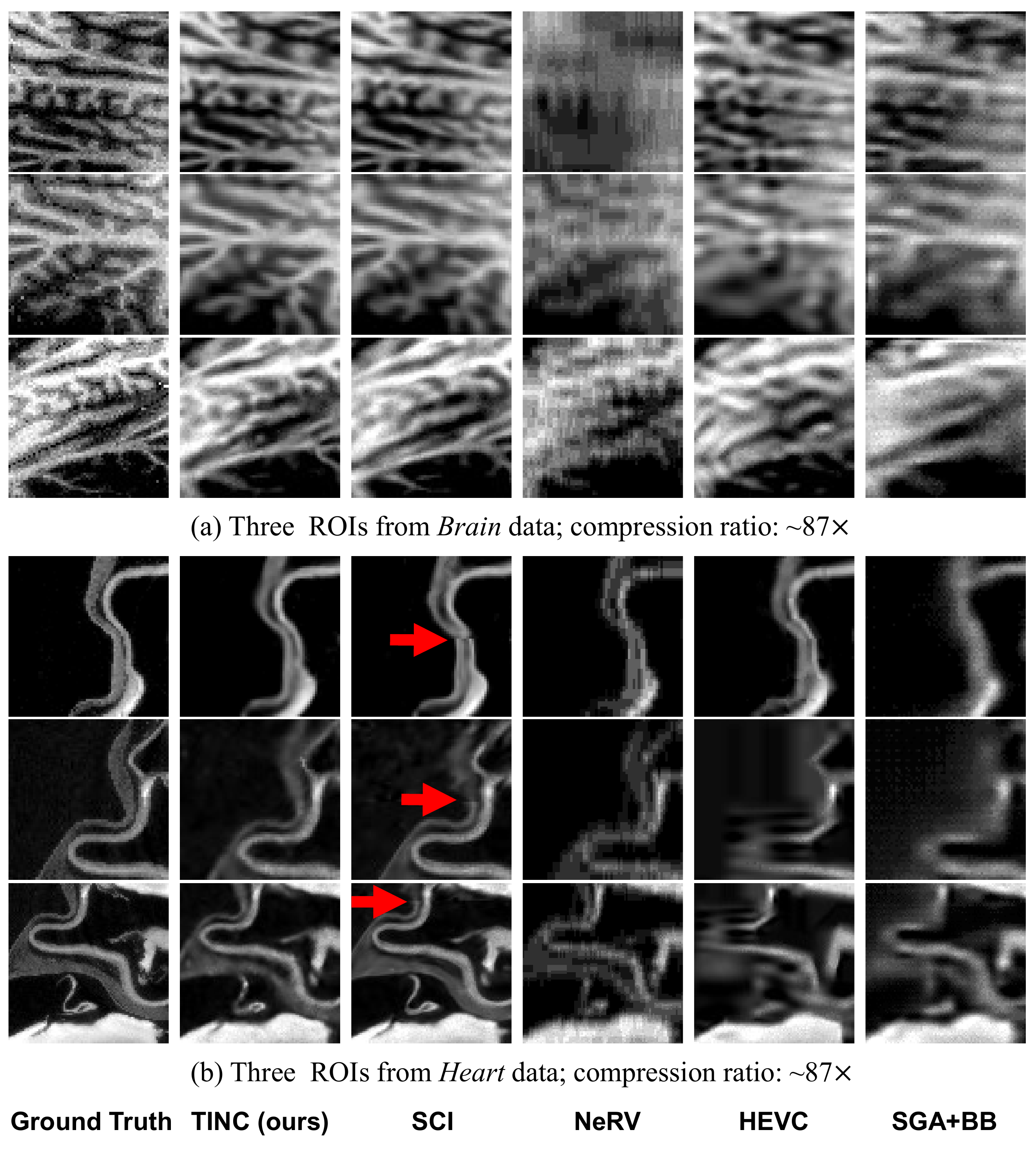}
   \caption{Visual comparisons of decompressed organs slices from \textit{Brain} and \textit{Heart} data between \M~and other benchmark methods under similar compression ratio around $87\times$.}
   \label{fig: compare_details}
\end{figure}

\vspace{2mm}
\noindent{\bf Comparison of running time.\quad}
We summarized the time required for compression and decompression for each algorithm in our experiments, as shown in the Supplementary Tables \ref{tab: speed on medical} and \ref{tab: speed on biological}. Comparatively, INR based compressor (\M, SCI, NeRV, and NeRF) is slower in compression stage but can decompress faster. It is worth noting that biomedical data need to be compressed only once before saving or transmission, but decompressed frequently for processing and analysis. Therefore, \M~is advantageous in biomedical fields.

\subsection{Flexibility settings for different data}
\begin{figure}[t]
%\vspace{-2mm}
\centering
 \begin{subfigure}[b]{0.22\textwidth}
     \centering
     \includegraphics[width=\textwidth]{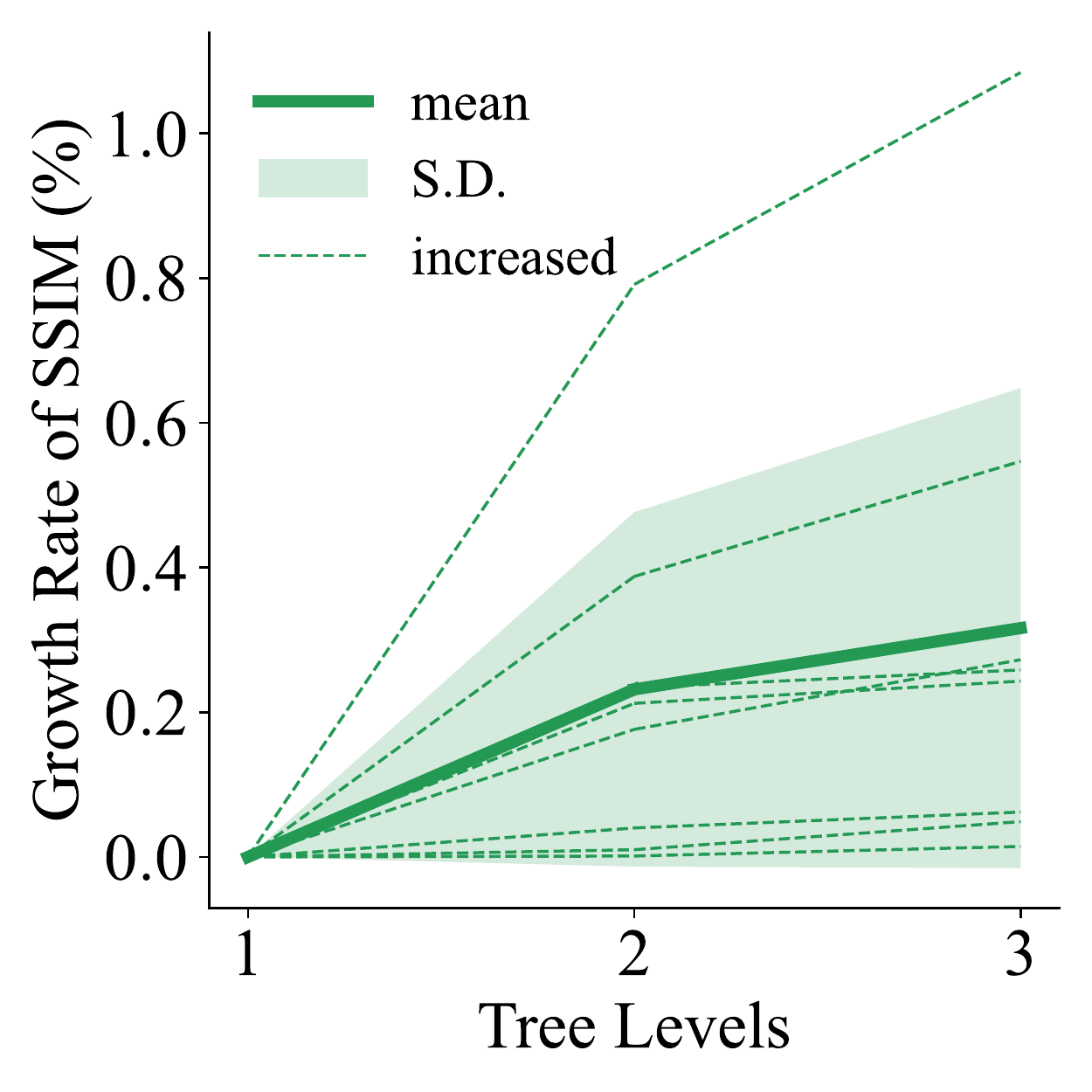}
     \caption{}
 \end{subfigure}
 \hfill
  \begin{subfigure}[b]{0.22\textwidth}
     \centering
     \includegraphics[width=\textwidth]{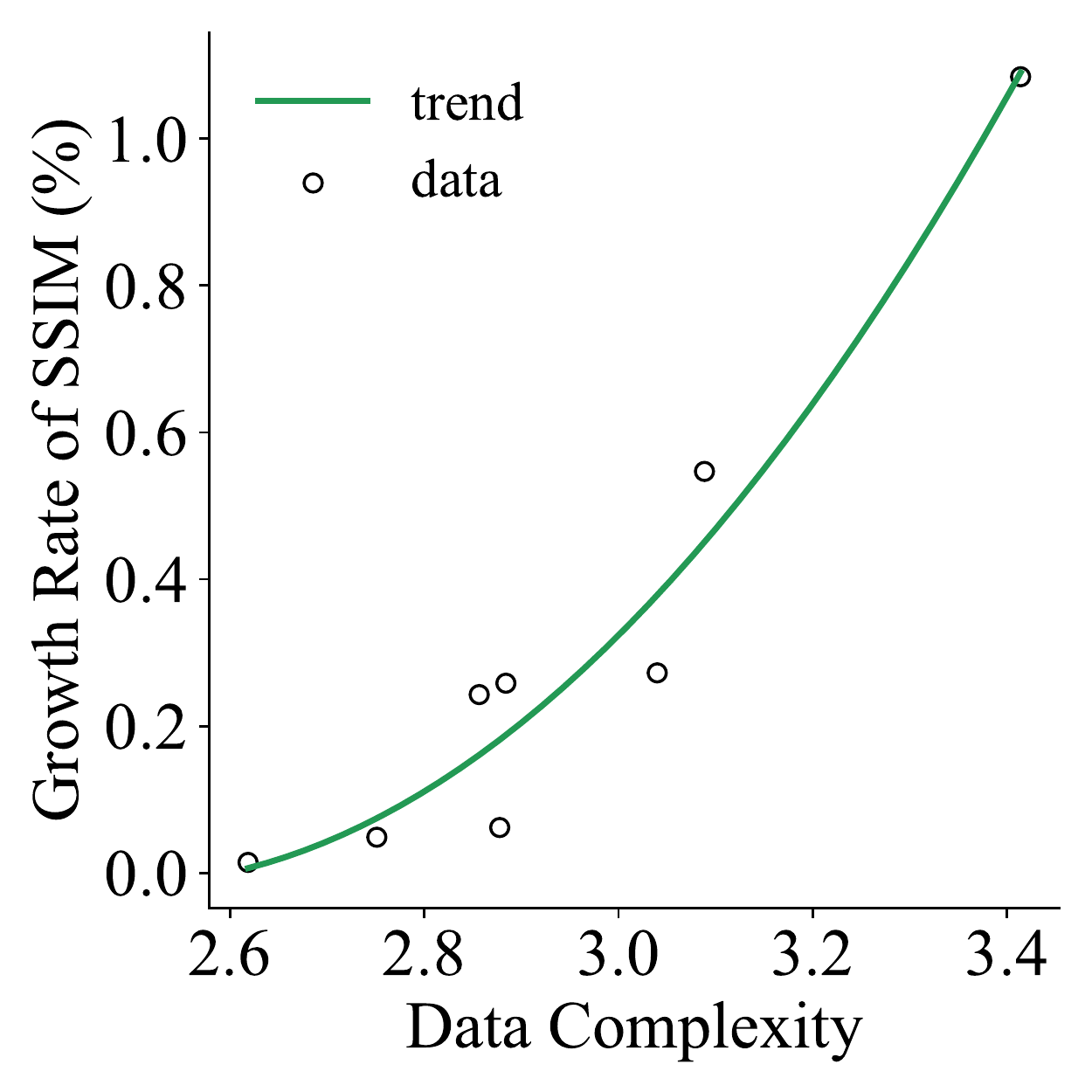}
     \caption{}
 \end{subfigure}
\caption{Effect of tree levels $L$ on \M's compression fidelity, in terms of SSIM on 8 medical data under 256$\times$ compression ratio.
(a) The growth rate of SSIM for each data when increasing the tree levels. The dashed lines represent the change in each data. The solid line represents the mean of changes, and the filled area represents the standard deviation.
(b) The scatter plot of each data's complexity and the growth rate of SSIM when increasing tree levels from 1 to 3. The solid line represents the trend of change, estimated by a quadratic fit.
}
\vspace{-2mm}
\label{fig:effect of tree layers}
\end{figure}

\noindent{\bf Setting of tree levels $L$.~~~~}
\label{sec: The number of tree levels exp}
We evaluated \M~at different tree levels $L$ on medical data under different compression ratios.
As shown in Fig.~\ref{fig:effect of tree layers}(a), the performance of \M~was evaluated in terms of SSIM on 8 medical data, under 256$\times$ compression ratio. To better demonstrate the effect of increasing tree levels, we used the growth rate of SSIM, calculated by normalizing all SSIM values by the one when $L$=1. 
The SSIM scores on all the test data have a steadily increasing growth rate (dashed lines) as the number of tree depth increases from 1 to 3.
Experimental results on other compression ratios show consistent trend (Supplementary Figure.~\ref{figsup: growth rate when increasing the tree levels}).

We further investigated the relationship between the data complexity and the  performance gain when increasing tree levels. 
For data complexity, we adopt the spectrum concentration defined in \cite{yang2022sci}, which measures the percentage of high frequency information contained in the data. 
The results are plotted in Fig.~\ref{fig:effect of tree layers}(b), with the solid line being a quadratic fitting of the growth rate for the 8 medical data using least squares. From the plot one can see that the performance gain increases monotonously with data complexity. In other words, the benefit of using a large tree is more significant when the target data contains more high-frequency information. This trend is consistent with our reasoning in \cref{sec:cross level param alloc} that a large $L$ favours parameter sharing at coarser levels and leave more parameters for representing the details. 
Experimental results on other compression ratios and other data are consistent with this (Supplementary Figure ~\ref{figsup:  scatter plot of each data's complexity and the growth rate}).

In addition, we studied the performance gain of using a deep tree structure under limited bandwidth budget $|\mathbf{\Theta}|$, e.g., at high compression ratio. 
Supplementary Figure.~\ref{figsup: growth rate when increasing tree levels on different compression ratios} plots the performance improvement at increasing compression ratios, calculated from 8 medical data volumes. Almost all data have a smaller growth rate of SSIM or PSNR when compression ratio increases, i.e. reducing the number of available parameters. At high compression ratio (512$\times$), some data are even of inferior fidelity using a deep tree, which is in line with our analysis in \cref{sec:cross level param alloc} that one should set a proper $L$ for choosing the right partitioning granularity under the constraint from available budget.

\vspace{2mm}
\noindent{\bf Inter-level parameter allocation.~~~~}
\label{sec: Inte-level parameter allocation exp}
We tested the performance at different parameter allocation proportions across levels on medical data under different compression ratios. 
Assuming there are $K^l$ nodes at $l$th level, we denote the number of allocated parameters at this level as $|\mathbf{\Theta}^l|$ and ratio with respect to that of its adjacent upper level as   $r^l=\frac{|\mathbf{\Theta}^l|/K^l}{|\mathbf{\Theta}^{l-1}|/K^{l-1}}$. 
Simply speaking, $r^l$=1 means even inter-level allocation, a $r^l$ larger than 1 means imposing higher priority to the shallow levels and vice versa. 
We let $r^l$ of all levels take 0.8,1 and 1.2 to represent the three typical cases of parameter allocation. %: more to to the shallow, even, and more to the deep, respectively.

\begin{figure}[t]
%\vspace{-2mm}
\centering
 \begin{subfigure}[b]{0.22\textwidth}
     \centering
     \includegraphics[width=\textwidth]{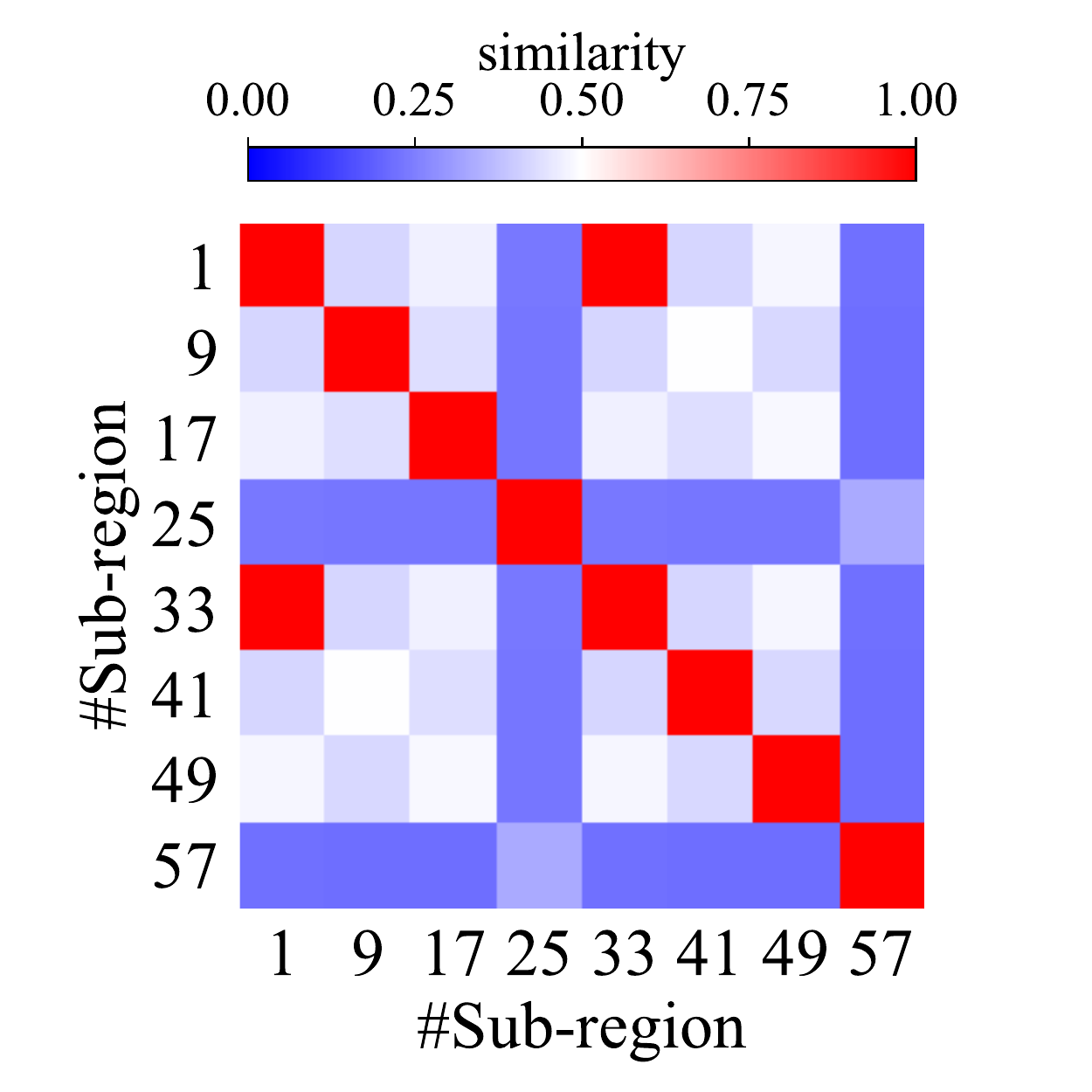}
     \caption{}
 \end{subfigure}
 \hfill
  \begin{subfigure}[b]{0.11\textwidth}
     \centering
     \includegraphics[width=\textwidth]{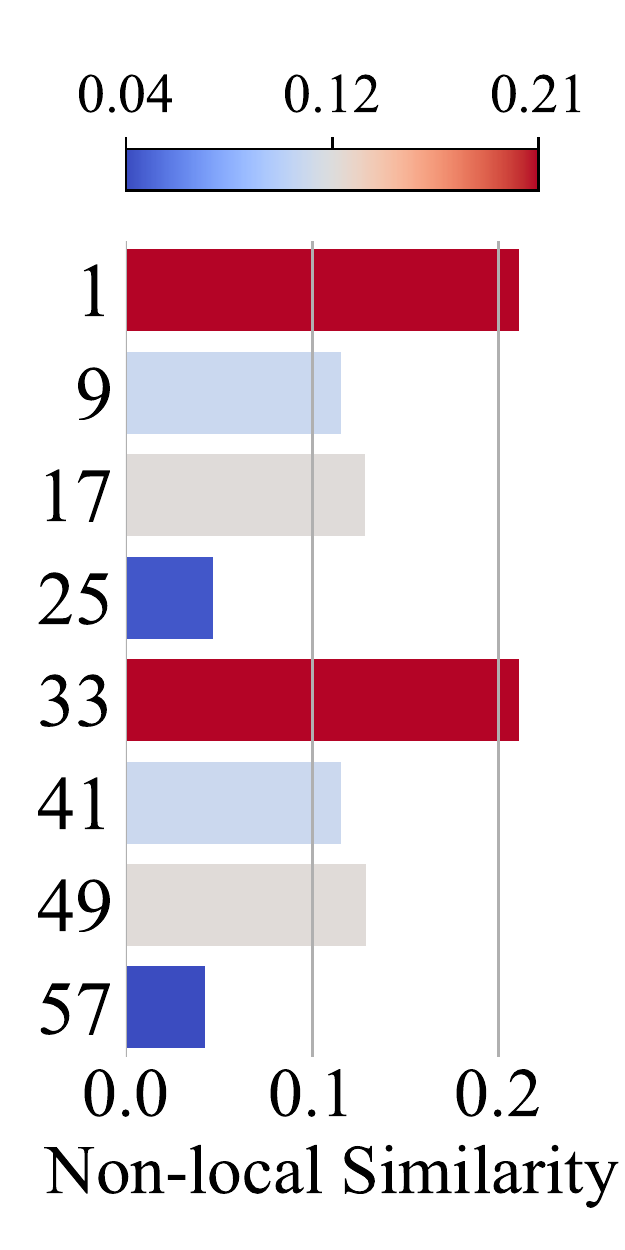}
     \caption{}
 \end{subfigure}
 \hfill
  \begin{subfigure}[b]{0.11\textwidth}
     \centering
     \includegraphics[width=\textwidth]{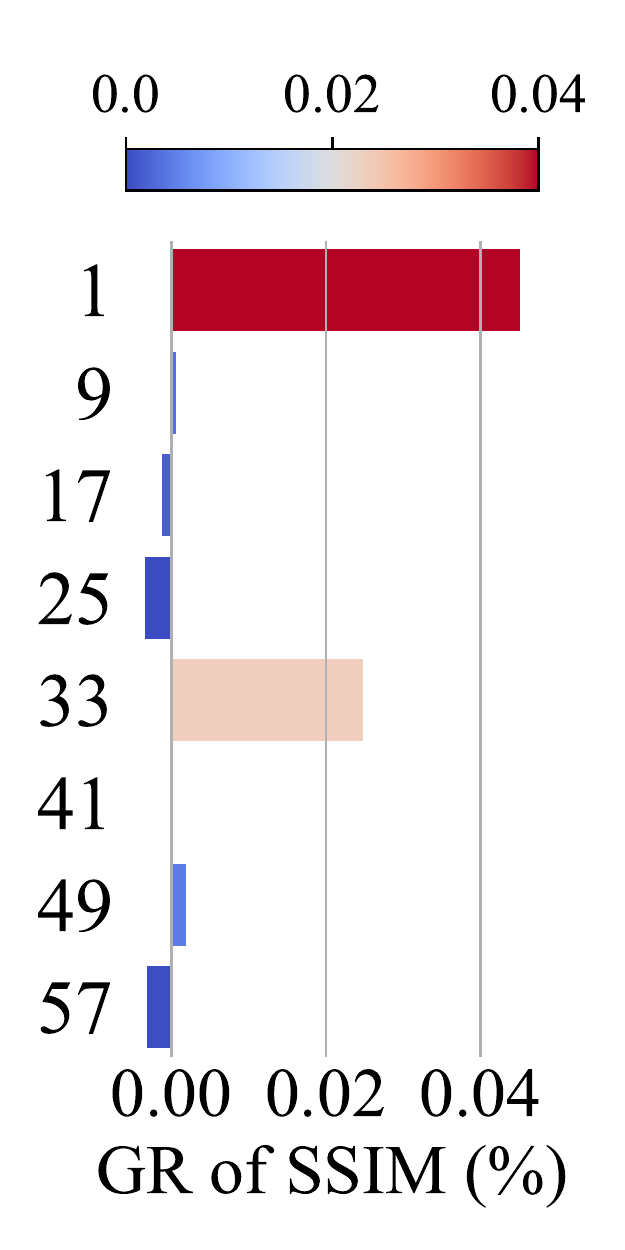}
     \caption{}
 \end{subfigure}
\caption{Effect of allocating more parameters to the shallow level on \M's compression fidelity for each sub-regions in a \textit{Brain} data under 512$\times$ compression ratio. All sub-figures share the same y-axis label.
(a) The heatmap of similarities between 8 equally spaced distant sub-regions. The serial numbers of the regions represent their z-curve order.
(b) The non-local similarity of each region.
(c) The growth rate of SSIM for each region when allocating more parameters to the shallow level.
}
\vspace{-2mm}
\label{fig:effect of tilting to shallow}
\end{figure}

We first investigated the effect of allocating more parameters to the shallow level, intending to share parameters among far apart similar blocks. As shown in Fig.~\ref{fig:effect of tilting to shallow}(a), we divided a \textit{Brain} data into 64 sub-regions, laid out in z-curve order, and selected 8 equally spaced distant regions. Then, we calculated the SSIM of these regions with respect to each other and used min-max normalization to obtain their non-local similarities. It can be seen that \#1 and \#33 regions are of high similarity. For convenient display, we also draw the bar plot of the normalized marginal sums of the similarity matrix in Fig.~\ref{fig:effect of tilting to shallow}(b), which tells that  \#1 and \#33 are both of high similarity to other blocks, while \#25 and \#57 regions are highly distinctive from others. 
%how similar a region is to the others, the similarities of each region to the others were summed up and converted to a ratio by Softmax function to facilitate comparison between regions, as shown in Fig.~\ref{fig:effect of tilting to shallow}(b).

To quantitatively measure the performance change at different blocks when allocating more parameters to the shallow level, we plot he growth rate of SSIM under $512\times$ ratio, as shown in Fig.~\ref{fig:effect of tilting to shallow}(c). Analyzing jointly with (b), we can find that the two regions with high non-local similarity, \#1 and \#33, gain a large boost, while those with the lowest similarity, \#25 and \#57, suffer a loss of fidelity. This is consistent with the intuition and  analysis in \cref{sec:cross level param alloc} that allocating more parameters to the nodes at shallow levels would benefit representing distant but similar regions, but at expense of slight degeneration at very individual regions.
Experimental results on other compression ratios and other data are shown in Supplementary Figures \ref{figsup: shallow level brain data heatmap} and \ref{figsup: shallow level heart data heatmap}.

We then investigated how to allocate parameters across levels according to the global consistency of the medical data. We divided the target data into 64 sub-regions, calculated the marginal sums of the similarity matrix. 
We change the setting of $r^l$ and the results are shown in the Supplementary Figure.~\ref{figsup: inter-level parameter allocation global consistency}. The results show allocating more parameters to the shallow level is more likely to improve fidelity when the global consistency of the data is greater than 0.7, and allocating more to the deep level is better when the global consistency lower than 0.6. Otherwise, an even allocation may be a better choice.

\vspace{2mm}
\noindent{\bf Intra-level parameter allocation.~~~~}
\label{sec: Intra-level parameter allocation exp}
We tested the approach of allocating parameters to the nodes at the same level according to their importance proposed in \cref{sec:cross level param alloc}, on biological data and under different compression ratios. 
Specifically,
% we denote the number of parameters allocated to the $i$th node at $l$th level as $|\mathbf{\Theta}^l_i|$, and the importance of corresponding data region as $I^l_i$. We let $|\frac{|\mathbf{\Theta}^l_i|}{|\mathbf{\Theta}^l_j|}=\frac{I^l_i}{I^l_j}$ for all levels
we let the number of allocated parameters be proportional to their importance.  %the number of parameters of every nodes in the same layer be proportional to their importance to represent the case of allocating more parameters to the more important regions, abbreviated as important allocation. And we let those be the same to represent the case of even allocation.
For example, with regard to the biological data such as \textit{Neurons} the importance can be measured by the percentage of the number of valid signals, where the valid signal can be determined either by the result of traced neurons or by the intensity value. Here we chose the latter for simplicity.

We first compare the compression fidelity of 6 \textit{Neurons} data using the above two parameter allocation approaches under different compression ratios, as shown in Supplementary Figure.~\ref{figsup: intra-level parameter allocation approaches for 6 Neurons data}. It can be seen that important allocation significantly improves the fidelity compared to even allocation.

\begin{figure}[t]
%\vspace{-2mm}
\centering
 \begin{subfigure}[b]{0.22\textwidth}
     \centering
     \includegraphics[width=\textwidth]{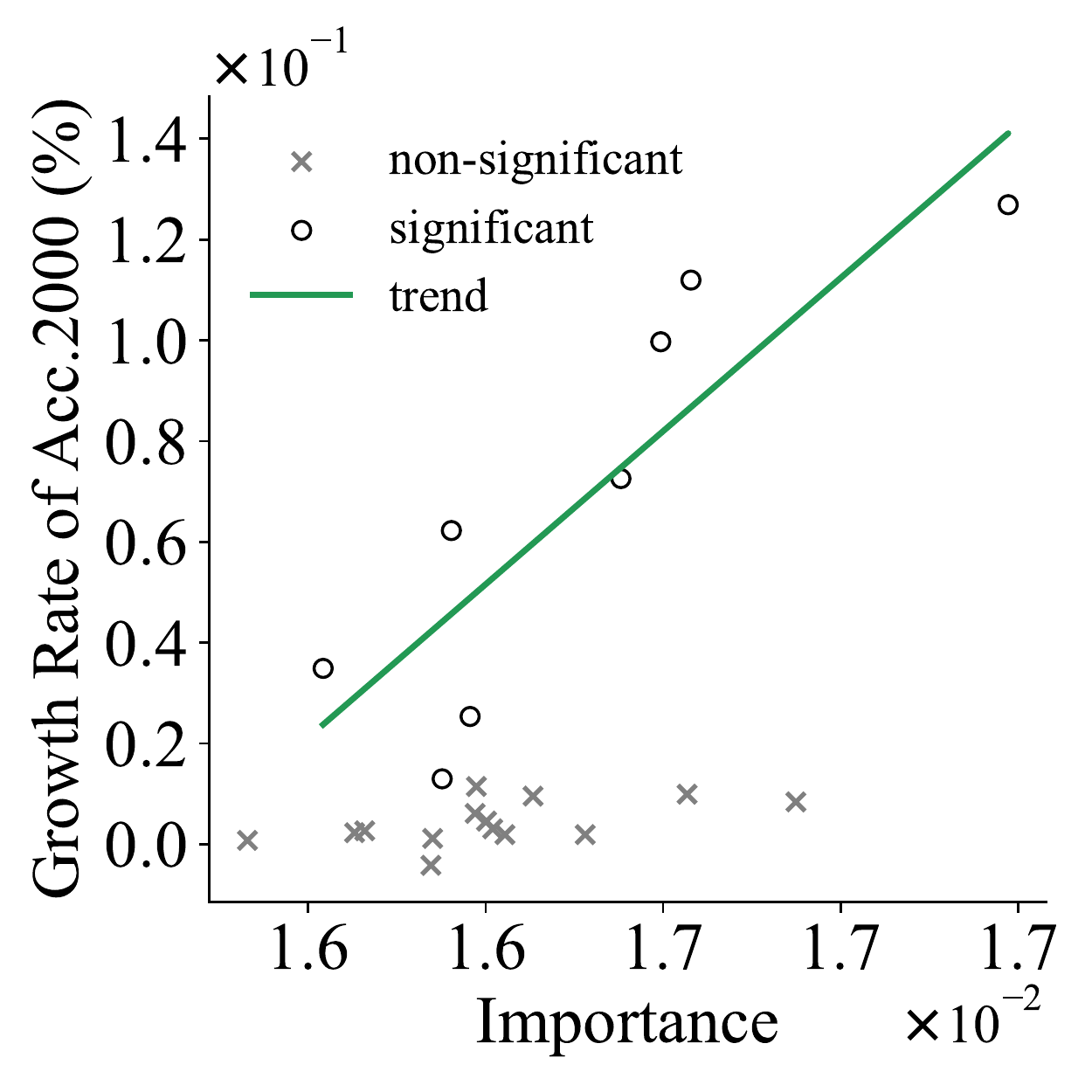}
     \caption{}
 \end{subfigure}
 \hfill
  \begin{subfigure}[b]{0.22\textwidth}
     \centering
     \includegraphics[width=\textwidth]{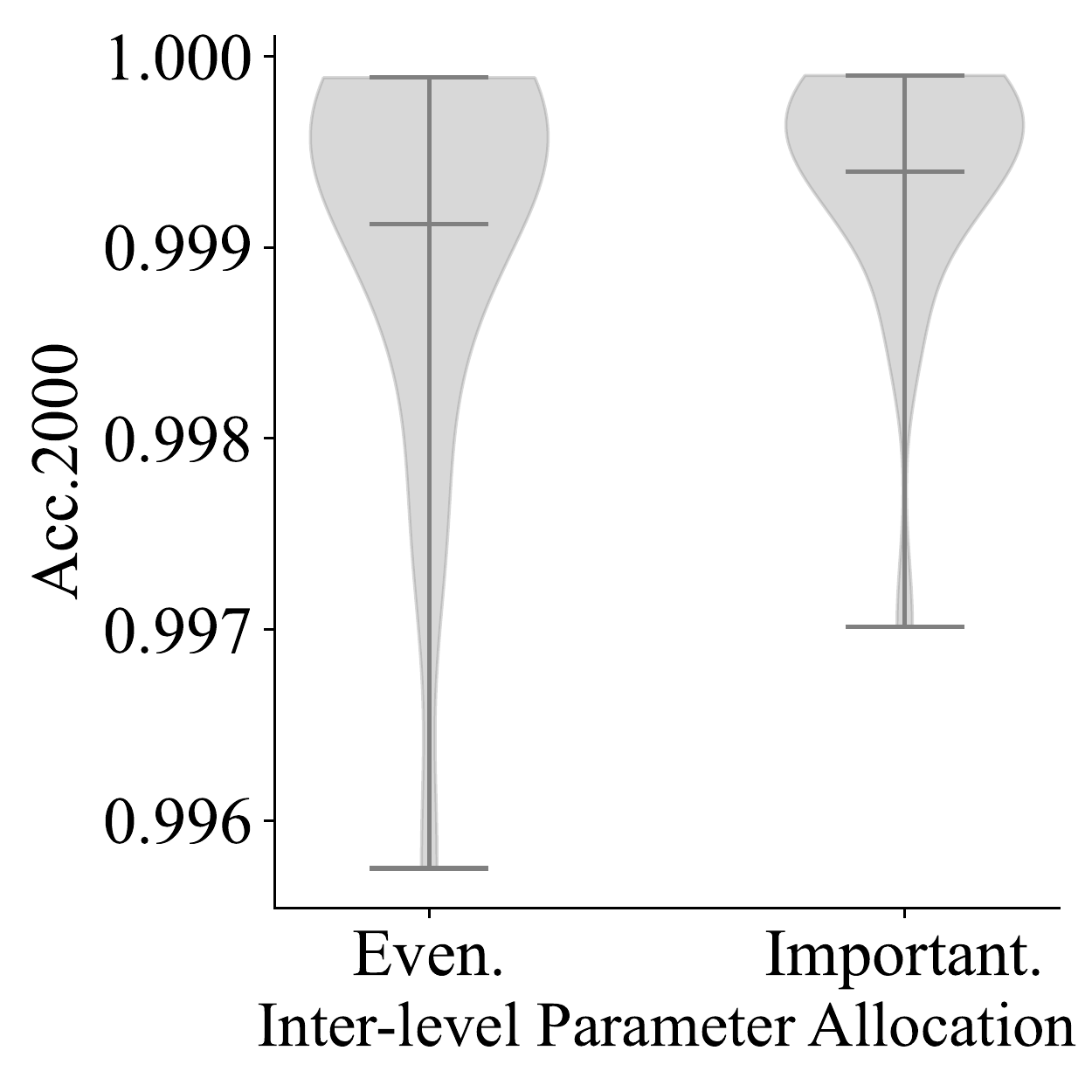}
     \caption{}
 \end{subfigure}
\caption{Effect of important allocation on \M's compression fidelity, in terms of Acc.2000 on a \textit{Neurons} data under 64$\times$ compression ratio.
(a) The growth rate of Acc.2000 for each region. The regions with and without significant growth rate are represented as cross and square respectively. The solid line represents the trend of change, estimated by a linear fit.
(b) The violin plots of Acc.2000 for all these regions. The shaded area represent the distribution of Acc.2000 across all regions. The three horizontal lines from top to bottom represent the maximum, average and minimum values respectively.
}
\vspace{-6mm}
\label{fig:effect of AOI}
\end{figure}

We then look further into the performance at all the sub-regions in the same target data.
As shown in Fig.~\ref{fig:effect of AOI}(a), we divided a \textit{Neurons} data into 64 sub-regions and compute their growth rate of Acc.2000 when switching from even allocation to important allocation. 
It should be noted that we omitted those regions that have reached 1 (the beast value) under both allocation approaches in the subsequent experiments.
From the scatter diagram we can see three groups of data: (i) Some regions show non-significant change (less than 10\% of the maximum change of the others, denoted as cross markers) using important allocation under 64$\times$ compression ratio, because they are already with sufficient parameters. (iii) Other regions show significant improvement after introducing important allocation, denoted with circles. We performed a linear fit using least squares, as shown by the solid line.
Overall, allocating parameters according to importance is helpful for the compression fidelity of important regions while at expenses of little fidelity degeneration in other less important ones.

For a statistical analysis, we plotted the Acc.2000 of all 64 sub-regions under even and important allocation, as shown in Fig.~\ref{fig:effect of AOI}(b). We found that importance based allocation can significantly improve those regions that have very poor compression fidelity under even allocation. This is mainly because that those regions with poor fidelity tend to contain more complex structures (with more neural signals), and thus are of higher importance scores and thus allocated more parameters accordingly. 
Experimental results on other compression ratios are detailed in Supplementary Figure.~\ref{figsup: important allocation under 512 compression ratio}.

%最后，至于如何根据我们算法的这一特性，去设计一套优美的参数分配方法，甚至是自适应的方法，我们将在未来工作中进一步探究。

\section{Conclusions}

\noindent\textbf{Summary and discussions.~~~~}
In this paper we propose a hierarchical parameter sharing mechanism for INR and implement it based on a tree-structured MLP for large and complex data compression tasks. The proposed approach \M~makes use of the local redundancy via block wise compression with separate MLPs and non-local redundancy via organizing these MLPs in tree structure to share parameters hierarchically. Such implementation 
forms a single network for the complex target data, not only incorporate the similarity distributed far apart and nonuniformly, and avoid discontinuity between adjacent regions as well.

Extensive experiments show that this hierarchical parameter sharing mechanism introduced in \M~greatly improves the compression capability and wide applicability of INR. Our method outperforms state-of-the-art compressors including commercial tools (JPEG, H.264, HEVC), data-driven methods (DVC, SSF, SGA+BB), and existing INR based ones (SCI, NeRV, and NeRF) on diverse medical and biological data.

We also analyze \M's flexibility to different cases via experimentally studying the effect of three key settings---number of tree levels, intra-level and inter-level parameter allocation methods, and provide highlights on proper parameter setting and strategy selection according to the features of target data. %on \M's compression fidelity on biomedical data with different characteristics under different compression ratios. And accordingly, the usage strategies of these features and their mechanisms are given for different experimental scenarios.

% Our method inherits the benefit of INR, such as the ability to precisely regulate the compression ratio and the freedom to tailor fidelity of different local regions.

\noindent\textbf{Limitations and future extensions.~~~~}
%速度；训练策略；其他更复杂的超参数配置
Similar to all current  INR based compression methods, \M~is of high decompression speed but slow in compression, since it takes time to pursue the MLPs matching the target data. We plan to combine meta-learning to find the best initialization parameters for each organ to speed up \M.

% We plan to combine meta-learning to speed up \M~ and optimize the training and sampling strategy for MLP on the tree structured network. 
%Although decompression is very fast, the drawback of all current INR based methods is the slow compression speed, 
% As an emerging technique, our approach is limited in several aspects. Firstly, current implementation requires relatively longer time than commercial compression methods, since it takes time to search for the proper network parameters fitting the target data.

The tree-shaped MLPs can be of different topological structure and brings a very large number of hyperparameter combinations. We would like to further investigate adaptive selection of tree structure to cover non-local redundancy across the data volume and hyperparameters to optimize the compression performance.

In addition, we plan to extend the parameter sharing mechanism to facilitate sharing network parameters among nearby regions in a feature space other than spatial space in \M. To find such a feature space for medical data will be one of our future extensions. 

We also would like to further develop \M~to be a general INR based compressor that can encode the intrinsic structure of arbitrary dimensional data, and apply it to other biomedical data such as 4D ultrasound imaging results, FMRI data, etc.
% to cover non-local redundancy across the data volume and hyperparameters to optimize the compression performance.

\section*{Acknowledgements}
We acknowledge MoE Key Laboratory of Biomedical Photonics (Huazhong University of Science and Technology) for sharing their mouse brain-wide microscopic data.

This work is supported by Beijing Natural Science Foundation (Grant No. Z200021), the National Natural Science Foundation of China (Grant Nos. 61931012, 62088102) and  Project of Medical Engineering Laboratory of Chinese PLA General Hospital (Grant No. 2022SYSZZKY21).
%%%%%%%%% REFERENCES
{\small
\bibliographystyle{ieee_fullname}
\bibliography{egbib}
}

%%%%%%%%%%%%%%%%%%%%%%%%%%%

\newpage
\onecolumn
\setcounter{figure}{0}
\captionsetup[figure]{name={Supplementary Figure.},labelsep=period}
\setcounter{table}{0}
\captionsetup[table]{name={Supplementary Table.},labelsep=period}

\begin{figure}[H]
\centering
\includegraphics[width=\linewidth]{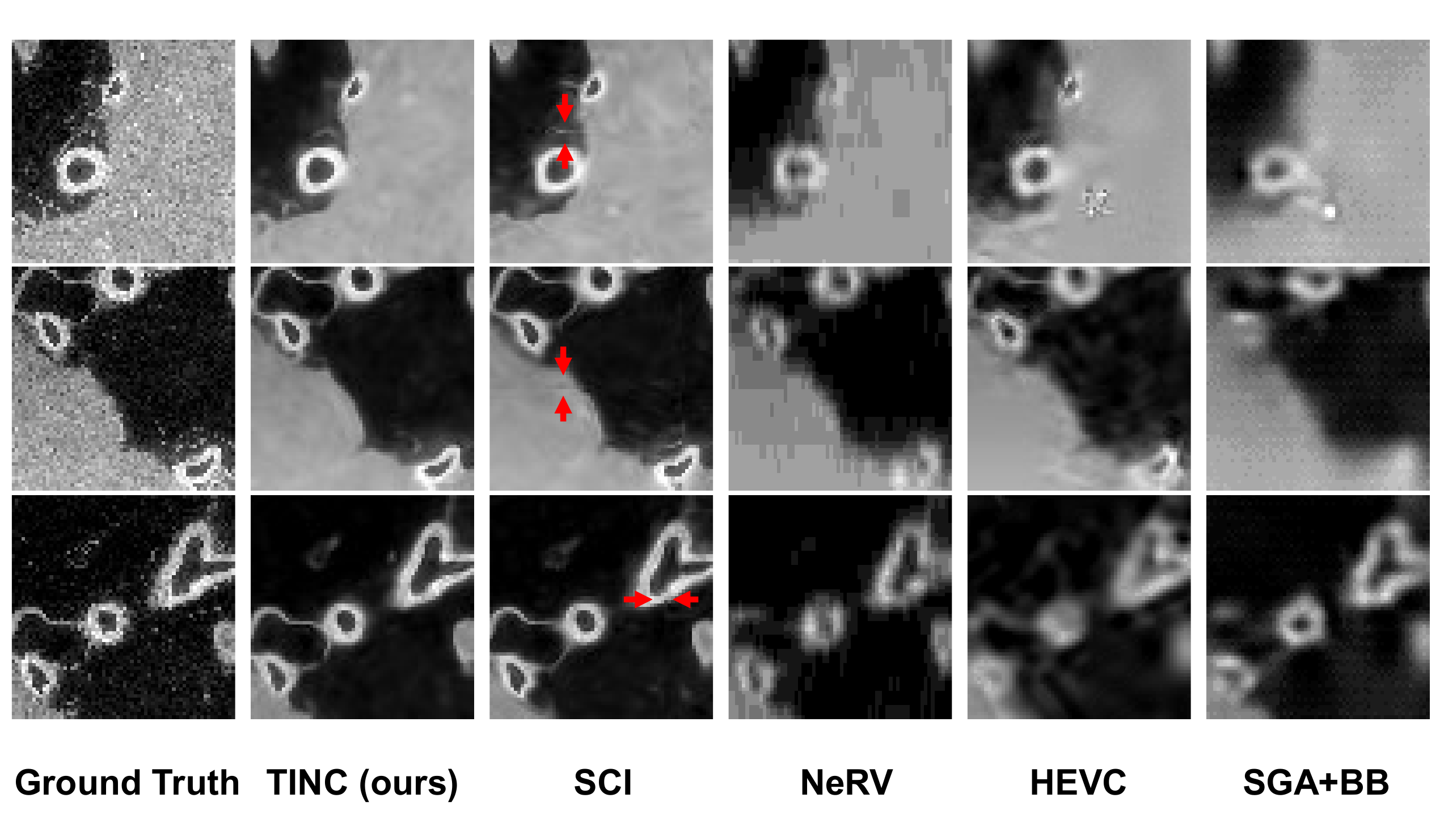}
\caption{Visual comparisons with state-of-the-arts on \textit{Kidney} data---2D slices from 3D volumes around 256$\times$ compression ratio. The red arrows highlight the blocking artifacts produced by SCI.
}
\label{figsup: compare_details_kidney}
\end{figure}

\begin{figure}[H]
\centering
 \begin{subfigure}[b]{0.3\textwidth}
     \centering
     \includegraphics[width=\textwidth]{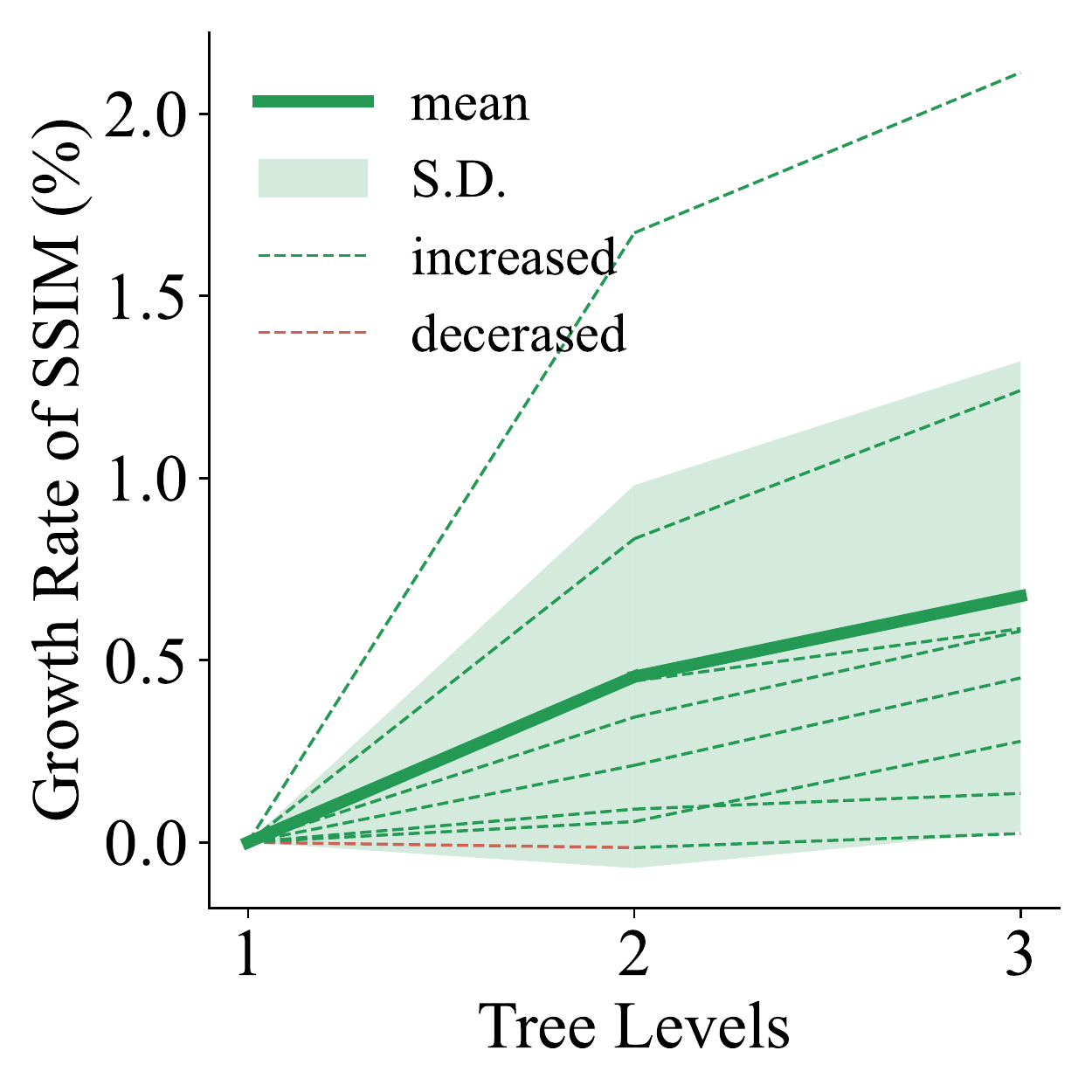}
     \caption{64$\times$}
 \end{subfigure}
  \begin{subfigure}[b]{0.3\textwidth}
     \centering
     \includegraphics[width=\textwidth]{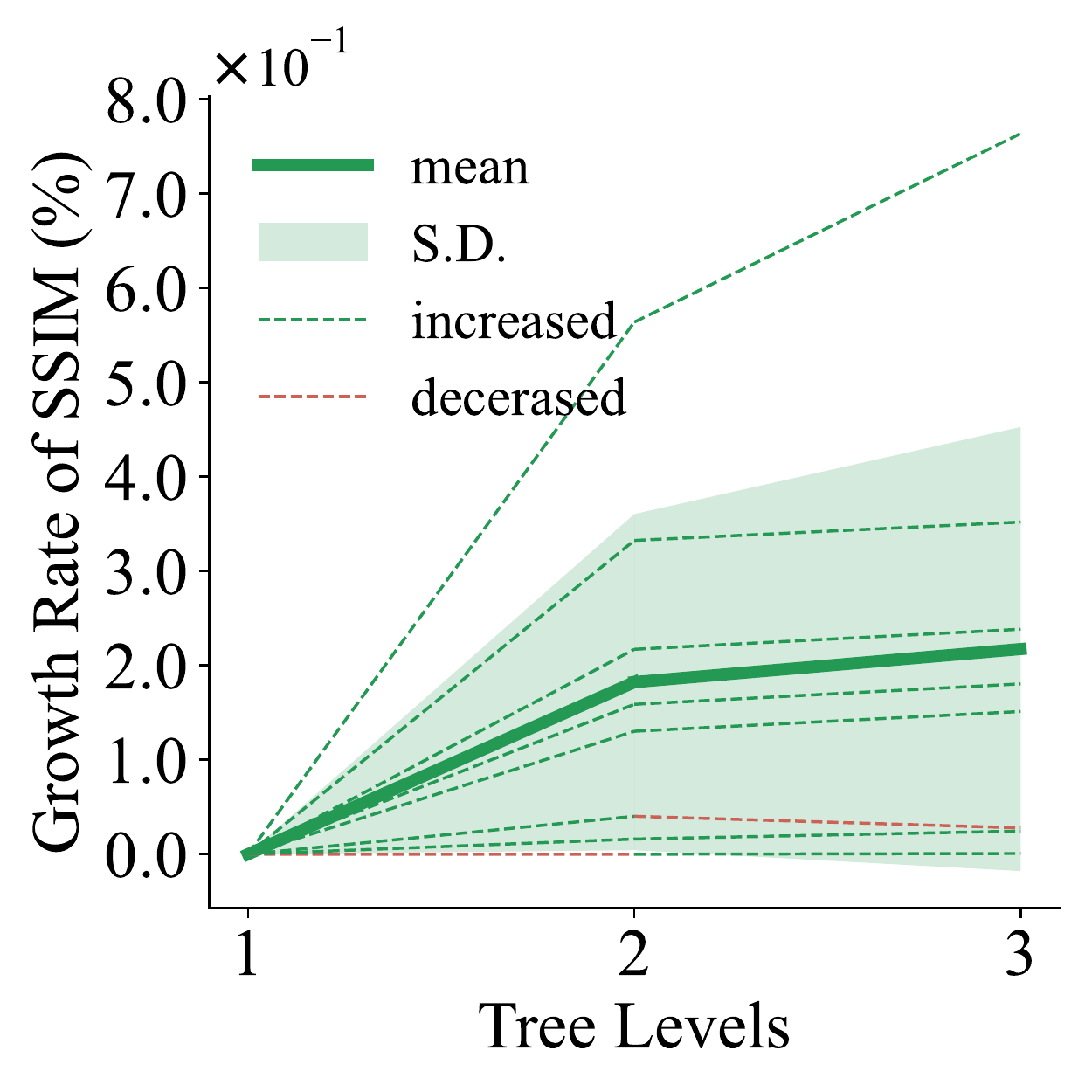}
     \caption{512$\times$}
 \end{subfigure}

 \begin{subfigure}[b]{0.3\textwidth}
     \centering
     \includegraphics[width=\textwidth]{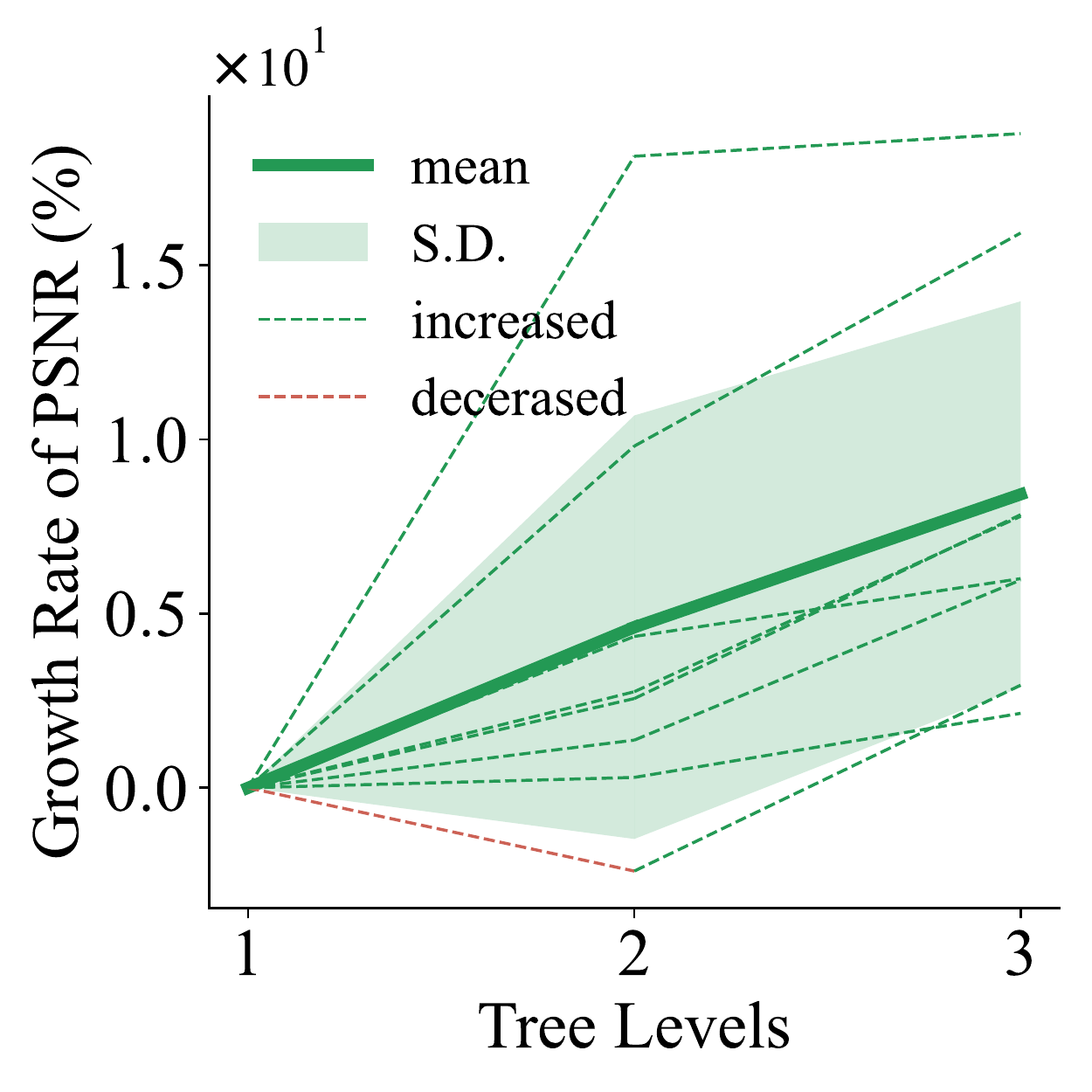}
     \caption{64$\times$}
 \end{subfigure}
  \begin{subfigure}[b]{0.3\textwidth}
     \centering
     \includegraphics[width=\textwidth]{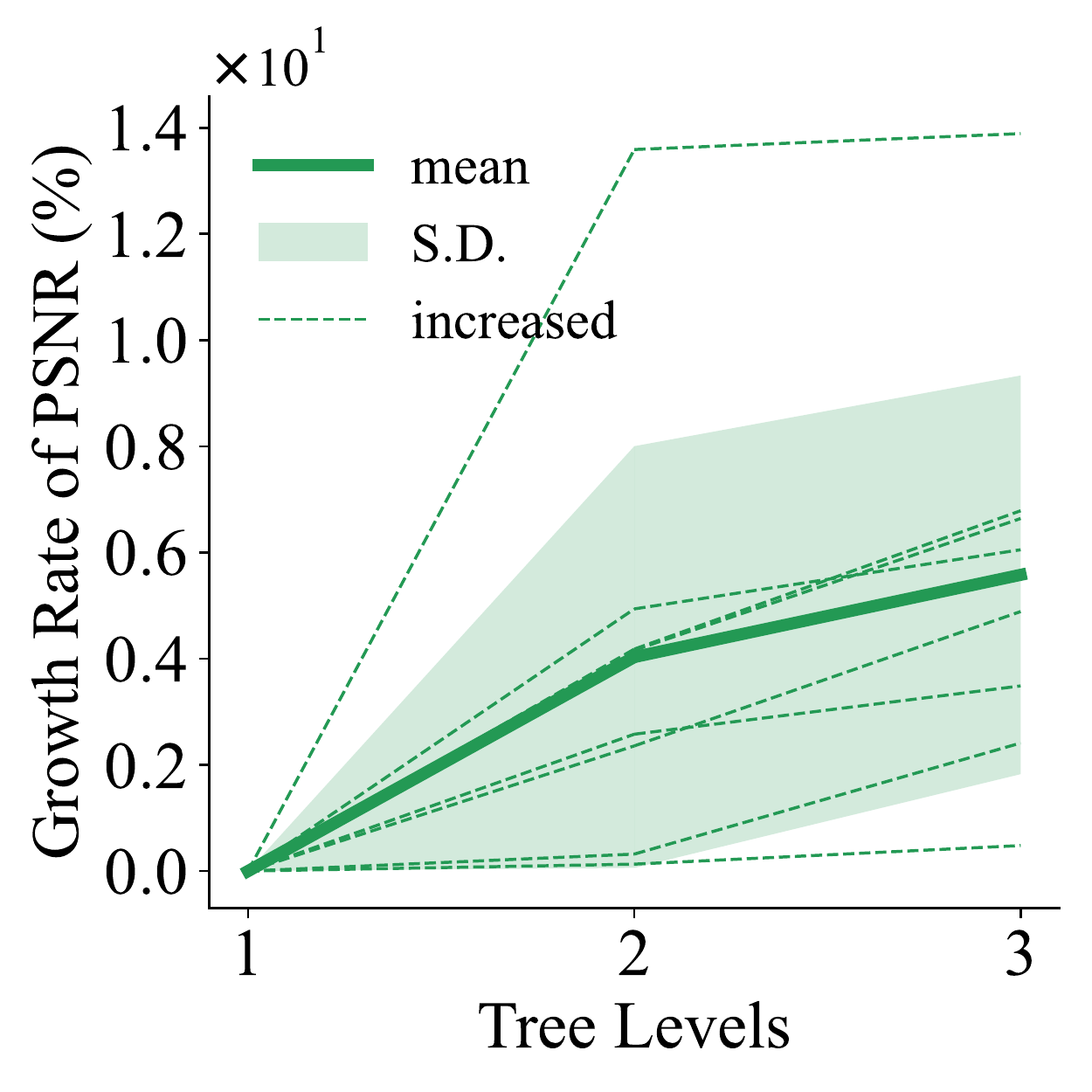}
     \caption{256$\times$}
 \end{subfigure}
  \begin{subfigure}[b]{0.3\textwidth}
     \centering
     \includegraphics[width=\textwidth]{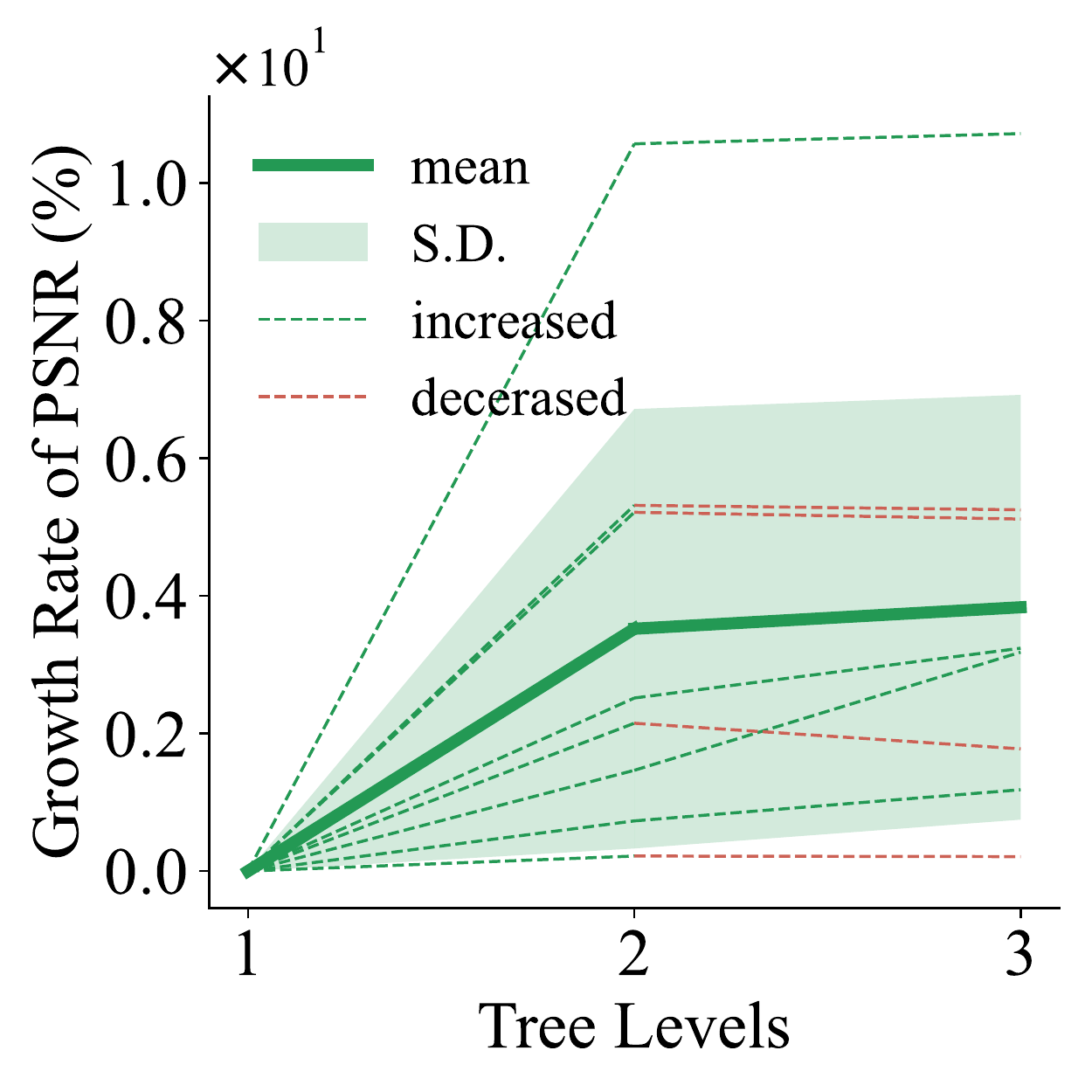}
     \caption{512$\times$}
 \end{subfigure}
\caption{The growth rate of SSIM and PSNR for each data when increasing the tree levels under different compression ratios. The compression ratios are labeled at the bottom of each sub-figure. The dashed lines represent the change in each data. The solid line represents the mean of changes, and the filled area represents the standard deviation, where the positive and negative change are highlighted with green and red respectively.}
\label{figsup: growth rate when increasing the tree levels}
\end{figure}

\begin{figure}[H]
\centering
 \begin{subfigure}[b]{0.3\textwidth}
     \centering
     \includegraphics[width=\textwidth]{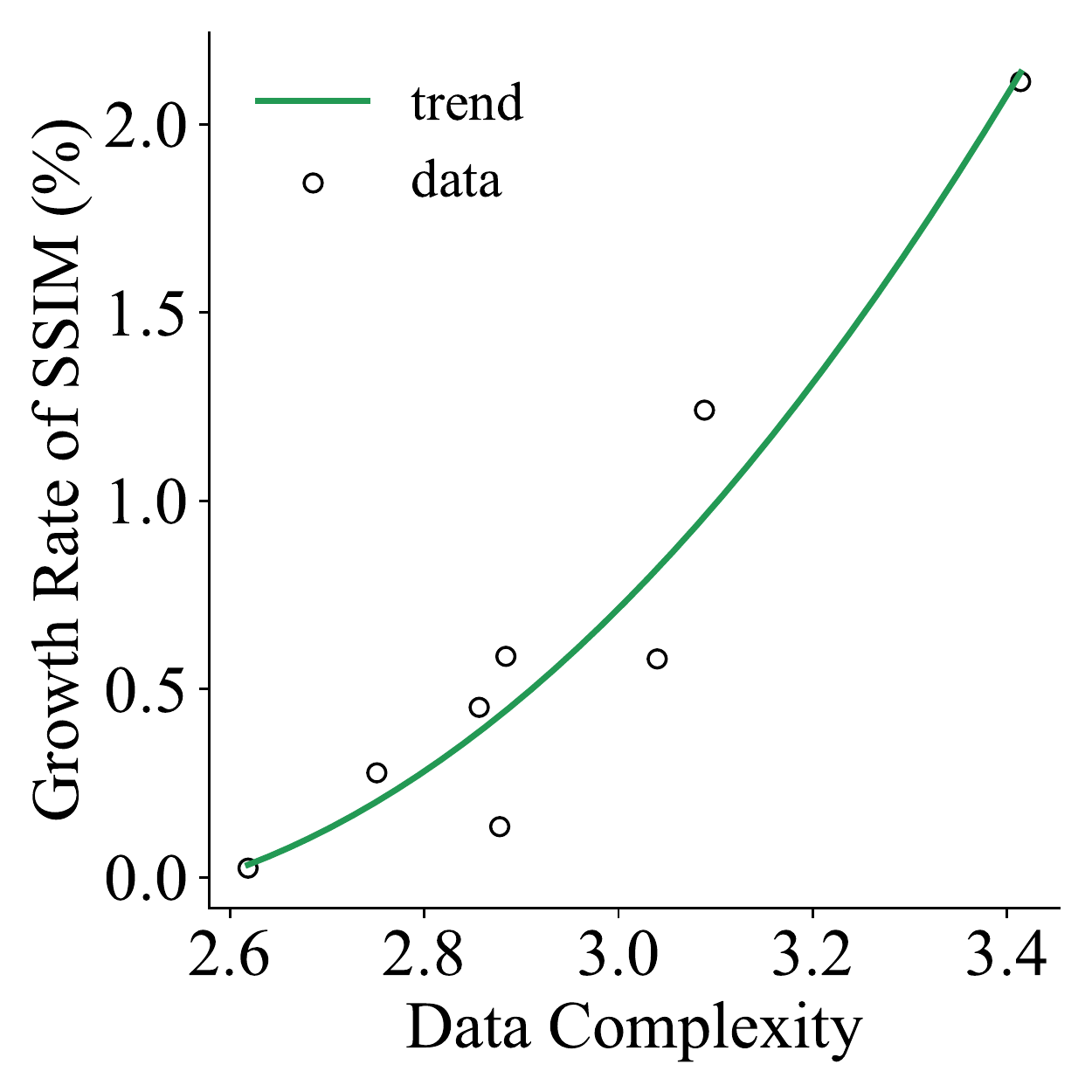}
     \caption{64$\times$}
 \end{subfigure}
  \begin{subfigure}[b]{0.3\textwidth}
     \centering
     \includegraphics[width=\textwidth]{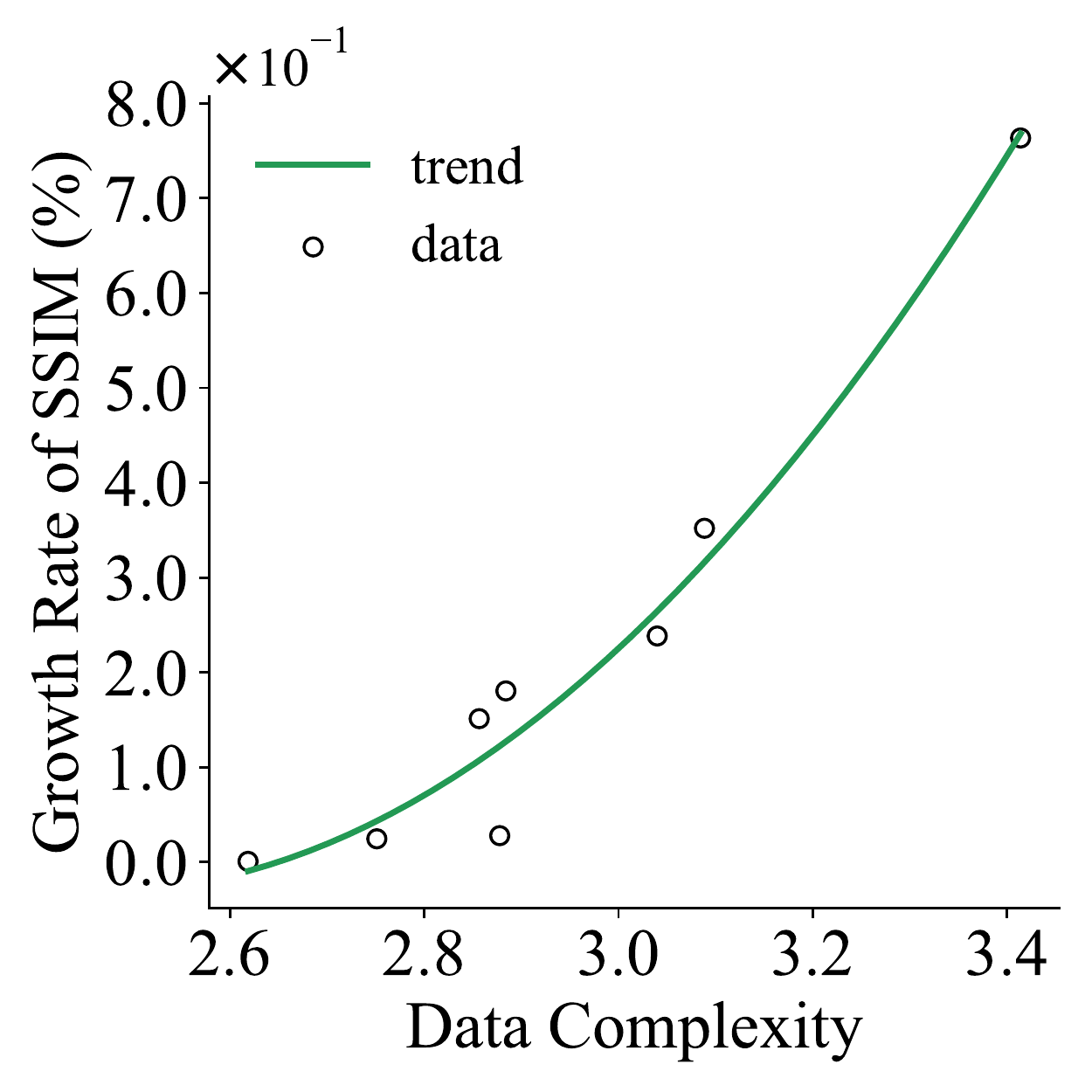}
     \caption{512$\times$}
 \end{subfigure}

 \begin{subfigure}[b]{0.3\textwidth}
     \centering
     \includegraphics[width=\textwidth]{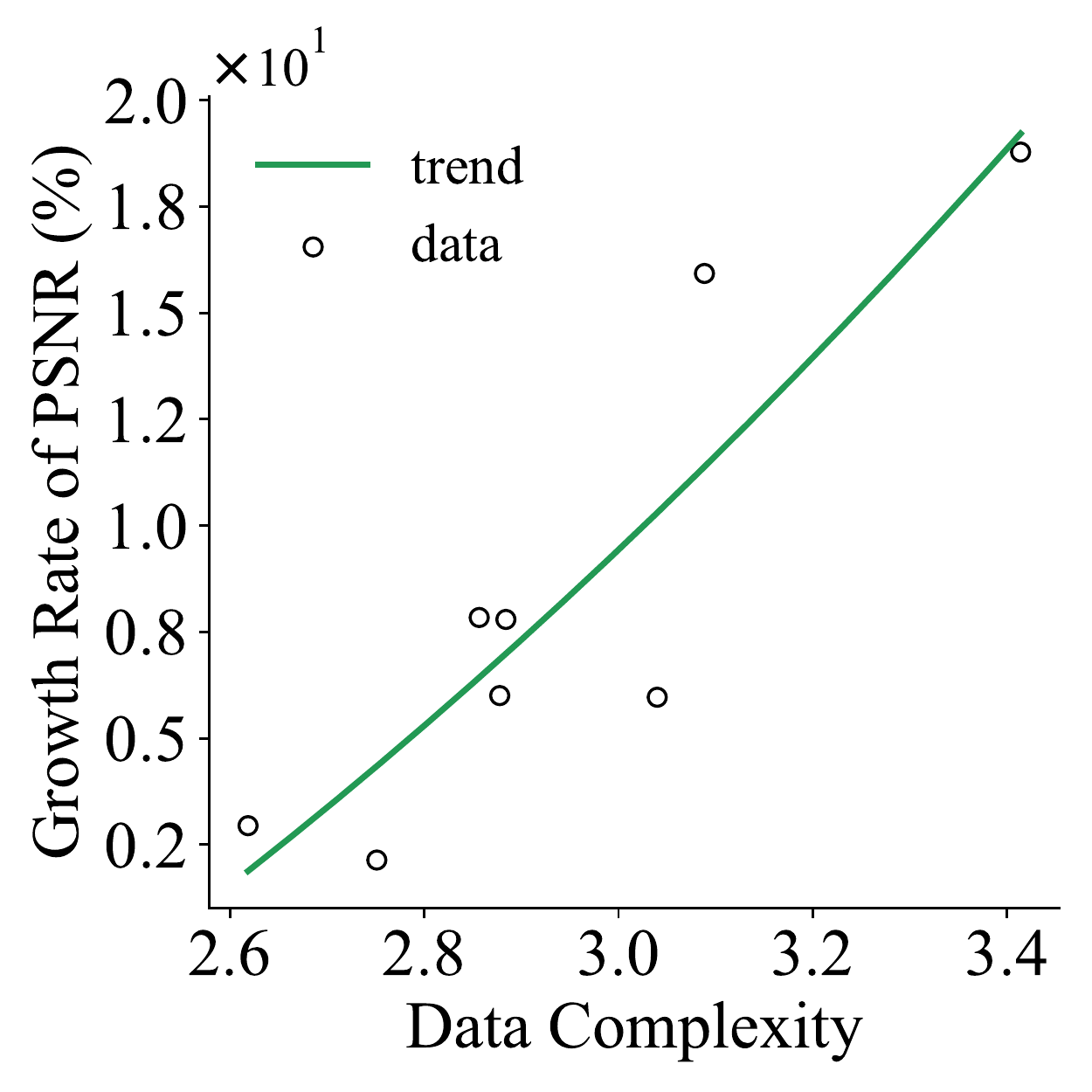}
     \caption{64$\times$}
 \end{subfigure}
  \begin{subfigure}[b]{0.3\textwidth}
     \centering
     \includegraphics[width=\textwidth]{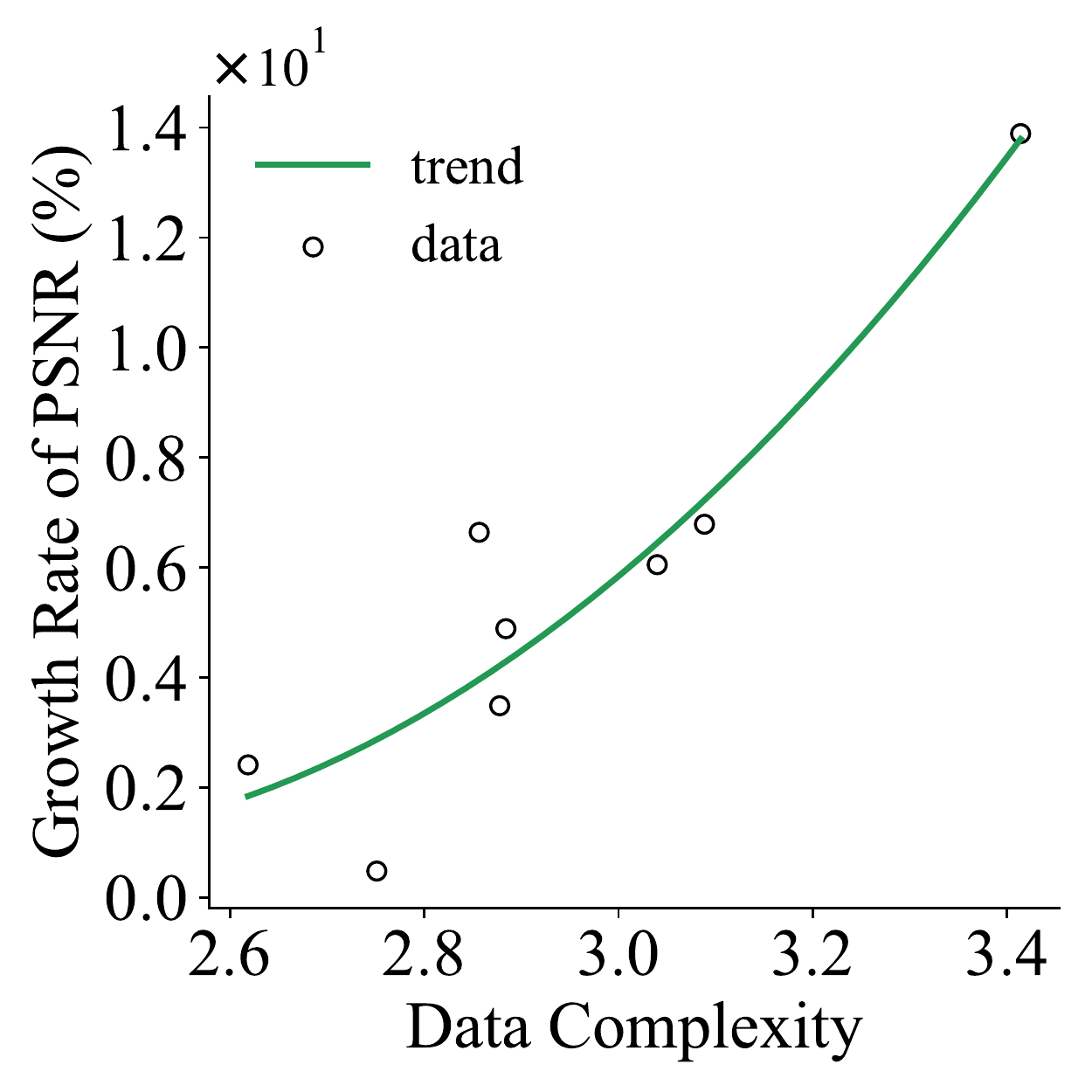}
     \caption{256$\times$}
 \end{subfigure}
  \begin{subfigure}[b]{0.3\textwidth}
     \centering
     \includegraphics[width=\textwidth]{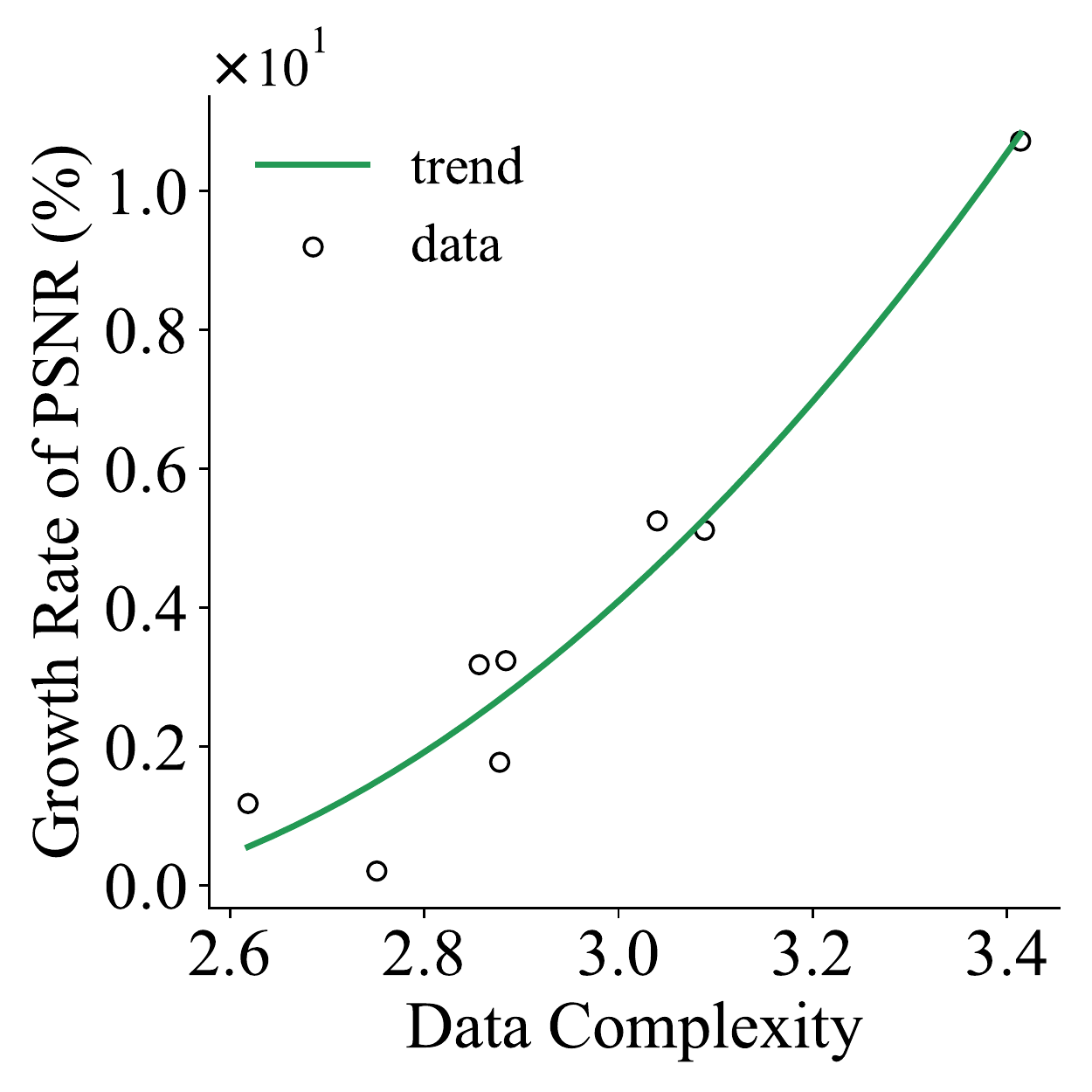}
     \caption{512$\times$}
 \end{subfigure}
\caption{The scatter plot of each data's complexity and the growth rate of SSIM and PSNR when increasing tree levels from 1 to 3 under different compression ratios. The compression ratios are labeled at the bottom of each sub-figure. The solid line represents the trend of change, estimated by a quadratic fit.}
\label{figsup:  scatter plot of each data's complexity and the growth rate}
\end{figure}

\begin{figure}[H]
\centering
  \begin{subfigure}[b]{0.35\textwidth}
     \centering
     \includegraphics[width=\textwidth]{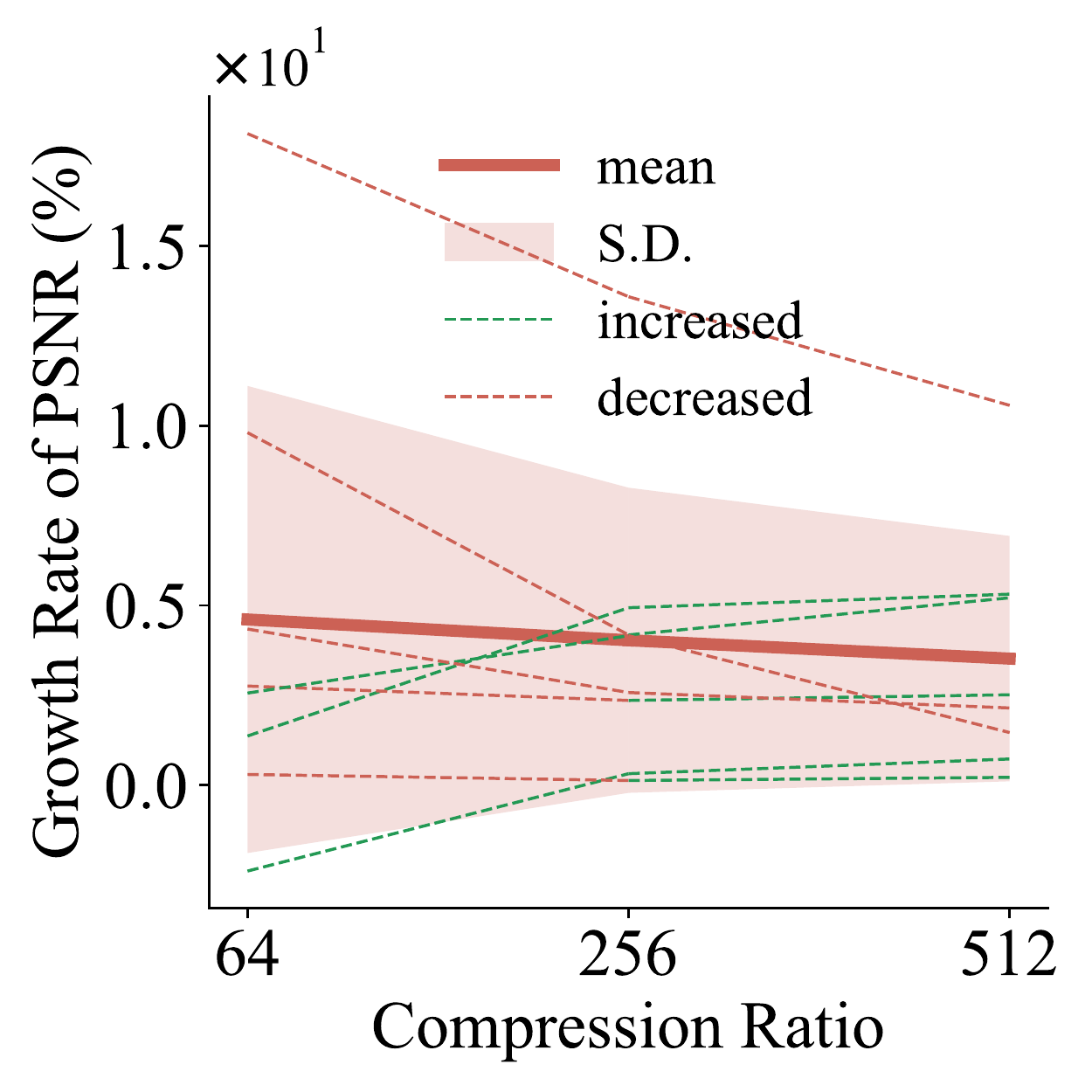}
     \caption{from 1 to 2}
 \end{subfigure}
   \begin{subfigure}[b]{0.35\textwidth}
     \centering
     \includegraphics[width=\textwidth]{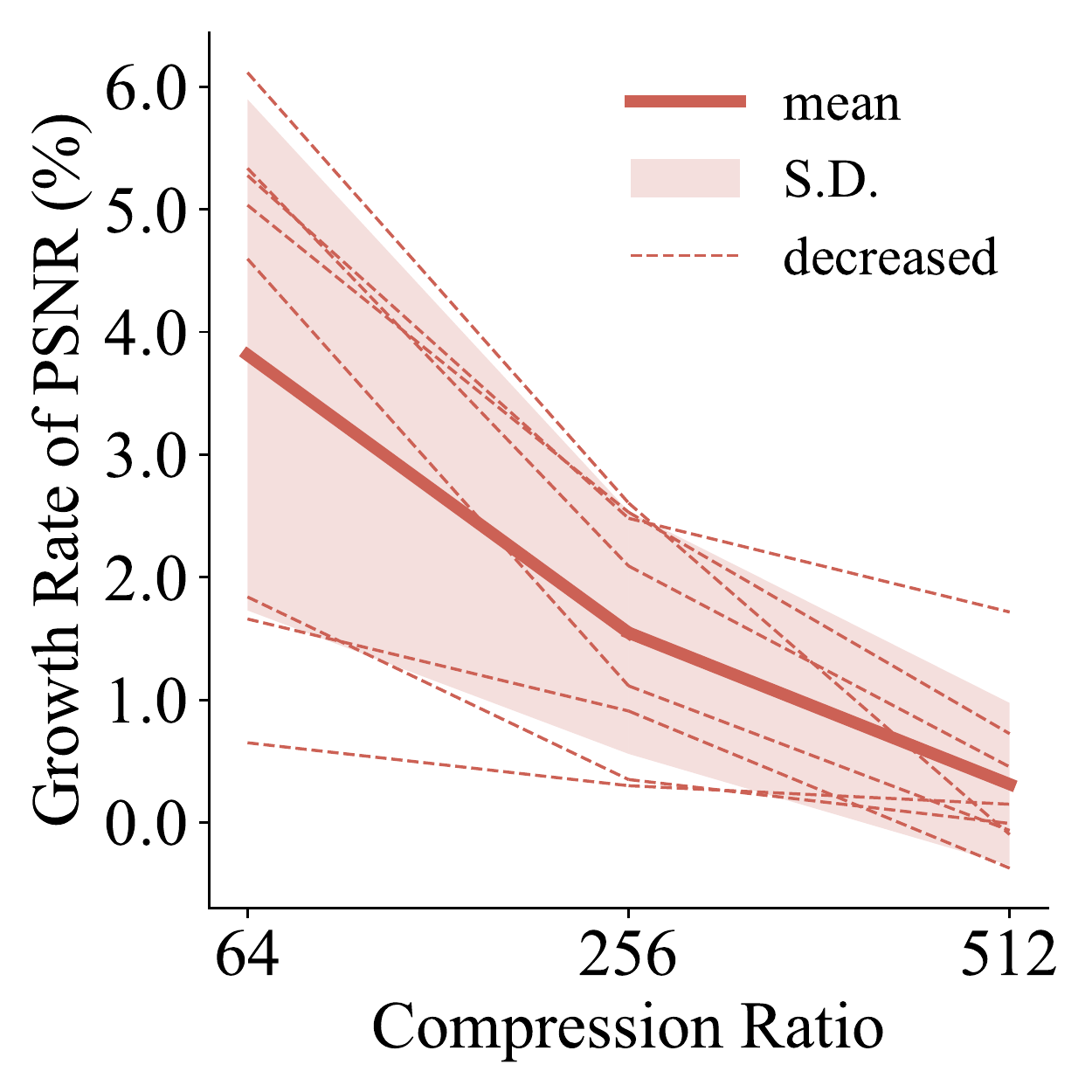}
     \caption{from 2 to 3}
 \end{subfigure}

   \begin{subfigure}[b]{0.35\textwidth}
     \centering
     \includegraphics[width=\textwidth]{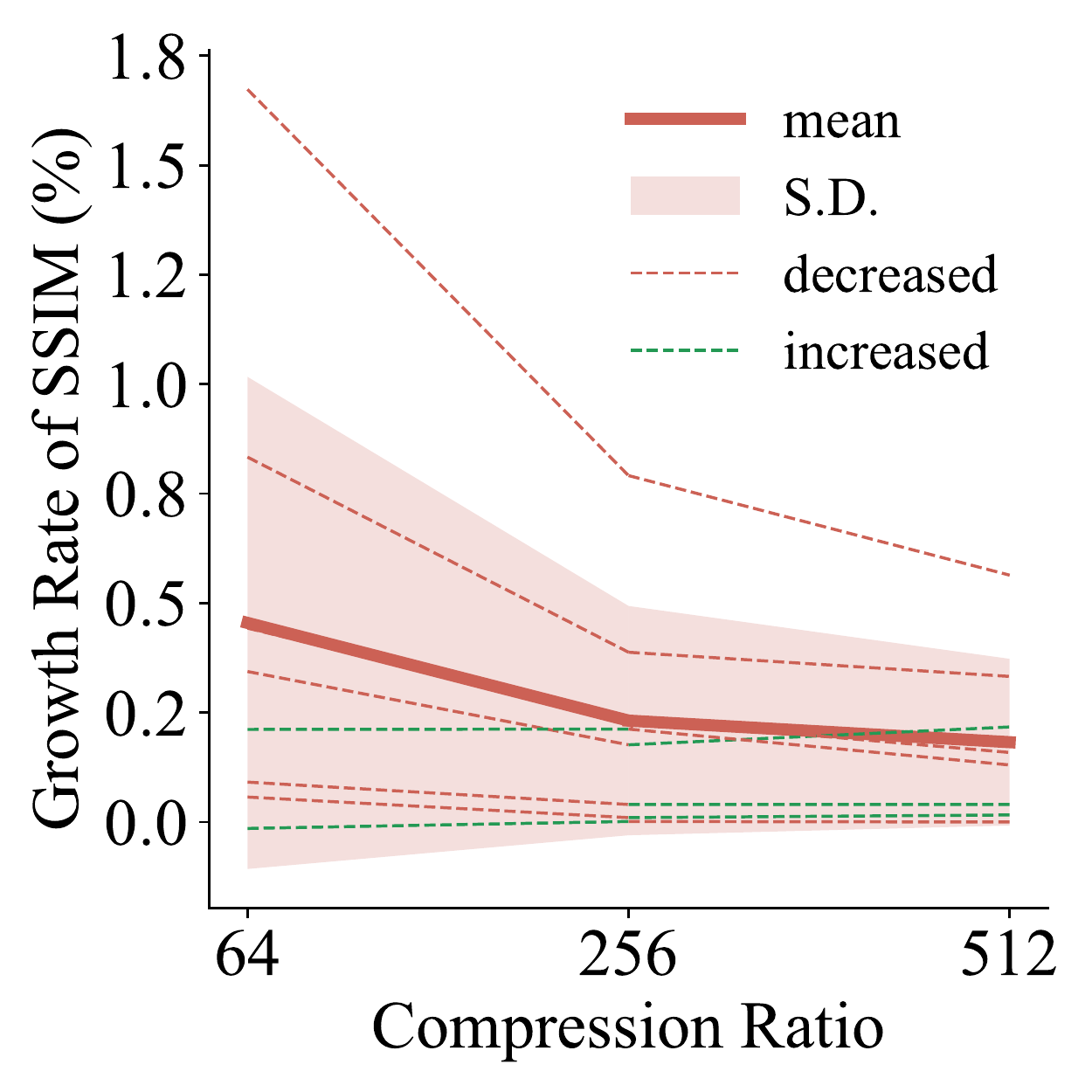}
     \caption{from 1 to 2}
 \end{subfigure}
   \begin{subfigure}[b]{0.35\textwidth}
     \centering
     \includegraphics[width=\textwidth]{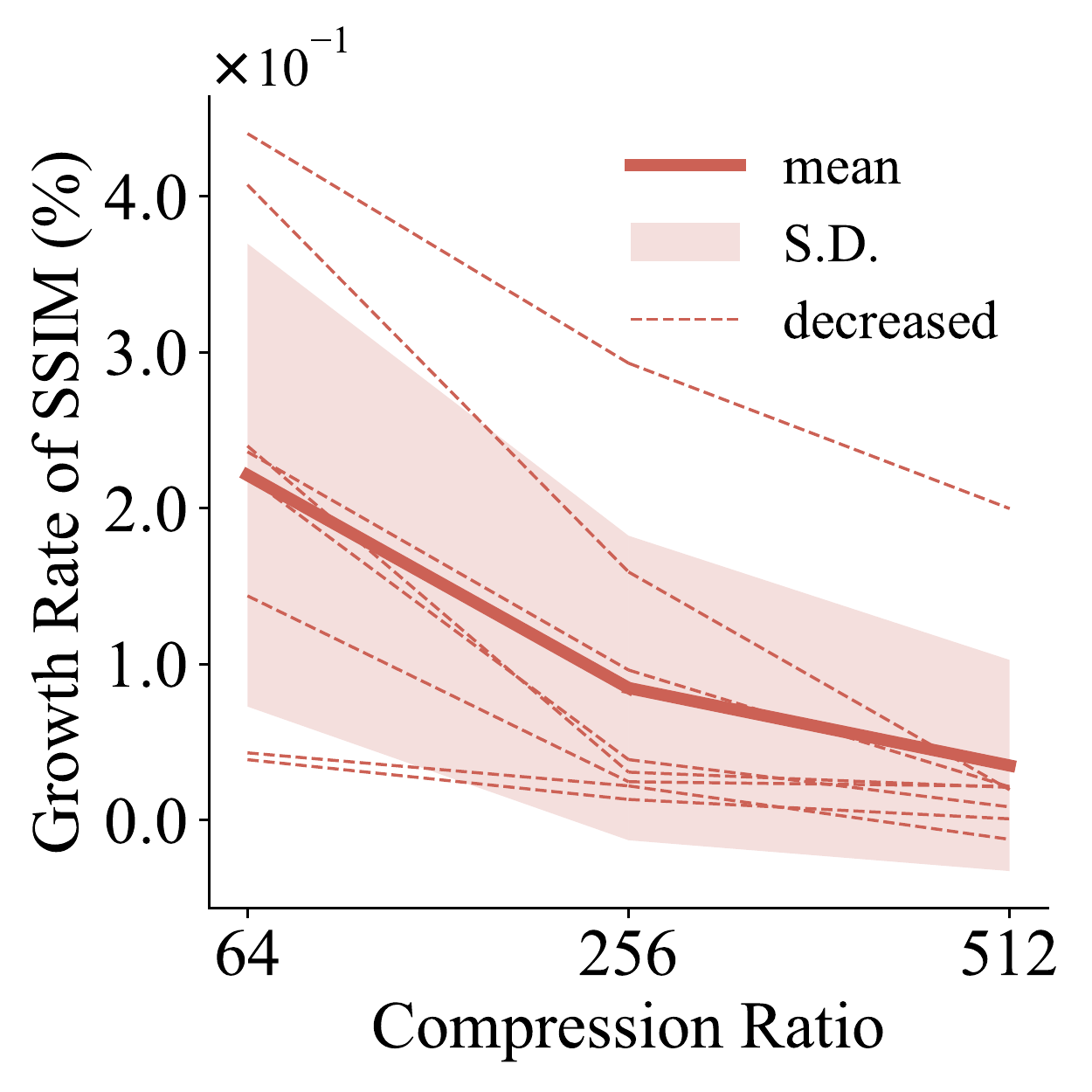}
     \caption{from 2 to 3}
 \end{subfigure}
\caption{The growth rate of SSIM and PSNR when increasing tree levels from 1 to 2 and from 2 to 3 for each data on different compression ratios. The changes of tree levels are labeled at the bottom of each sub-figure. The dashed lines represent the change in each data. The solid line represents the mean of changes, and the filled area represents the standard deviation, where the positive and negative change are highlighted with green and red respectively.}
\label{figsup: growth rate when increasing tree levels on different compression ratios}
\end{figure}

\begin{figure}[H]
\centering
 \begin{subfigure}[b]{0.4\textwidth}
     \centering
     \includegraphics[width=\textwidth]{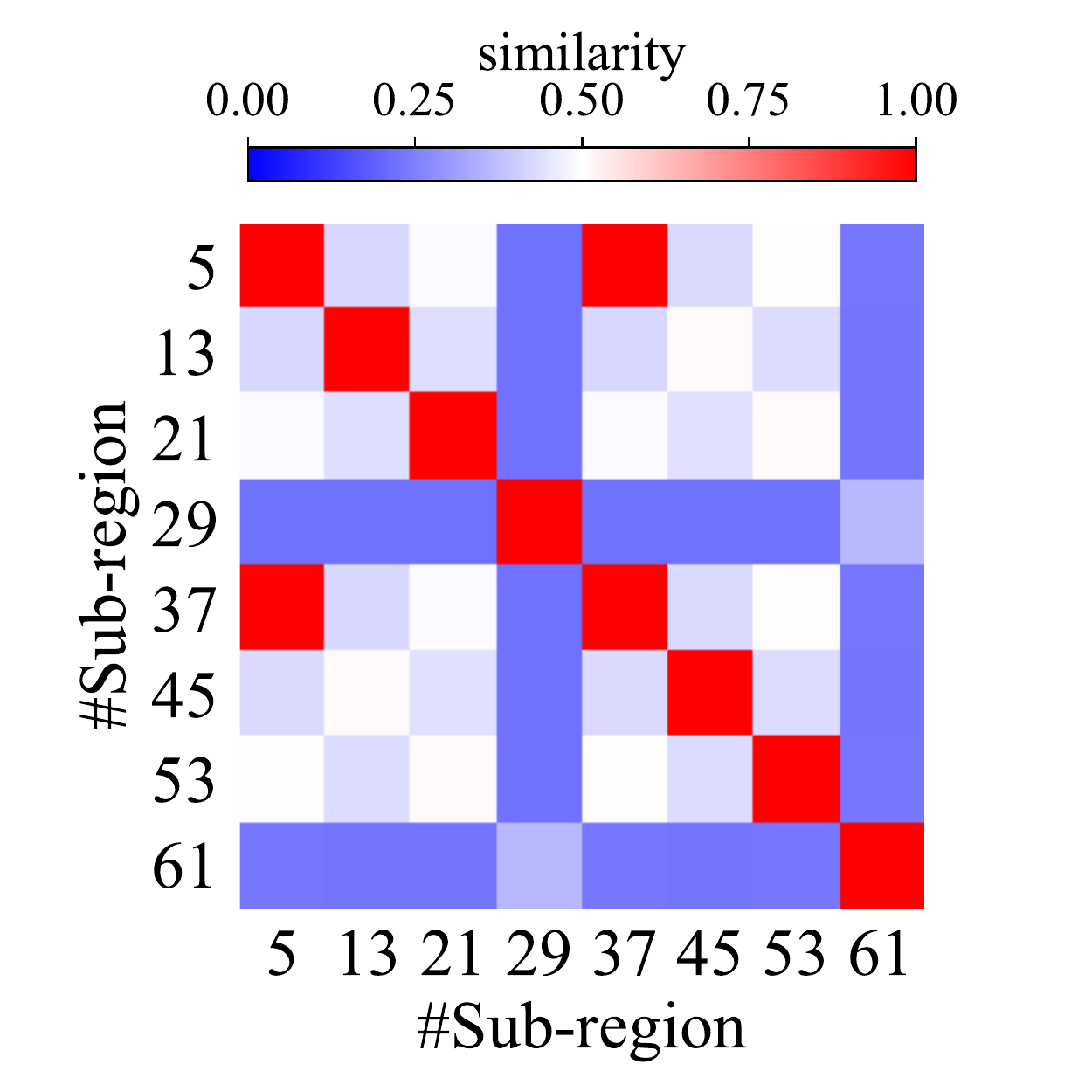}
     \caption{}
 \end{subfigure}
  \begin{subfigure}[b]{0.2\textwidth}
     \centering
     \includegraphics[width=\textwidth]{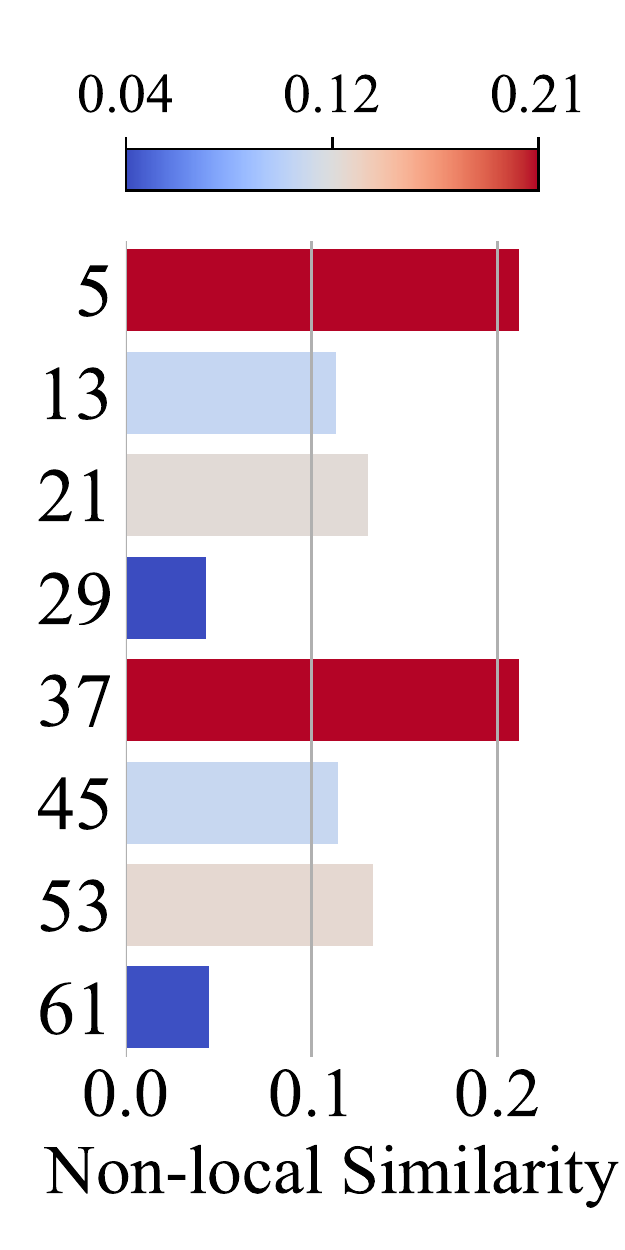}
     \caption{}
 \end{subfigure}
  \begin{subfigure}[b]{0.2\textwidth}
     \centering
     \includegraphics[width=\textwidth]{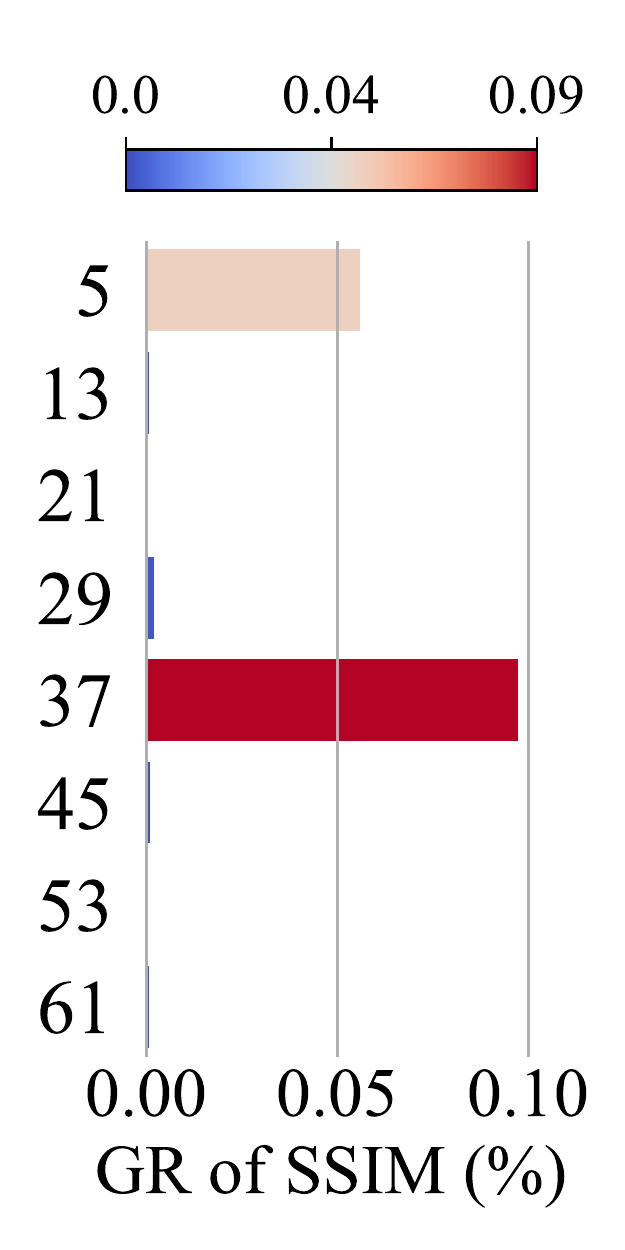}
     \caption{}
 \end{subfigure}
\caption{Effect of allocating more parameters to the shallow level on \M's compression fidelity for each sub-regions in a \textit{Brain} data under 256$\times$ compression ratio. All sub-figures share the same y-axis label.
(a) The heatmap of similarities between 8 equally spaced distant sub-regions. The serial numbers of the regions represent their z-curve order.
(b) The non-local similarity of each region.
(c) The growth rate of SSIM for each region when allocating more parameters to the shallow level.}
\label{figsup: shallow level brain data heatmap}
\end{figure}

\begin{figure}[H]
\centering
 \begin{subfigure}[b]{0.4\textwidth}
     \centering
     \includegraphics[width=\textwidth]{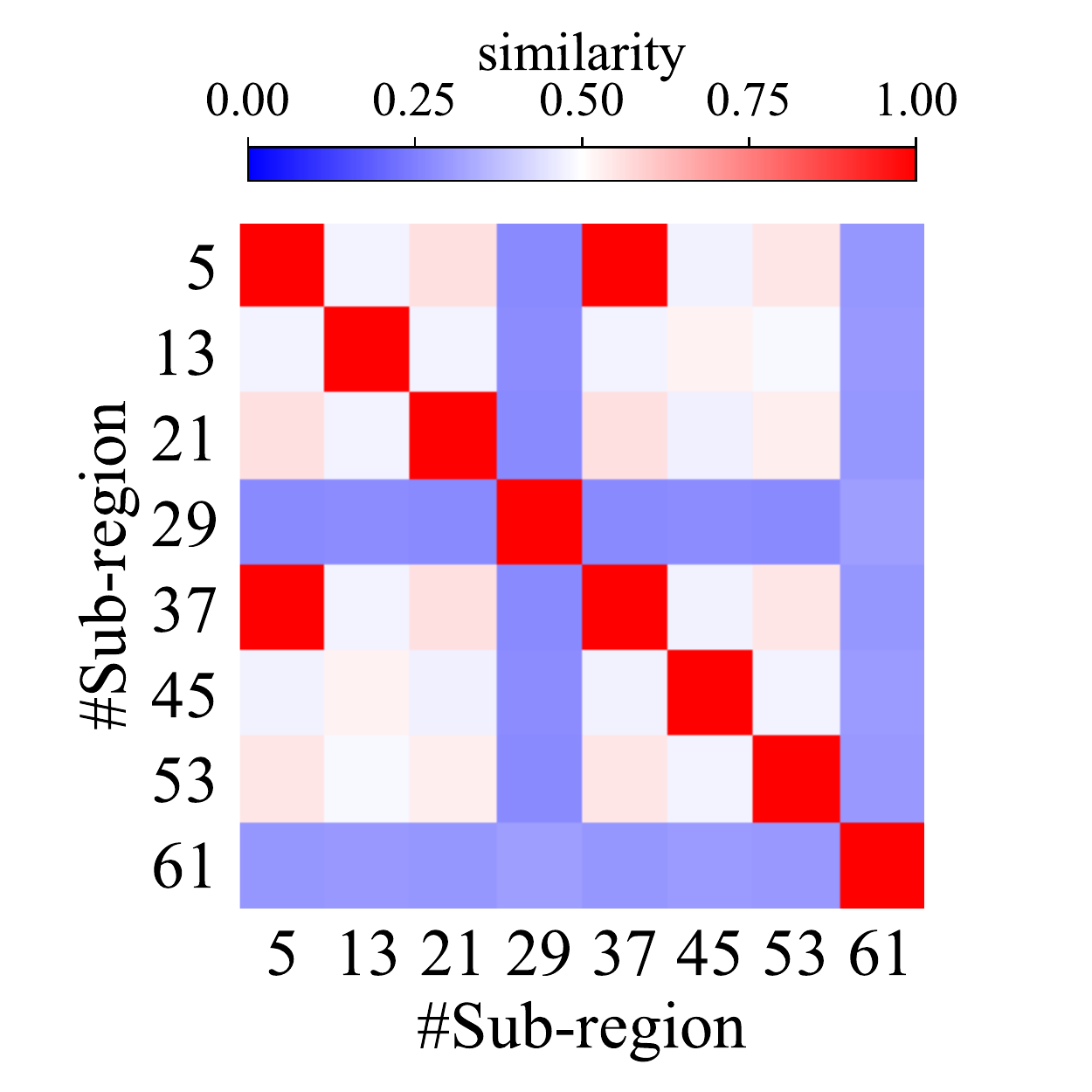}
     \caption{}
 \end{subfigure}
  \begin{subfigure}[b]{0.2\textwidth}
     \centering
     \includegraphics[width=\textwidth]{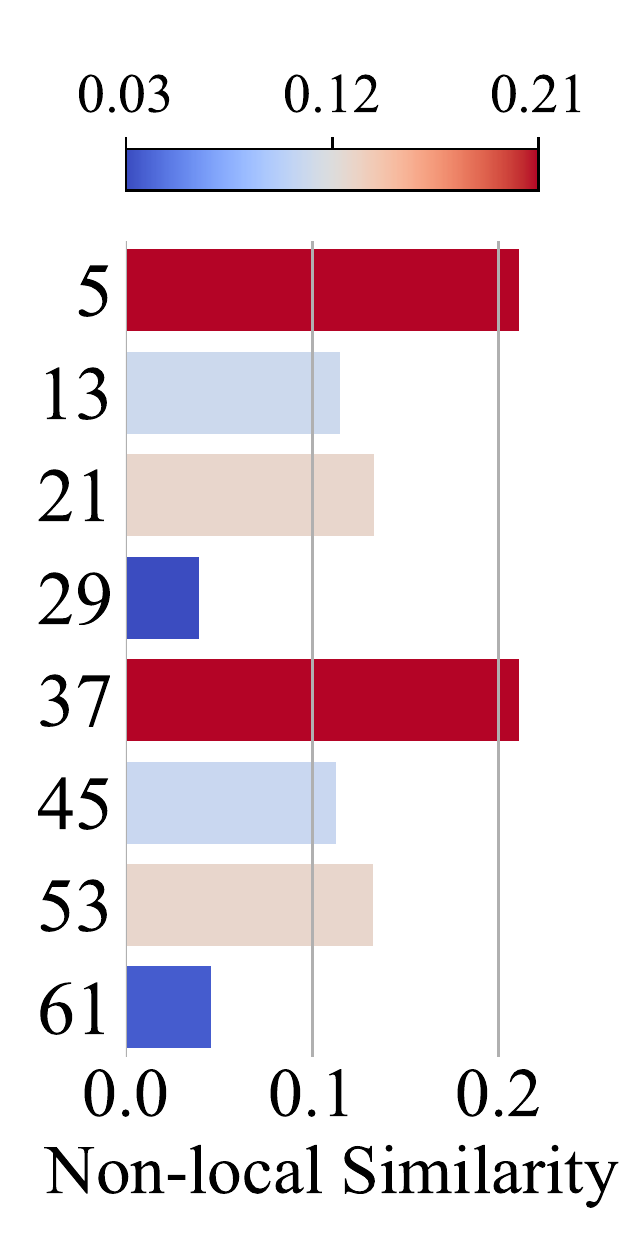}
     \caption{}
 \end{subfigure}
  \begin{subfigure}[b]{0.2\textwidth}
     \centering
     \includegraphics[width=\textwidth]{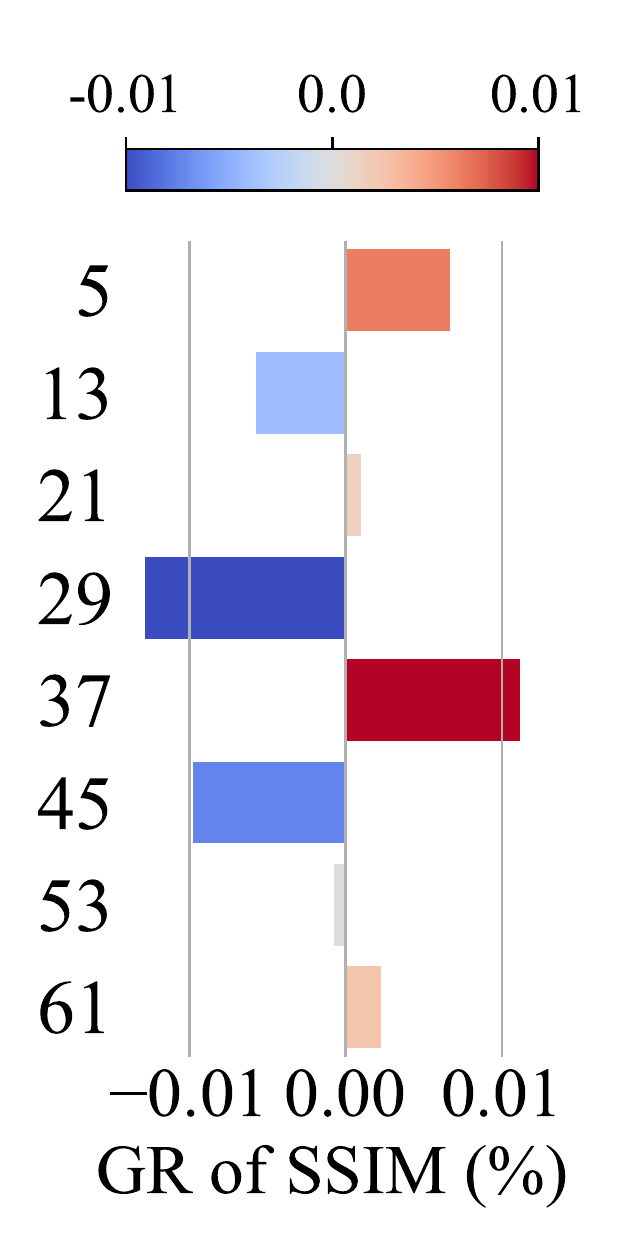}
     \caption{}
 \end{subfigure}
\caption{Effect of allocating more parameters to the shallow level on \M's compression fidelity for each sub-regions in a \textit{Heart} data under 512$\times$ compression ratio. With the same layout in Supplementay Figure.~\ref{figsup: shallow level brain data heatmap}}
\label{figsup: shallow level heart data heatmap}
\end{figure}

\begin{figure}[H]
\centering
  \begin{subfigure}[b]{0.3\textwidth}
     \centering
     \includegraphics[width=\textwidth]{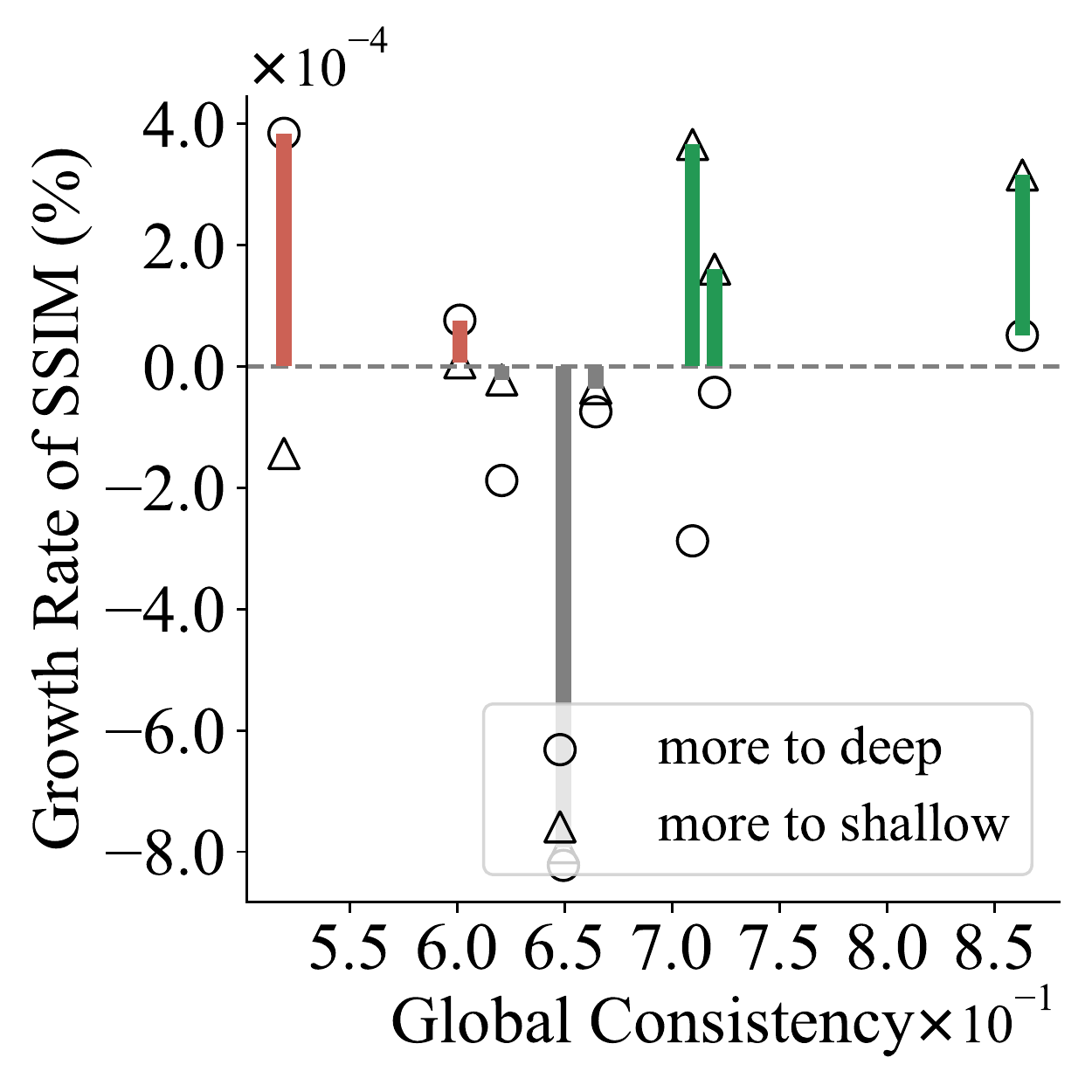}
     \caption{64$\times$}
 \end{subfigure}
  \begin{subfigure}[b]{0.3\textwidth}
     \centering
     \includegraphics[width=\textwidth]{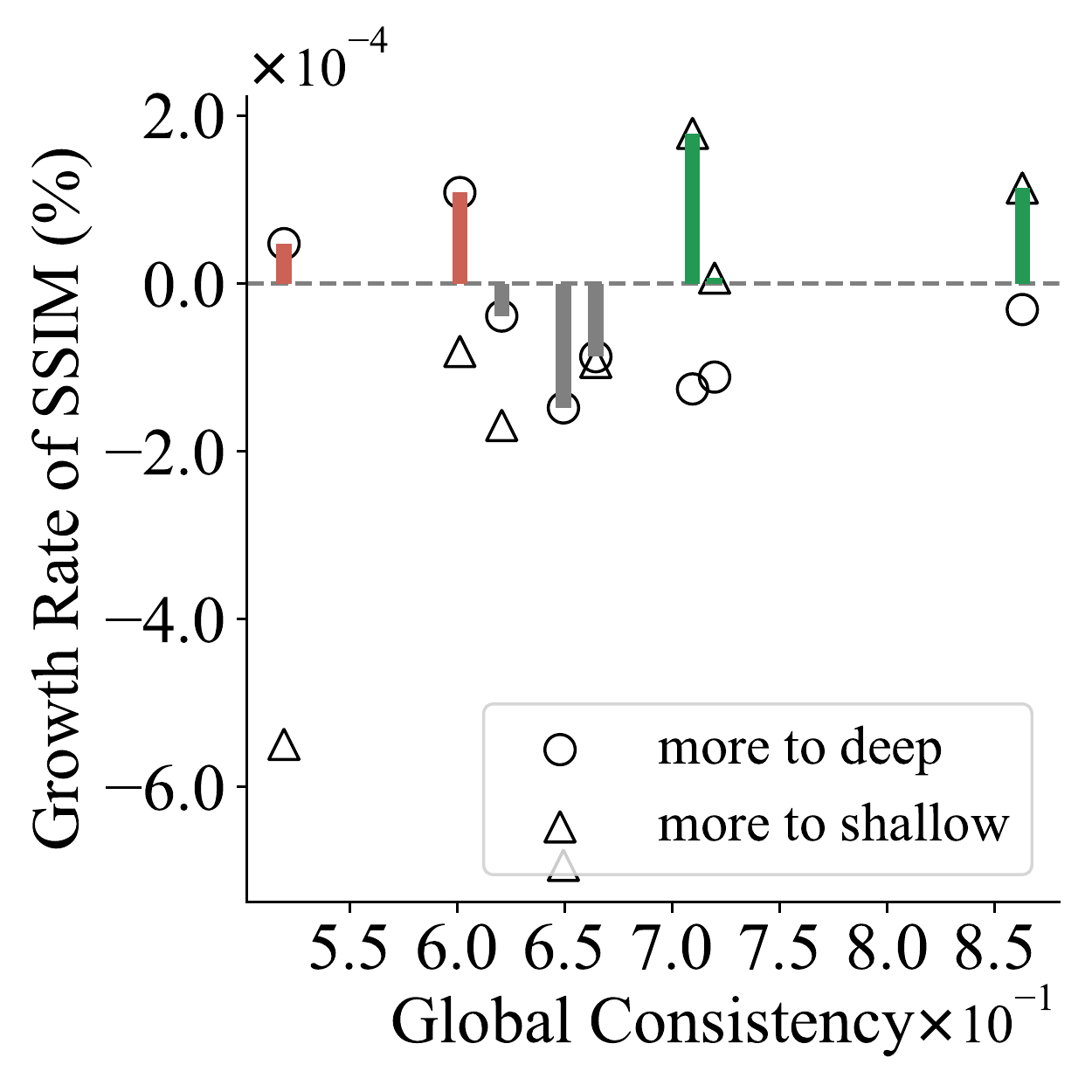}
     \caption{256$\times$}
 \end{subfigure}
  \begin{subfigure}[b]{0.3\textwidth}
     \centering
     \includegraphics[width=\textwidth]{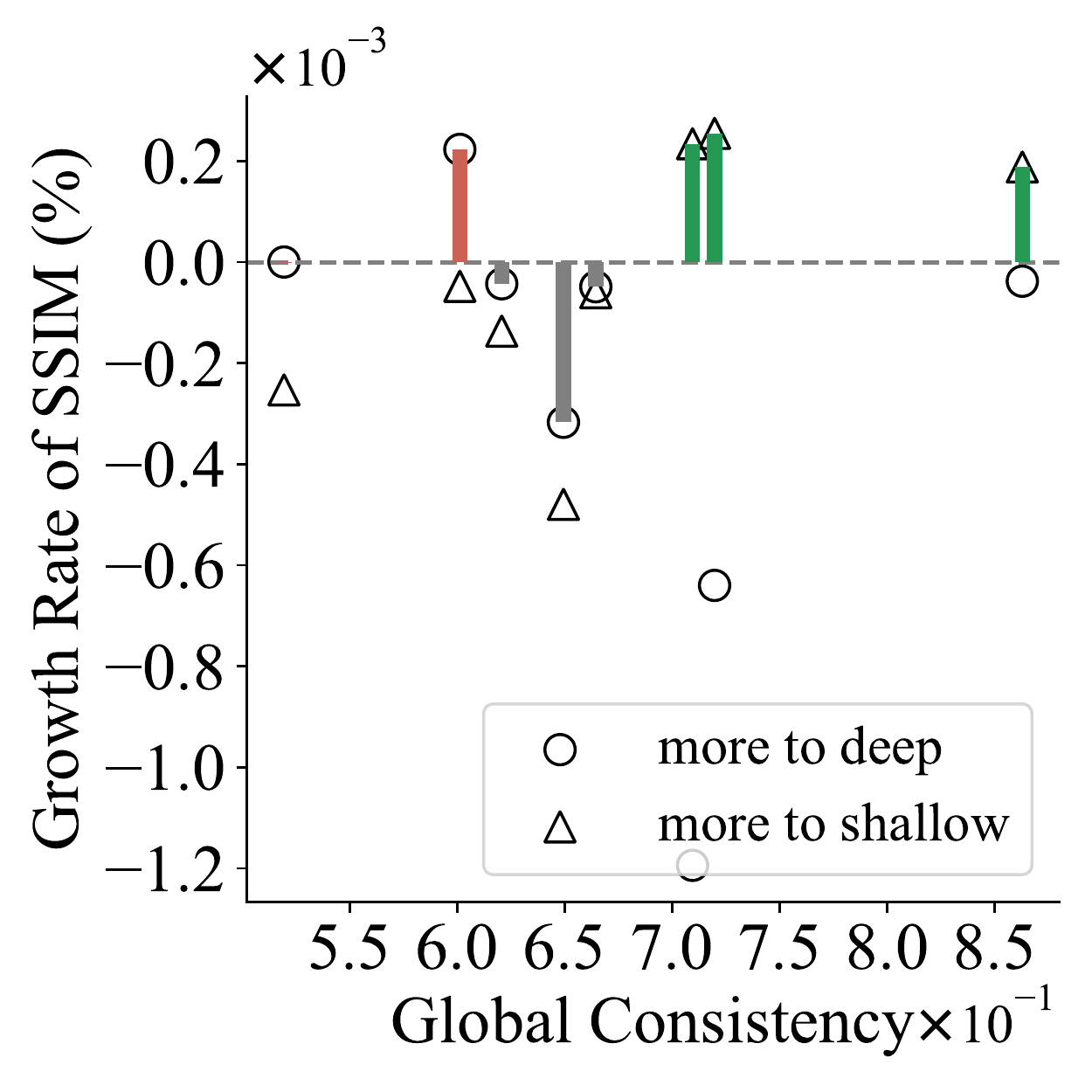}
     \caption{512$\times$}
 \end{subfigure}

 \begin{subfigure}[b]{0.3\textwidth}
     \centering
     \includegraphics[width=\textwidth]{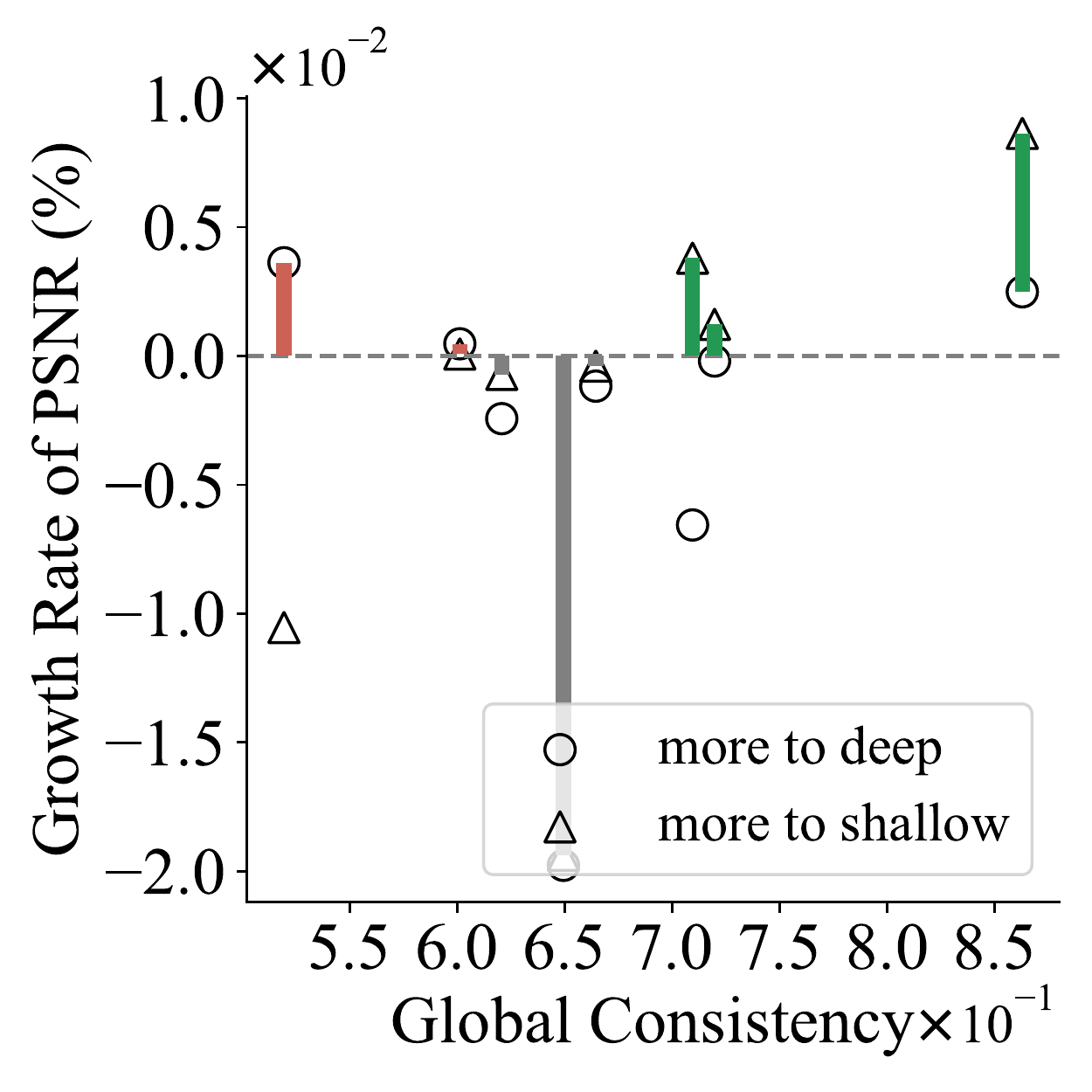}
     \caption{64$\times$}
 \end{subfigure}
  \begin{subfigure}[b]{0.3\textwidth}
     \centering
     \includegraphics[width=\textwidth]{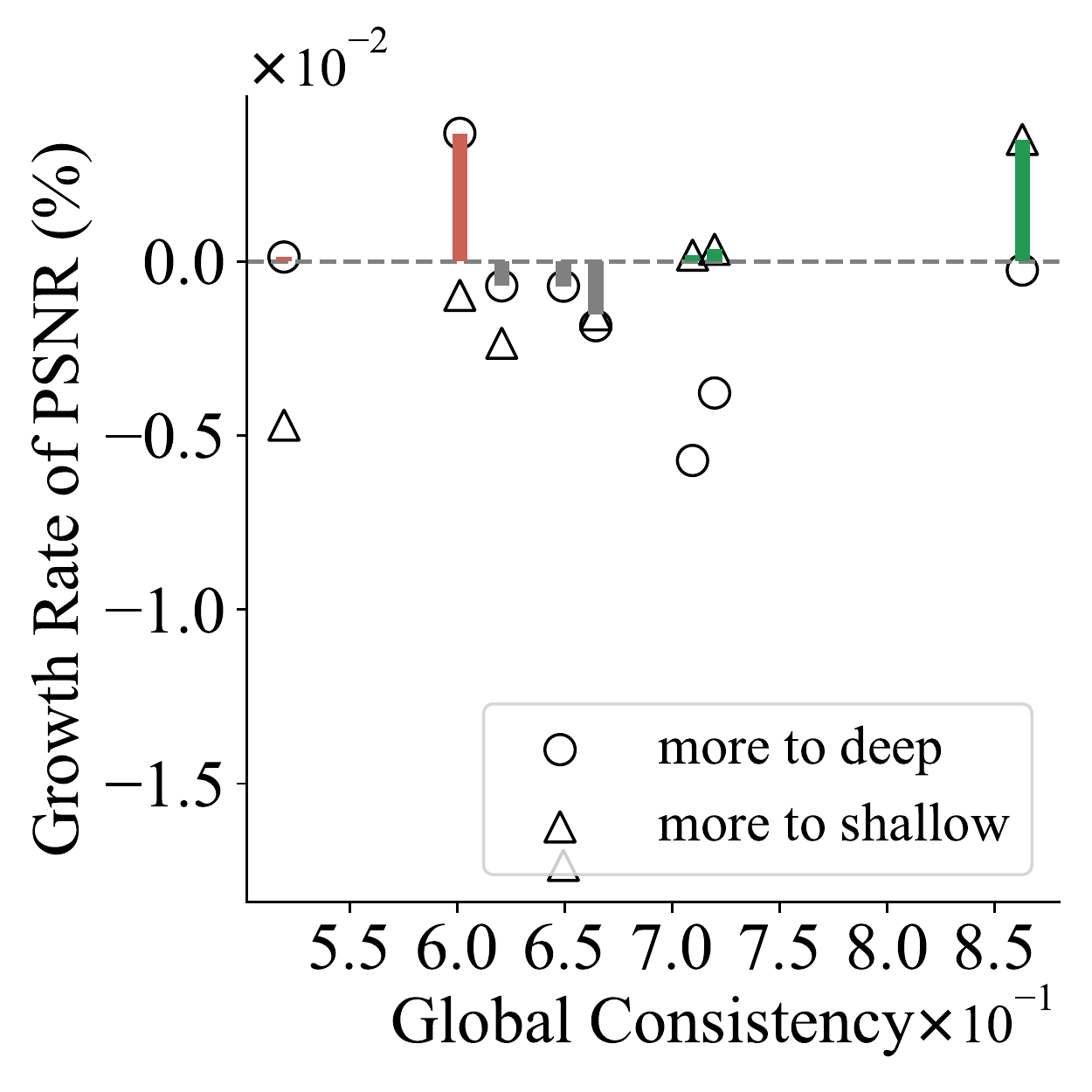}
     \caption{256$\times$}
 \end{subfigure}
  \begin{subfigure}[b]{0.3\textwidth}
     \centering
     \includegraphics[width=\textwidth]{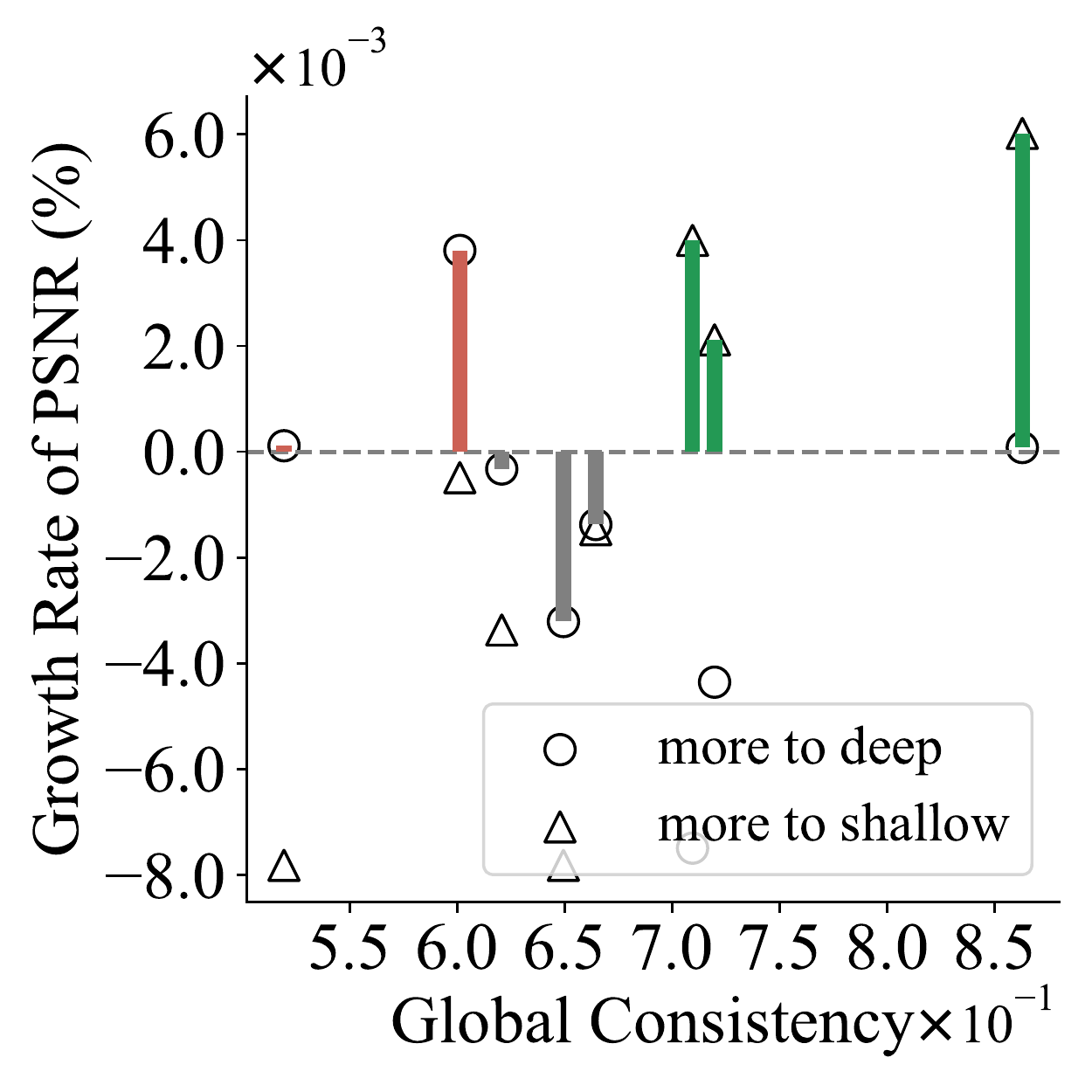}
     \caption{512$\times$}
 \end{subfigure}
\caption{Effect of three inter-level parameter allocation approaches on \M's compression fidelity for medical data with different global consistency under different compression ratios. The compression ratios are labeled at the bottom of each sub-figure. We used the even allocation as baseline and calculated the growth rate of SSIM and PSNR for the other two approaches, represented as squares and triangles. We marked the gap between the best and second-best approaches, and use red, gray, and green colors to indicate the three cases of allocation more to the deep being the best, even allocation being the best, and allocation more to the shallow being the best, respectively}
\label{figsup: inter-level parameter allocation global consistency}
\end{figure}

\begin{figure}[H]
\centering
  \begin{subfigure}[b]{0.15\textwidth}
     \centering
     \includegraphics[width=\textwidth]{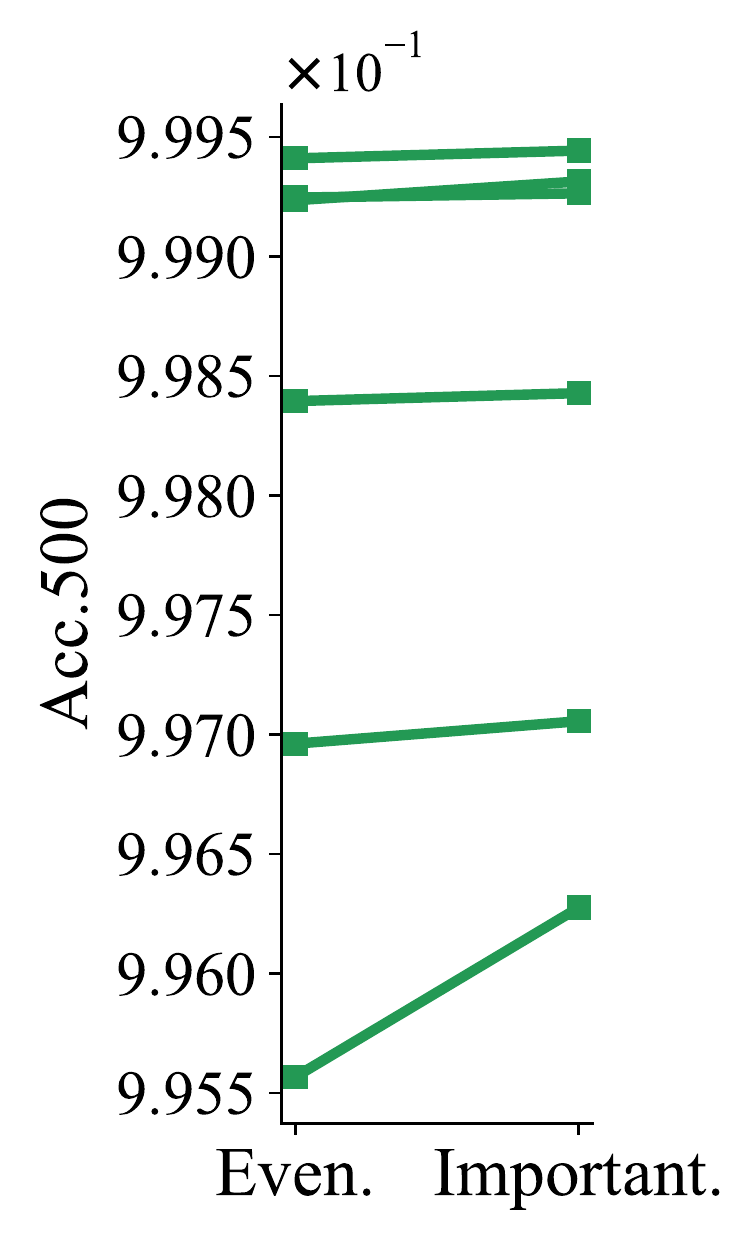}
     \caption{64$\times$}
 \end{subfigure}
  \begin{subfigure}[b]{0.15\textwidth}
     \centering
     \includegraphics[width=\textwidth]{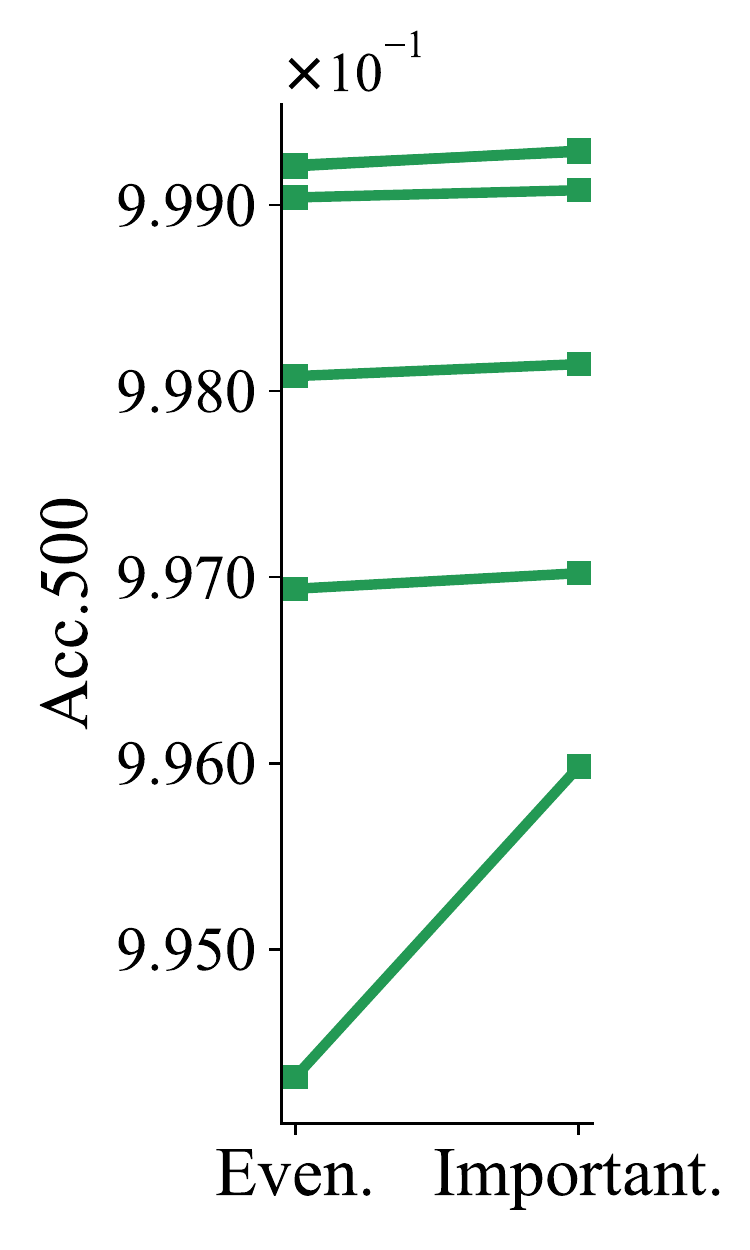}
     \caption{256$\times$}
 \end{subfigure}
  \begin{subfigure}[b]{0.15\textwidth}
     \centering
     \includegraphics[width=\textwidth]{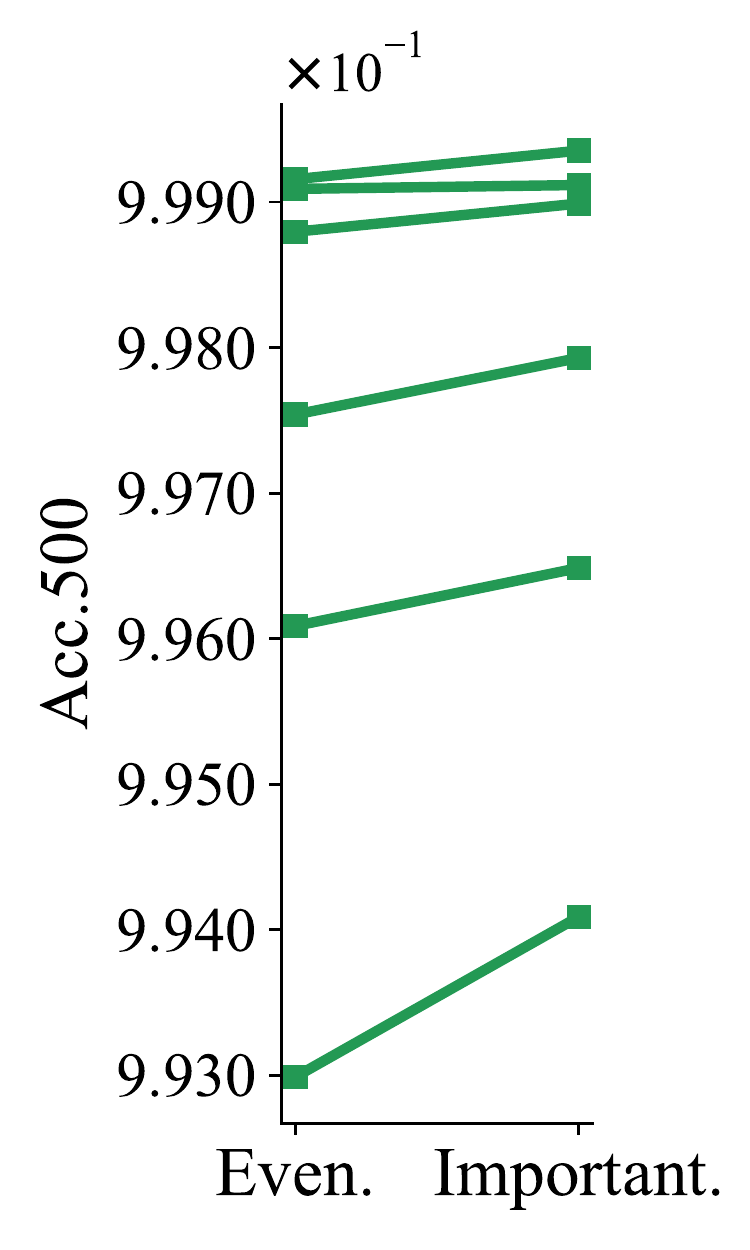}
     \caption{512$\times$}
 \end{subfigure}
 \begin{subfigure}[b]{0.15\textwidth}
     \centering
     \includegraphics[width=\textwidth]{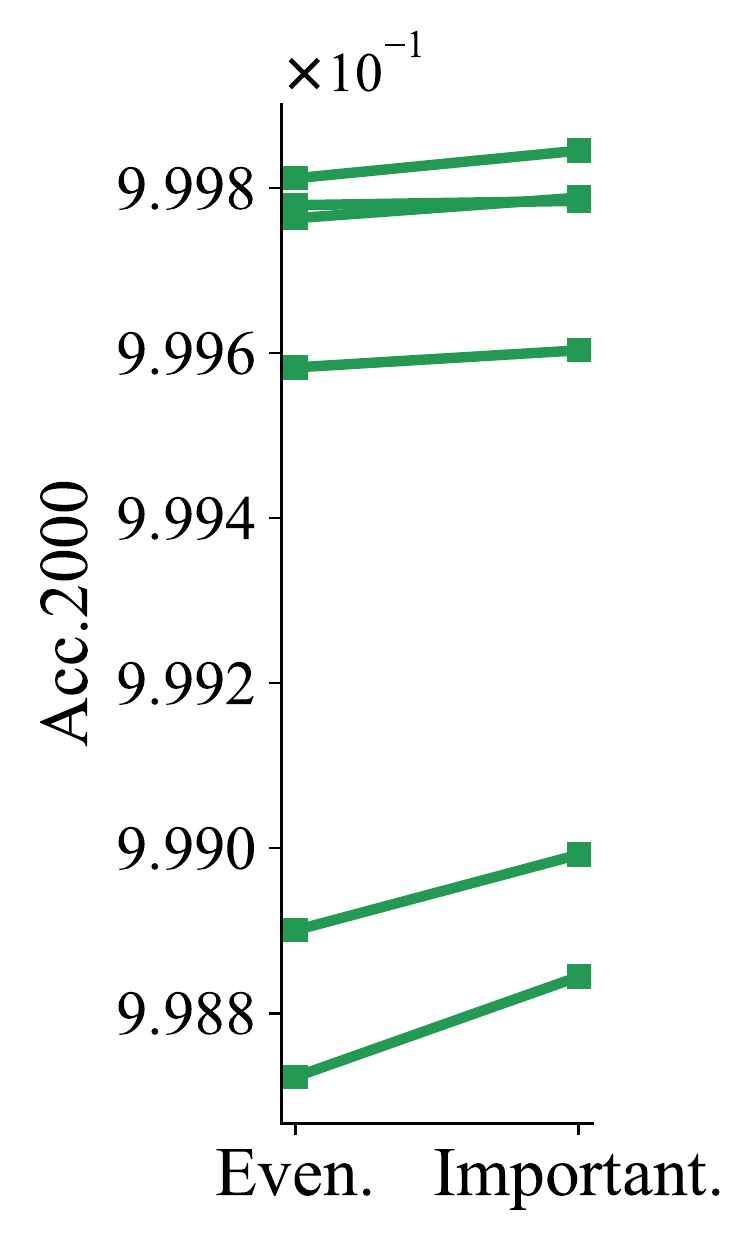}
     \caption{64$\times$}
 \end{subfigure}
  \begin{subfigure}[b]{0.15\textwidth}
     \centering
     \includegraphics[width=\textwidth]{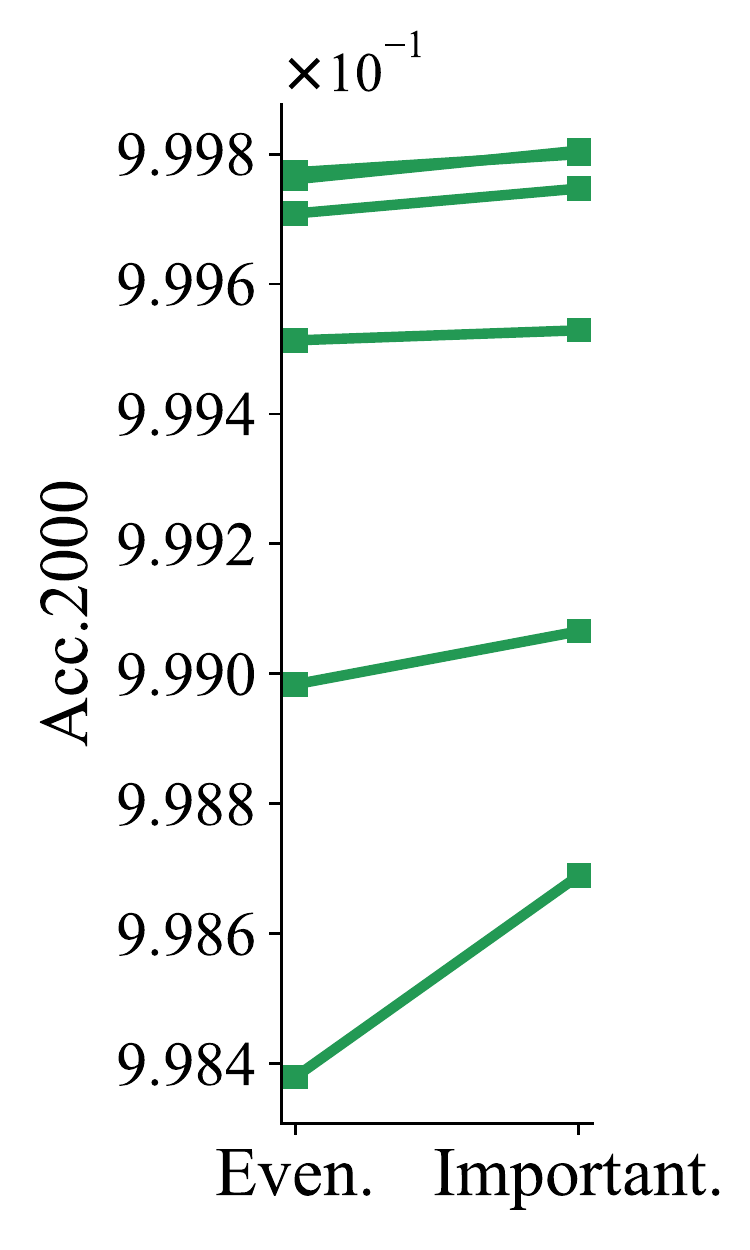}
     \caption{256$\times$}
 \end{subfigure}
  \begin{subfigure}[b]{0.15\textwidth}
     \centering
     \includegraphics[width=\textwidth]{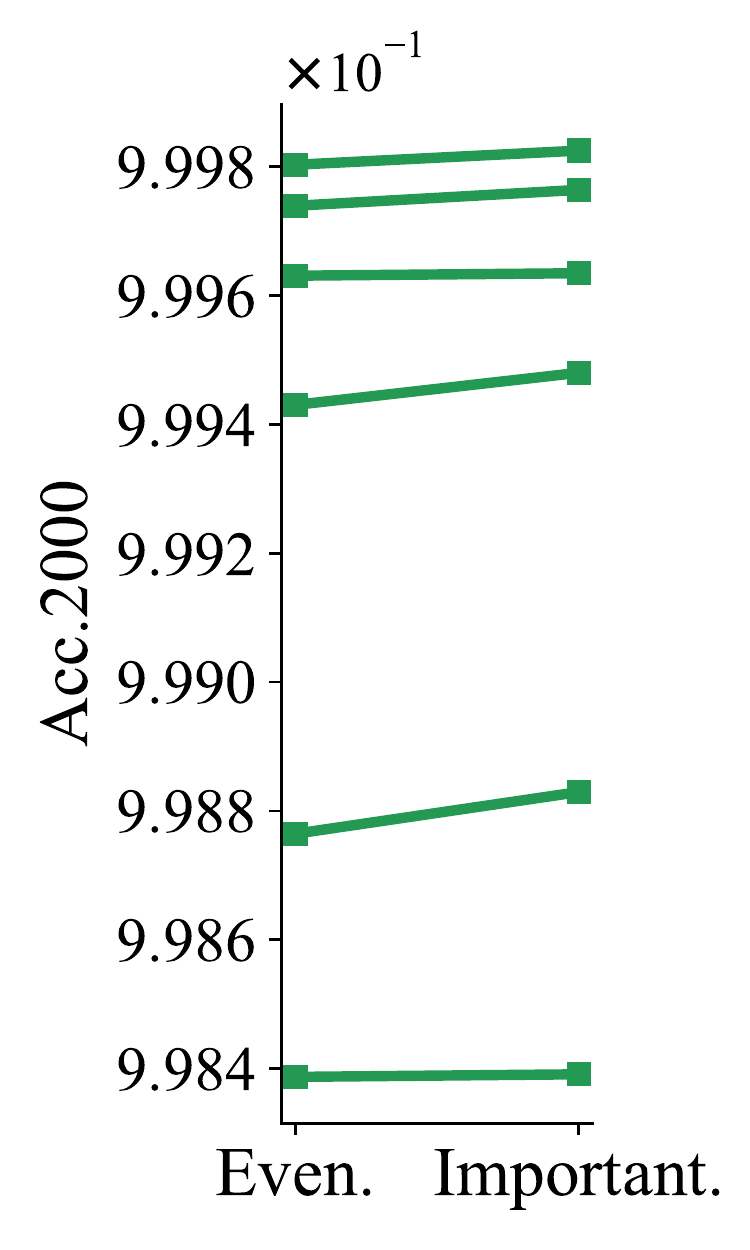}
     \caption{512$\times$}
 \end{subfigure}
\caption{Comparison of different intra-level parameter allocation approaches for 6 \textit{Neurons} data under different compression ratios. The compression ratios are labeled at the bottom of each sub-figure.}
\label{figsup: intra-level parameter allocation approaches for 6 Neurons data}
\end{figure}

\begin{figure}[H]
\centering
  \begin{subfigure}[b]{0.3\textwidth}
     \centering
     \includegraphics[width=\textwidth]{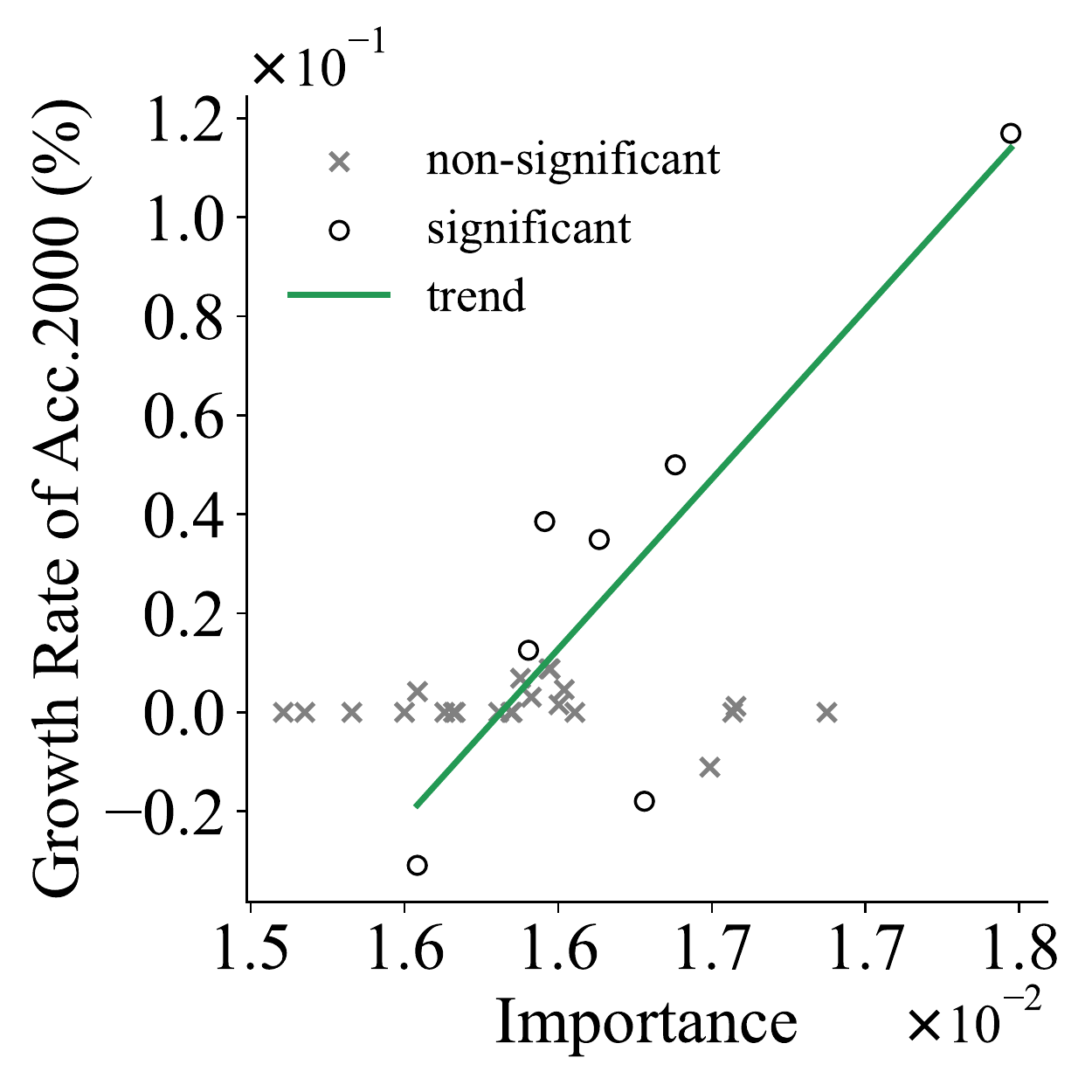}
     \caption{}
 \end{subfigure}
  \begin{subfigure}[b]{0.3\textwidth}
     \centering
     \includegraphics[width=\textwidth]{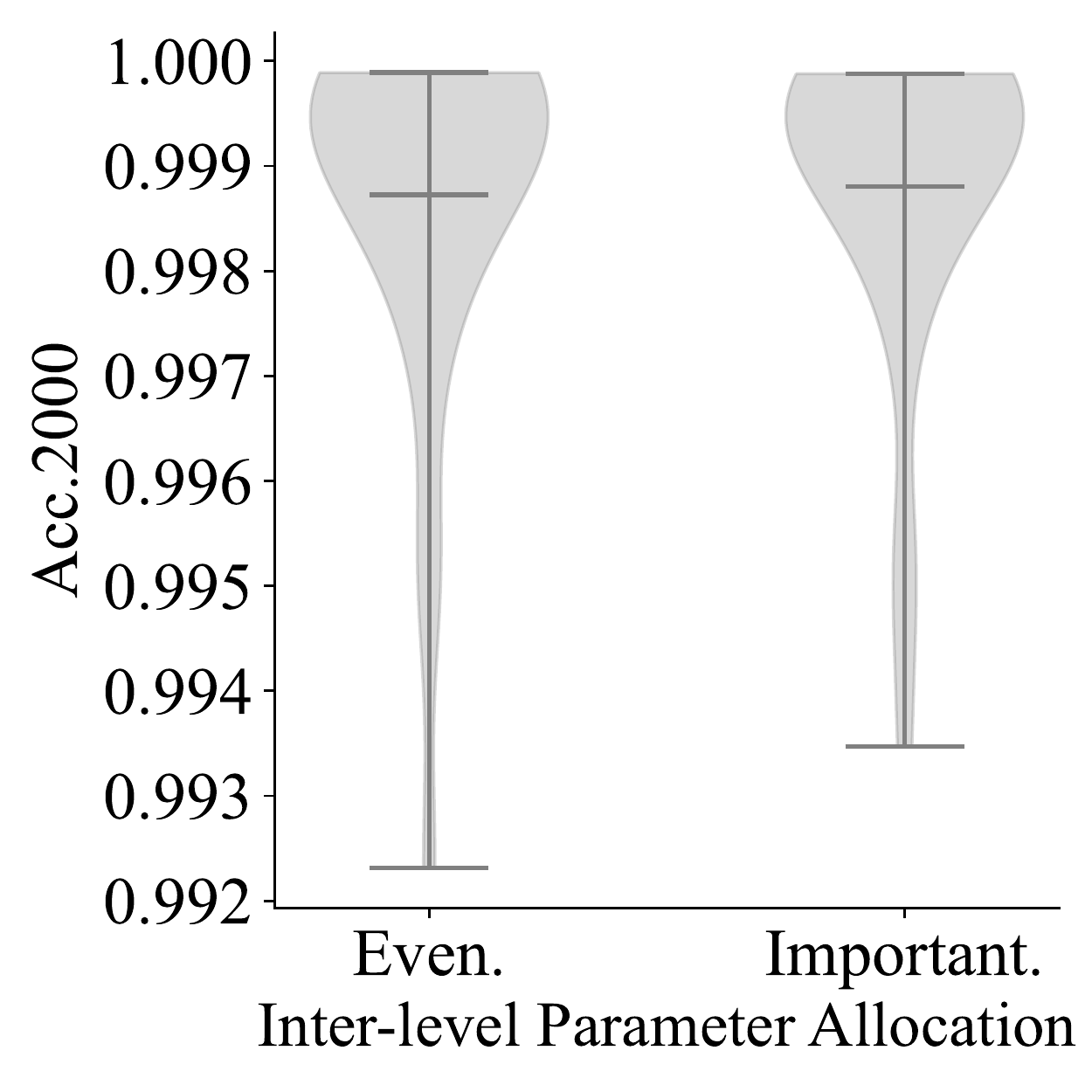}
     \caption{}
 \end{subfigure}
\caption{Effect of important allocation on \M's compression fidelity, in terms of Acc.2000 on a \textit{Neurons} data under 512$\times$ compression ratio.
(a) The growth rate of Acc.2000 for each region. The regions with and without significant growth rate are represented as cross marker and square respectively. The solid line represents the trend of change, estimated by a linear fit.
(b) The violin plots of Acc.2000 for all these regions. The shaded area represent the distribution of Acc.2000 across all regions. The three horizontal lines from top to bottom represent the maximum, average and minimum values respectively.}
\label{figsup: important allocation under 512 compression ratio}
\end{figure}

\newpage
\newpage
\newpage
\section*{Supplementary Tables}

\begin{table}[H]
\centering
\caption{The compression time and decompression time for each method on medical data under around 512$\times$ compression ratio. JPEG was excluded due to lower compression ratio. The decompression time were multiplied by 1,000 to simulate the real situation where the data needs to be decompressed frequently for processing and analysis}
\begin{tabular}{c||cccc||cc||ccc}
\hline
\multirow{2}{*}{Methods} &
  \multicolumn{4}{c||}{INR based} &
  \multicolumn{2}{c||}{Commercial} &
  \multicolumn{3}{c}{Data driven} \\ \cline{2-10} 
 &
  \multicolumn{1}{c|}{TINC(ours)} &
  \multicolumn{1}{c|}{SCI} &
  \multicolumn{1}{c|}{NeRV} &
  NeRF &
  \multicolumn{1}{c|}{H.264} &
  HEVC &
  \multicolumn{1}{c|}{DVC} &
  \multicolumn{1}{c|}{SGA+BB} &
  SSF \\ \hline
\begin{tabular}[c]{@{}c@{}}Compression\\ (seconds)\end{tabular} &
  \multicolumn{1}{c|}{2310} &
  \multicolumn{1}{c|}{2913} &
  \multicolumn{1}{c|}{2392} &
  489.2 &
  \multicolumn{1}{c|}{1.670} &
  1.000 &
  \multicolumn{1}{c|}{58.82} &
  \multicolumn{1}{c|}{2701} &
  7.086 \\ \hline
\begin{tabular}[c]{@{}c@{}}Decompression\\ *1k(seconds)\end{tabular} &
  \multicolumn{1}{c|}{217.4} &
  \multicolumn{1}{c|}{160.8} &
  \multicolumn{1}{c|}{195.3} &
  99.15 &
  \multicolumn{1}{c|}{370.1} &
  276.8 &
  \multicolumn{1}{c|}{183.0} &
  \multicolumn{1}{c|}{280.0} &
  400.0 \\ \hline
\end{tabular}
\label{tab: speed on medical}
\end{table}

\begin{table}[H]
\centering
\caption{The compression time and decompression time for each method on biological data under around 512$\times$ compression ratio with the same layout as Supplementary Table.~\ref{tab: speed on medical}}
\begin{tabular}{c||cccc||cc||ccc}
\hline
\multirow{2}{*}{Methods} &
  \multicolumn{4}{c||}{INR based} &
  \multicolumn{2}{c||}{Commercial} &
  \multicolumn{3}{c}{Data driven} \\ \cline{2-10} 
 &
  \multicolumn{1}{c|}{TINC(ours)} &
  \multicolumn{1}{c|}{SCI} &
  \multicolumn{1}{c|}{NeRV} &
  NeRF &
  \multicolumn{1}{c|}{H.264} &
  HEVC &
  \multicolumn{1}{c|}{DVC} &
  \multicolumn{1}{c|}{SGA+BB} &
  SSF \\ \hline
\begin{tabular}[c]{@{}c@{}}Compression\\ (seconds)\end{tabular} &
  \multicolumn{1}{c|}{1999} &
  \multicolumn{1}{c|}{1445} &
  \multicolumn{1}{c|}{1603} &
  769.6 &
  \multicolumn{1}{c|}{0.900} &
  2.414 &
  \multicolumn{1}{c|}{38.99} &
  \multicolumn{1}{c|}{3231} &
  3.940 \\ \hline
\begin{tabular}[c]{@{}c@{}}Decompression\\ *1k(seconds)\end{tabular} &
  \multicolumn{1}{c|}{216.3} &
  \multicolumn{1}{c|}{201.1} &
  \multicolumn{1}{c|}{98.64} &
  84.43 &
  \multicolumn{1}{c|}{448.6} &
  501.1 &
  \multicolumn{1}{c|}{295.6} &
  \multicolumn{1}{c|}{279.3} &
  127.7 \\ \hline
\end{tabular}
\label{tab: speed on biological}
\end{table}

\end{document}